\documentclass[journal]{IEEEtran}
\usepackage{cite}
\usepackage{epsfig}
\usepackage{epstopdf}
\usepackage{bm}
\usepackage{amsmath}
\usepackage{amssymb}
\usepackage{graphicx}
\usepackage{subfigure}
\usepackage{overpic}
\usepackage{booktabs}
\usepackage{multirow}
\usepackage{threeparttable}
\usepackage{enumitem}
\usepackage{color}
\usepackage[x11names]{xcolor}
\usepackage{caption}
\usepackage[pagebackref=true,breaklinks=true,colorlinks,bookmarks=false]{hyperref}

\DeclareCaptionLabelSeparator{periodspace}{.\quad}
\def\Fix#1{{\color{black}{{#1}}}}

\newcommand\eg{\emph{e.g.}} 
 
\newcommand\ien{\emph{i.e.}}
\newcommand\cf{\emph{cf.}} 
\newcommand\etc{\emph{etc.}}

\newcommand\etal{\emph{et al.}}

\begin{document}

\title{Learning Dense and Continuous Optical Flow from an Event Camera}

\author{Zhexiong~Wan,
        Yuchao~Dai,~\IEEEmembership{Member,~IEEE,}
        and Yuxin Mao% <-this % stops a space
\thanks{This research was supported in part by National Natural Science Foundation of China (62271410, 61871325, 61901387), National Key Research and Development Program of China (2018AAA0102803) and Zhejiang Lab (NO.2021MC0AB05).}
\thanks{Zhexiong Wan (wanzhexiong@mail.nwpu.edu.cn), Yuchao Dai (corresponding author, daiyuchao@nwpu.edu.cn) and Yuxin Mao are with School of Electronics and Information, Northwestern Polytechnical University, Xi'an, Shaanxi, 710129, China.}
\thanks{Digital Object Identifier 10.1109/TIP.2022.3220938}
\thanks{Project page https://npucvr.github.io/DCEIFlow/}
}

\markboth{IEEE Transactions on Image Processing,~Vol.~xx, No.~xx, November~2022}%
{Shell \MakeLowercase{\textit{et al.}}: Bare Demo of IEEEtran.cls for IEEE Journals}

\maketitle

\begin{abstract}
Event cameras such as DAVIS can simultaneously output high temporal resolution events and low frame-rate intensity images, which own great potential in capturing scene motion, such as optical flow estimation. Most of the existing optical flow estimation methods are based on two consecutive image frames and can only estimate \emph{discrete flow} at a fixed time interval. Previous work has shown that \emph{continuous flow} estimation can be achieved by changing the quantities or time intervals of events. However, they are difficult to estimate reliable \emph{dense flow}, especially in the regions without any triggered events. In this paper, we propose a novel deep learning-based dense and continuous optical flow estimation framework from a single image with event streams, which facilitates the accurate perception of high-speed motion. Specifically, we first propose an event-image fusion and correlation module to effectively exploit the internal motion from two different modalities of data. Then we propose an iterative update network structure with bidirectional training for optical flow prediction. Therefore, our model can estimate reliable dense flow as two-frame-based methods, as well as estimate temporal continuous flow as event-based methods. Extensive experimental results on both synthetic and real captured datasets demonstrate that our model outperforms existing event-based state-of-the-art methods and our designed baselines for accurate dense and continuous optical flow estimation. 
\end{abstract}


\IEEEpeerreviewmaketitle

%%%%%%%%%%%%%%%%%%%%%%%%%%% Introduction %%%%%%%%%%%%%%%%%%%%%%
\section{Introduction}

\IEEEPARstart{E}{vent} cameras are bio-inspired vision sensors that can trigger brightness change asynchronously and independently at each pixel with a microsecond time resolution~\cite{event:Gallego_eventsurvey_TPAMI_2020}.
Unlike the conventional frame-based shutter cameras that capture full resolution images at a fixed frame rate, event cameras such as DVS~\cite{eventdevice:dvscamera} and DAVIS~\cite{eventdevice:daviscamera} can output a discrete event stream at a very small and not fixed event rate.
In particular, the DAVIS event camera~\cite{eventdevice:daviscamera} can simultaneously output image and event streams.
The event data stream can be regarded as a frame sequence with up to millions of frames-per-second (fps)~\cite{eventapp:Rebecq_HighSpeedHDREvent_TPAMI_2019, eventapp:Pan_BringingBlurryEvent_CVPR_2019, eventapp:Sabatier_EventFourier_TIP_2017}, which owns appealing advantages over the shutter frames, including high temporal resolution, high dynamic range, low latency, low redundancy, and low power consumption. 
These enable the broad applications of event cameras in feature tracking~\cite{eventapp:Tedaldi_Featuredavistrack_2016, eventapp:Gehrig_Asynchronouetrackdavis_eccv_2018, eventapp:Kueng_Davisfeaturetrack_IROS_2016}, depth estimation and 3D reconstruction~\cite{eventapp:Tulyakov_learningeventdepth_2019, eventapp:Kim_real6dof_eccv_2016, eventapp:Zhou_semieventreconstruction_eccv_2018, eventapp:Gallego_Event6dof_TPAMI_2017}, frame synthesis~\cite{eventapp:Rebecq_HighSpeedHDREvent_TPAMI_2019, eventapp:Jiang_learningeventdeblur_2020, eventapp:Cadena_SPADE_E2VID_TIP_2021}, \etc

\begin{table}[tbp]
\setlength\tabcolsep{1pt}
% \small
\centering
  \begin{tabular}{p{1.6cm} p{1.4cm} p{1.4cm} p{1.4cm} p{1.4cm} p{1.4cm}}
    \begin{minipage}[b]{1.5cm}
		\centering
		Image1 \\
		\raisebox{-.5\height}{($\bf{I_1}$)}
	\end{minipage}& 
    \begin{minipage}[b]{1.4cm}
		\centering
		\raisebox{-.5\height}{\includegraphics[width=\linewidth]{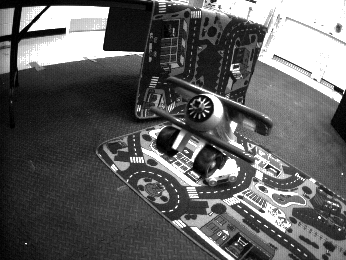}}
	\end{minipage}& 
    \begin{minipage}[b]{1.4cm}
		\centering
		\raisebox{-.5\height}{\includegraphics[width=\linewidth]{viz/continuous/table_seq_00_198/table_seq_00_198_image1.png}}
	\end{minipage}& 
    \begin{minipage}[b]{1.4cm}
		\centering
		\raisebox{-.5\height}{\includegraphics[width=\linewidth]{viz/continuous/table_seq_00_198/table_seq_00_198_image1.png}}
	\end{minipage}& 
    \begin{minipage}[b]{1.4cm}
		\centering
		\raisebox{-.5\height}{\includegraphics[width=\linewidth]{viz/continuous/table_seq_00_198/table_seq_00_198_image1.png}}
	\end{minipage}& 
    \begin{minipage}[b]{1.4cm}
		\centering
		\raisebox{-.5\height}{\includegraphics[width=\linewidth]{viz/continuous/table_seq_00_198/table_seq_00_198_image1.png}}
	\end{minipage} \cr
    \specialrule{0em}{0.5pt}{0.5pt}

    \begin{minipage}[b]{1.5cm}
		\centering
		Image2 \\
		\raisebox{-.5\height}{($\bf{I_2}$)}
	\end{minipage}& 
    \begin{minipage}[b]{1.4cm}
		\centering
		-
	\end{minipage}& 
    \begin{minipage}[b]{1.4cm}
		\centering
		\raisebox{-.5\height}{\includegraphics[width=\linewidth]{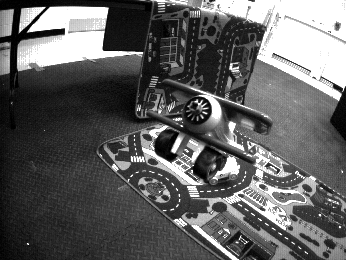}}
	\end{minipage}& 
    \begin{minipage}[b]{1.4cm}
		\centering
		-
	\end{minipage}& 
    \begin{minipage}[b]{1.4cm}
		\centering
		-
	\end{minipage}& 
    \begin{minipage}[b]{1.4cm}
		\centering
		\raisebox{-.5\height}{\includegraphics[width=\linewidth]{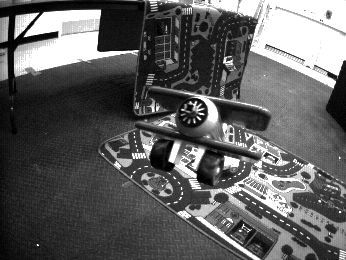}}
	\end{minipage} \cr
    \specialrule{0em}{0.5pt}{0.5pt}

    \begin{minipage}[b]{1.5cm}
		\centering
        \raisebox{-.5\height}{PWC-Net} \raisebox{-.5\height}{\cite{flow:Sun_PWCNet_TPAMI_2019} ($\bf{I_1}$+$\bf{I_2}$)}
	\end{minipage}& 
    \begin{minipage}[b]{1.4cm}
		\centering
		-
	\end{minipage}& 
    \begin{minipage}[b]{1.4cm}
		\centering
		\raisebox{-.5\height}{\includegraphics[width=\linewidth]{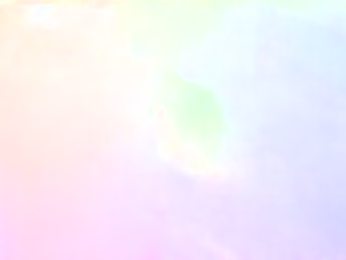}}
	\end{minipage}& 
    \begin{minipage}[b]{1.4cm}
		\centering
		-
	\end{minipage}& 
    \begin{minipage}[b]{1.4cm}
		\centering
		-
	\end{minipage}& 
    \begin{minipage}[b]{1.4cm}
		\centering
		\raisebox{-.5\height}{\includegraphics[width=\linewidth]{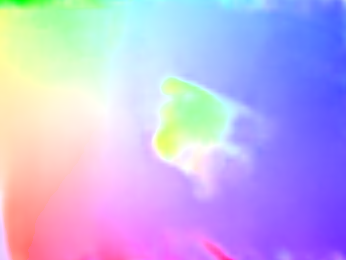}}
	\end{minipage} \cr
    \specialrule{0em}{0.5pt}{0.5pt}

    \begin{minipage}[b]{1.5cm}
		\centering
        \raisebox{-.5\height}{RAFT}
        \raisebox{-.5\height}{\cite{flow:Teed_RAFT_ECCV_2020} (\textbf{$\bf{I_1}$+$\bf{I_2}$})}
	\end{minipage}& 
    \begin{minipage}[b]{1.4cm}
		\centering
		-
	\end{minipage}& 
    \begin{minipage}[b]{1.4cm}
		\centering
		\raisebox{-.5\height}{\includegraphics[width=\linewidth]{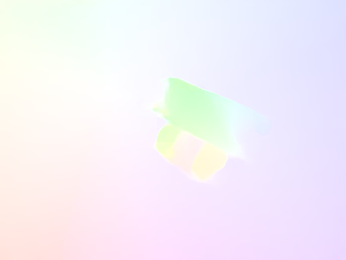}}
	\end{minipage}& 
    \begin{minipage}[b]{1.4cm}
		\centering
		-
	\end{minipage}& 
    \begin{minipage}[b]{1.4cm}
		\centering
		-
	\end{minipage}& 
    \begin{minipage}[b]{1.4cm}
		\centering
		\raisebox{-.5\height}{\includegraphics[width=\linewidth]{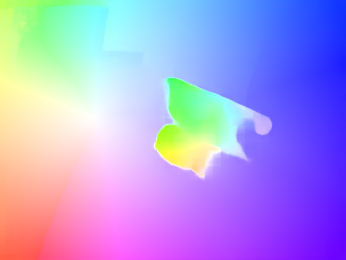}}
	\end{minipage} \cr
    \specialrule{0em}{0.5pt}{0.5pt}

    \begin{minipage}[b]{1.5cm}
		\centering
        Events \\
		\raisebox{-.5\height}{($\bf{E}$)}
	\end{minipage}& 
    \begin{minipage}[b]{1.4cm}
		\centering
		\raisebox{-.5\height}{\includegraphics[width=\linewidth]{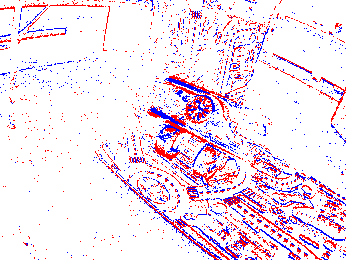}}
	\end{minipage}& 
    \begin{minipage}[b]{1.4cm}
		\centering
		\raisebox{-.5\height}{\includegraphics[width=\linewidth]{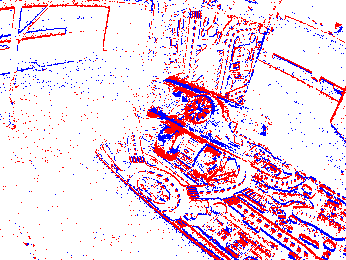}}
	\end{minipage}& 
    \begin{minipage}[b]{1.4cm}
		\centering
		\raisebox{-.5\height}{\includegraphics[width=\linewidth]{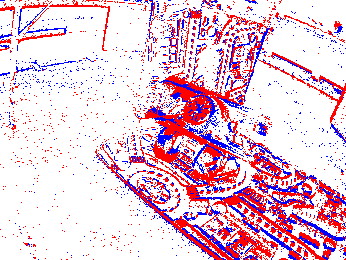}}
	\end{minipage}& 
    \begin{minipage}[b]{1.4cm}
		\centering
		\raisebox{-.5\height}{\includegraphics[width=\linewidth]{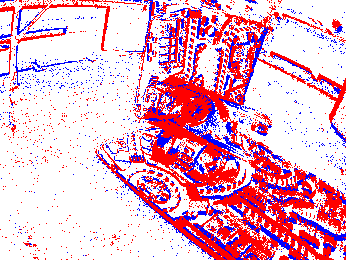}}
	\end{minipage}& 
    \begin{minipage}[b]{1.4cm}
		\centering
		\raisebox{-.5\height}{\includegraphics[width=\linewidth]{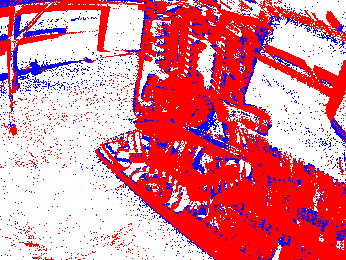}}
	\end{minipage} \cr
    \specialrule{0em}{0.5pt}{0.5pt}

    \begin{minipage}[b]{1.5cm}
		\centering
        \raisebox{-.5\height}{\scriptsize SpikeFlowNet} \raisebox{-.5\height}{\cite{eventflow:Lee_SpikeFlowNet_ECCV_2020} ($\bf{E}$)}
	\end{minipage}& 
    \begin{minipage}[b]{1.4cm}
		\centering
		\raisebox{-.5\height}{\includegraphics[width=\linewidth]{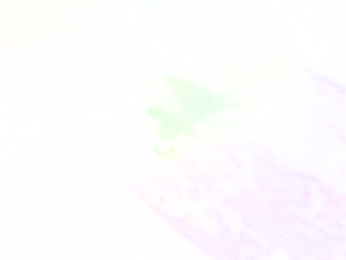}}
	\end{minipage}& 
    \begin{minipage}[b]{1.4cm}
		\centering
		\raisebox{-.5\height}{\includegraphics[width=\linewidth]{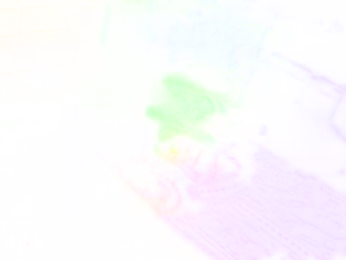}}
	\end{minipage}& 
    \begin{minipage}[b]{1.4cm}
		\centering
		\raisebox{-.5\height}{\includegraphics[width=\linewidth]{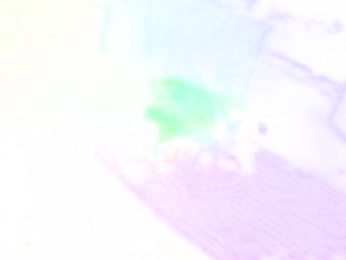}}
	\end{minipage}& 
    \begin{minipage}[b]{1.4cm}
		\centering
		\raisebox{-.5\height}{\includegraphics[width=\linewidth]{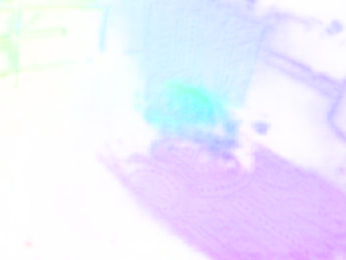}}
	\end{minipage}& 
    \begin{minipage}[b]{1.4cm}
		\centering
		\raisebox{-.5\height}{\includegraphics[width=\linewidth]{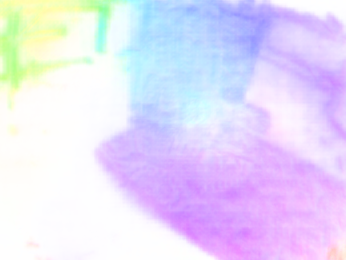}}
	\end{minipage} \cr
    \specialrule{0em}{0.5pt}{0.5pt}

    \begin{minipage}[b]{1.4cm}
		\centering
        \raisebox{-.5\height}{\scriptsize Stoffregen~\etal} \raisebox{-.5\height}{\cite{eventflow:Stoffregen_ReducingGAP_ECCV_2020} ($\bf{E}$)}
	\end{minipage}& 
    \begin{minipage}[b]{1.4cm}
		\centering
		\raisebox{-.5\height}{\includegraphics[width=\linewidth]{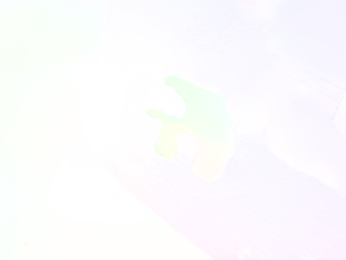}}
	\end{minipage}& 
    \begin{minipage}[b]{1.4cm}
		\centering
		\raisebox{-.5\height}{\includegraphics[width=\linewidth]{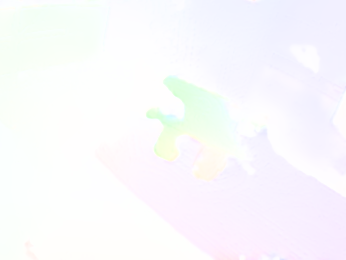}}
	\end{minipage}& 
    \begin{minipage}[b]{1.4cm}
		\centering
		\raisebox{-.5\height}{\includegraphics[width=\linewidth]{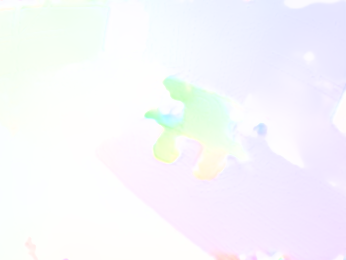}}
	\end{minipage}& 
    \begin{minipage}[b]{1.4cm}
		\centering
		\raisebox{-.5\height}{\includegraphics[width=\linewidth]{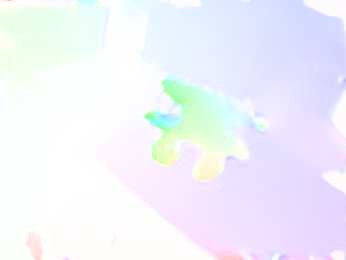}}
	\end{minipage}& 
    \begin{minipage}[b]{1.4cm}
		\centering
		\raisebox{-.5\height}{\includegraphics[width=\linewidth]{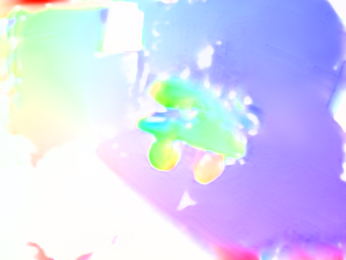}}
	\end{minipage} \cr
    \specialrule{0em}{0.5pt}{0.5pt}

    \begin{minipage}[b]{1.5cm}
		\centering
        \raisebox{-.5\height}{\scriptsize DCEIFlow} \raisebox{-.5\height}{{\scriptsize Ours}~($\bf{I_1}$+$\bf{E}$)}
	\end{minipage}& 
    \begin{minipage}[b]{1.4cm}
		\centering
		\raisebox{-.5\height}{\includegraphics[width=\linewidth]{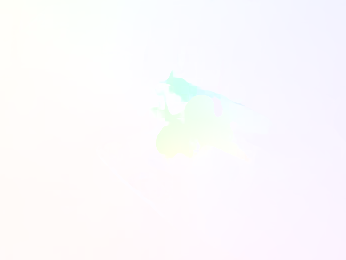}}
	\end{minipage}& 
    \begin{minipage}[b]{1.4cm}
		\centering
		\raisebox{-.5\height}{\includegraphics[width=\linewidth]{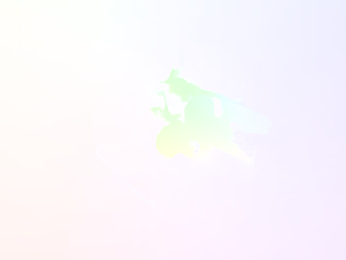}}
	\end{minipage}& 
    \begin{minipage}[b]{1.4cm}
		\centering
		\raisebox{-.5\height}{\includegraphics[width=\linewidth]{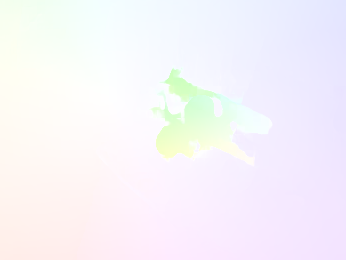}}
	\end{minipage}& 
    \begin{minipage}[b]{1.4cm}
		\centering
		\raisebox{-.5\height}{\includegraphics[width=\linewidth]{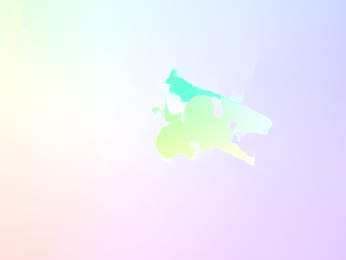}}
	\end{minipage}& 
    \begin{minipage}[b]{1.4cm}
		\centering
		\raisebox{-.5\height}{\includegraphics[width=\linewidth]{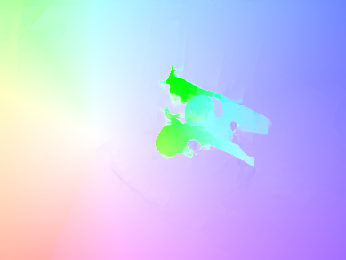}}
	\end{minipage} \cr
    \specialrule{0em}{0.5pt}{0.5pt}

    & 
    \begin{minipage}[b]{1.4cm}
		\centering
		$dt$=0.6
	\end{minipage}& 
    \begin{minipage}[b]{1.4cm}
		\centering
		$dt$=1.0
	\end{minipage}& 
    \begin{minipage}[b]{1.4cm}
		\centering
		$dt$=1.8
	\end{minipage}& 
    \begin{minipage}[b]{1.4cm}
		\centering
		$dt$=2.3
	\end{minipage}& 
    \begin{minipage}[b]{1.4cm}
		\centering
		$dt$=5.0
	\end{minipage} \cr
    \specialrule{0em}{0.5pt}{0.5pt}
  \end{tabular}

  \vspace{-8pt}
  \captionof{figure}[foo]{\textbf{Visual comparisons of continuous flow prediction with different time intervals}. $dt$ denotes the frame interval (\eg, $dt$=1.0 represents the time interval between two adjacent frames). The two-frame ($\bf{I_1}$+$\bf{I_2}$) approaches can only estimate dense optical flow between frames. The event-only ($\bf{E}$) approaches can estimate continuous optical flow, but cannot predict accurate dense flow. Our model can estimate both dense and continuous optical flow by fusing events and the first image ($\bf{I_1}$+$\bf{E}$). Best viewed on screen.
  }
  \vspace{-10pt}
  \label{viz:continuous_viz}
\end{table}

\begin{figure}[tbp] 
\centering 
\includegraphics[width=0.5\textwidth]{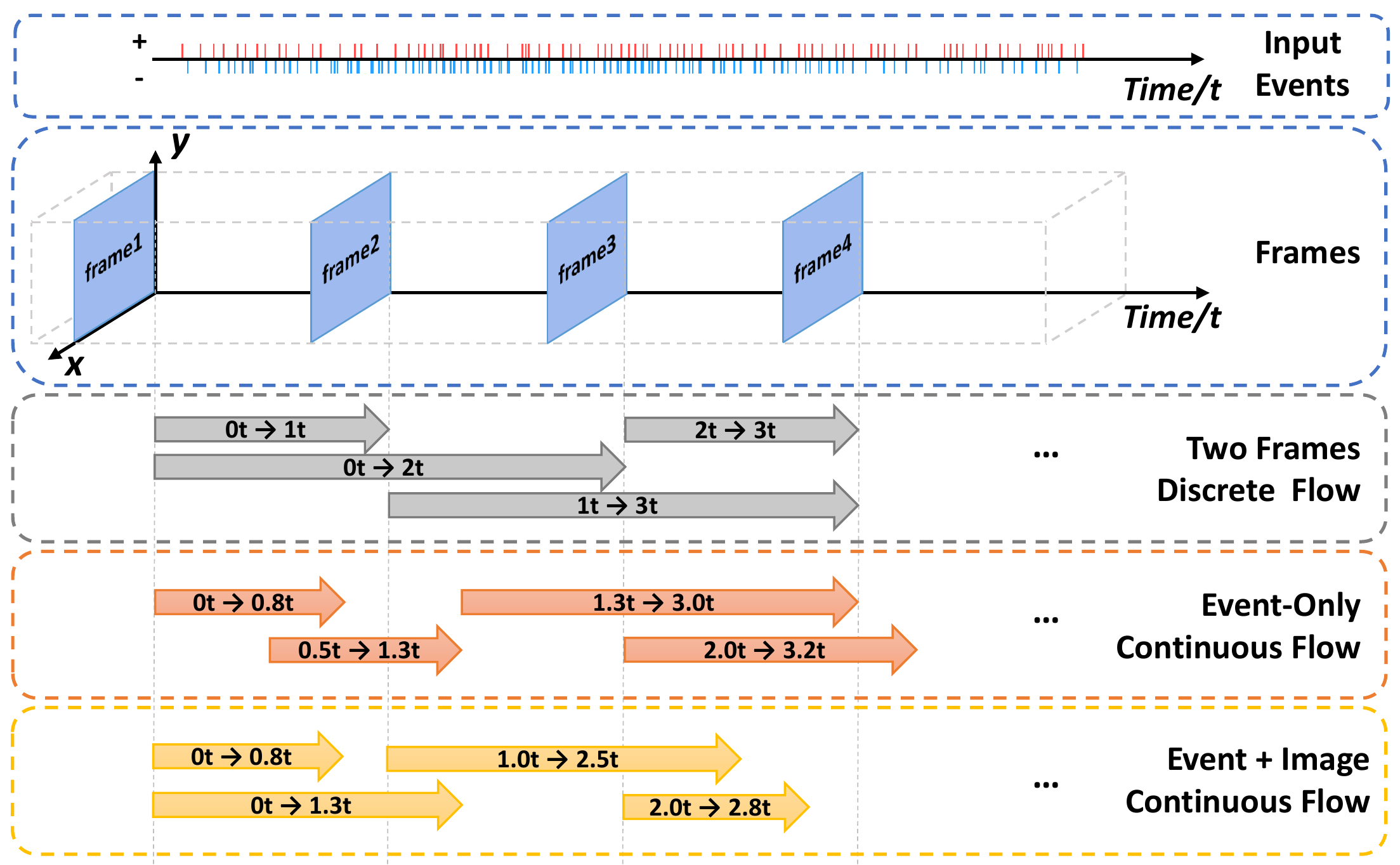}
\vspace{-15pt} 
\caption{Differences between \textbf{Discrete Flow} and \textbf{Continuous Flow}. Given a set of frames and event streams captured by a DAVIS camera. The two-frame optical flow estimation methods can only work with discrete integer frame intervals. The event-only approach can estimate continuous flow from any start to end time. Bringing the first image with events, we can still estimate continuous flow, but the start time is limited to the time of each frame.
}
\vspace{-10pt}
\label{fig:continuous_vs_discrete}
\end{figure}

Optical flow estimation aims to predict the motion between two moments by exploiting the photometric consistency. Most of the existing event-based optical flow estimation approaches~\cite{eventflow:Zhu_EVFlowNet_RSS_2018, eventflow:Zhu_EVFlowNet_CVPR_2019, eventflow:Lee_SpikeFlowNet_ECCV_2020, eventflow:Gallego_Unifyingcontrastmax_cvpr_2018, eventflow:Paredes_UnsupervisedHierarchicalFlow_TPAMI_2019} only use event streams. 
Although temporal continuous optical flow can be predicted, it is difficult to get reliable predictions in regions without any events, as shown in Fig.~\ref{viz:continuous_viz}. 
Thus, we consider fusing a single image with events to improve the reliability of dense optical flow estimation.
Due to the frame rate limitation of the first image, we cannot estimate the continuous flow from any start to end time like the event-only methods.
However, we can still estimate continuous and reliable dense flow at varying time intervals from a fixed frame. 
We have shown a schematic diagram in Fig.~\ref{fig:continuous_vs_discrete} to illustrate the differences between these three types of input settings.
The estimated continuous flow from a single image with events has notable significance for many event-based downstream applications, especially those associated with images, including image deblurring~\cite{eventapp:Pan_BringingBlurryEvent_CVPR_2019}, video synthesis~\cite{eventapp:tulyakov_timelens_CVPR_2021}, feature tracking~\cite{eventapp:Gehrig_Asynchronouetrackdavis_eccv_2018}, \etc

In this paper, we propose a novel deep learning-based optical flow estimation model from a single image with event streams, named DCEIFlow, which can effectively exploit the internal relation from two different modalities of data through our proposed event-image fusion and correlation module. Thus, our proposed DCEIFlow model can estimate reliable dense flow as two-frame-based methods, as well as estimate continuous flow as event-only methods.
In particular, we first propose an event-image fusion module to effectively fuse the features of the first image and events by a multi-layer convolution fusion network. 
The fused feature is regarded as a pseudo second image feature that is constrained by the real second image feature in training. Then we use it to construct the feature correlation. On this basis, we propose a network with an iterative update structure to learn the optical flow from the constructed correlations.
In addition, we propose a bidirectional flow training mechanism based on the reverse event, which can use the same network to estimate the backward flow by inputting the second image with the reversed event streams during the training stage. 

Due to the lack of an event dataset with dense flow annotation, we first pre-train our model on the two-frame dataset FlyingChairs2~\cite{flowdatasets:Ilg_FlyingChairs2_ECCV_2018} with simulated events, and then evaluate the pre-trained model on the real captured dataset MVSEC~\cite{eventdatasets:Zhu_MVSEC_RAL_2018}.
Evaluation results and visual comparisons on both synthetic and real captured datasets show a significant improvement over the existing event-only or fused single image state-of-the-arts.
Specifically, our pre-trained model achieves better results on MVSEC than existing methods with different time intervals ($dt$=1 and $dt$=4 frames).
We also compare our model with two baseline networks, one with only input events and the other directly concatenating the image feature with events. 
The results show that our model achieves better results with a smaller model size than the baseline methods. 
In addition, we perform visual comparisons on a highly dynamic real captured dataset EV-IMO~\cite{eventdatasets:mitrokhin_EVIMO_IROS_2019}. 
The results show that our model is superior to existing event-based methods in dense and accurate flow estimation and has advantages over existing two-frame-based methods for detailed optical flow estimation.

Our main contributions are summarized as follows:
\begin{enumerate}[fullwidth,itemindent=1em,label=(\arabic*)]
    \item We propose a novel deep learning-based dense and continuous optical flow estimation model from a single image with event streams, which can estimate reliable dense flow as the two-frame-based approaches, as well as estimate temporal continuous flow as the event-only approaches.
    \item We propose to build an event-image correlation to effectively exploit internal motion from two different modalities of data. We also propose bidirectional flow training based on reverse events to leverage the order of motion information in events.
    \item Extensive experiments on the MVSEC~\cite{eventdatasets:Zhu_MVSEC_RAL_2018} and EV-IMO~\cite{eventdatasets:mitrokhin_EVIMO_IROS_2019} datasets demonstrate that our proposed DCEIFlow model improves significantly compared to baselines and the existing event-based state-of-the-arts. We also verify the superiority of our method in dense and continuous optical flow estimation through further experiments and analysis.
\end{enumerate}

%%%%%%%%%%%%%%%%%%% Related Work %%%%%%%%%%%%%%%%%%%%%
\section{Related Work}
Optical flow estimation is a very active research area in computer vision, where various approaches have been proposed. In this section, we first review the development of learning-based two-frame optical flow estimation. Then we focus on event-based optical flow estimation, including dense and sparse flow from events or with a single image.

\subsection{Two-Frame-based Optical Flow Estimation}
Recently, the success of deep Convolutional Neural Networks (CNNs) has been extended to various computer vision tasks such as optical flow estimation~\cite{flow:hur_OpticalInDeepAge_2020}. 
The first end-to-end CNN regression approach for estimating optical flow is  FlowNet~\cite{flowdatasets:Dosovitskiy_FlowNet_FlyingChairs1_CVPR_2015}, which directly estimates flow from a pair of input images based on an encoder-decoder architecture. It has achieved a faster inference speed than optimization-based methods with higher accuracy. PWC-Net~\cite{flow:Sun_PWCNet_TPAMI_2019} exploits three well-known design principles from existing optimization approaches to deep learning scenarios, including pyramid structure, feature warping, and correlation construction. These key design principles have been widely used or improved in a series of recent works such as IRR-PWC~\cite{flow:Hur_IRR_CVPR_2019}, SelFlow~\cite{flow:Liu_SelFlow_CVPR_2019}, and VCN~\cite{flow:Yang_Volumetric_NeurIPS_2019}.
Recently, RAFT~\cite{flow:Teed_RAFT_ECCV_2020} introduced an all-pairs correlation iterative network structure and achieved significant improvements over existing methods.

\subsection{Event-based Optical Flow Estimation}
Since event streams only encode the pixel-level brightness changes discretely, they cannot directly represent the absolute brightness. It is difficult to find the spatial photometric consistency between sparse pixels and estimate dense optical flow. According to the input and output, the existing event-based methods can be divided into the following three categories.

\subsubsection{Sparse Flow from Events}
Before deep networks were widely used, Benosman~\etal~\cite{eventflow:Benosman_asynchronouseventflow_NN_2012} first proposed an event-based optical flow algorithm based on the Lucas-Kanade~\cite{lucaskanade_LKFlow_IJCAI_1981} brightness constancy assumption. However, it can only estimate the normal flow component perpendicular to the edge because the event data is usually triggered at the moving edge. 
After that, \cite{eventflow:Liu_AdaptiveFlow_BMCV_2018, eventflow:Gallego_Unifyingcontrastmax_cvpr_2018} can estimate full flow, which introduce tangential flow and contain more motion information compared to normal flow~\cite{event:Gallego_eventsurvey_TPAMI_2020}. 
Recently, \cite{eventflow:Paredes_UnsupervisedHierarchicalFlow_TPAMI_2019} proposes to use SNNs~\cite{Maass_networks_SNN_NN_1997} to estimate sparse flow efficiently, but it is not widely used because of the limited application of sparse flow.

\subsubsection{Dense Flow from Events}
Recent event-based methods tend to estimate dense optical flow, which can provide more spatial information than sparse flow.
EV-FlowNet~\cite{eventflow:Zhu_EVFlowNet_RSS_2018} is an end-to-end optical flow network learning from events in a self-supervised manner. 
It uses the grey image captured by DAVIS as the unsupervised supervision in training.
After that, Zhu~\etal~\cite{eventflow:Zhu_EVFlowNet_CVPR_2019} proposed an unsupervised training framework by using the predicted flow to remove the motion blur in the input events.
EST~\cite{eventapp:Gehrig_event_representations_ICCV_2019}, Matrix-LSTM~\cite{eventflow:Cannici_Matrix-LSTM_ECCV_2020} explore different event representations. SpikeFlowNet~\cite{eventflow:Lee_SpikeFlowNet_ECCV_2020}, LIF-EV-FlowNet~\cite{eventflow:Hagenaars_SelfSpikeFlow_Neurips_2021}, STE-FlowNet~\cite{eventflow:Ding_STEFlowNet_aaai_2022}, E-RAFT~\cite{eventflow:Gehrig_DenseRAFTFlow_3DV_2021} and Li~\etal~\cite{eventflow:Li_LightWeightEventFlow_ICPR_2021} explore the effects of introducing SNNs, recurrent structure and reducing network parameters.
Because they only use events, the predicted dense flow in the regions without any triggered events, such as the constant brightness region, is relatively unreliable compared to the regions with events. 

\subsubsection{Dense Flow from Single Image with Events}
In order to obtain more reliable dense estimates, researchers consider combining events with the image which contains per-pixel absolute intensity. Bardow~\etal~\cite{eventflow:Bardow_simultaneousflowintensity_2016} jointly reconstructed the intensity image and estimated flow from events, but the accuracy of flow depends on the image reconstruction quality.
Pan~\etal~\cite{eventflow:Pan_SingleImageFlow_CVPR_2020} proposed to jointly use a set of events with a single image and introduced an event-based brightness constancy as the objective function for optimization.
However, these non-convex optimization-based methods are not only time-consuming but also require complex post-processing and tuning. 
Very recently, Fusion-FlowNet~\cite{eventflow:Lee_Fusion_FlowNet_ICRA_2022} directly inputs events and an image to an end-to-end dual-branch fusion network. This concatenate fusion scheme is simple to explore the internal relationship between two modalities of data, and their results can not show the advantages of introducing the image.

%%%%%%%%%%%%%%%%%%% Approach %%%%%%%%%%%%%%%%%%%%%
\section{Approach}
Given the event streams and first image, we build a learning-based framework for estimating dense and continuous optical flow.
Our framework consists of five stages: (1) event volume representation, (2) event and image feature extraction, (3) event-image feature fusion, (4) event-image all-pairs correlation construction, and (5) iterative flow updater.
Based on this, we propose a bidirectional optical flow training mechanism to constrain the training process.

\subsection{Preliminaries}

\subsubsection{Event camera model}
The output data streams of the event camera can be regarded as a finite quaternion sequence, which represents the per-pixel brightness changes during a period of time. Each event contains its space-time coordinate and a binary polarity representing its brightness change:
\begin{equation}
\begin{aligned}
e_i&=\{\mathbf{x}_i, t_i, p_i\},
\end{aligned}
\end{equation}
\begin{equation}
|\log (L(\mathbf{x},t + \Delta t)) - \log (L(\mathbf{x},t)| \ge c ,
\end{equation}
where $L(\mathbf{x}_i,t_i)$ is the brightness at camera coordinate $\mathbf{x}_i=(x_i, y_i)$ and microsecond timestamp $t_i$. When the logarithmic domain brightness changes reach the threshold $c$ after $\Delta t$ time, an event $e$ is triggered. The polarity $p=\pm 1$ indicates the direction of brightness change.

\subsubsection{\textbf{Discrete Flow} and \textbf{Continuous Flow}}
Within the standard setup, the optical flow is estimated from two image frames to represent the displacement of the corresponding pixels from the first frame to the second frame.
Because the images from the shutter camera are usually at a fixed frame rate, two-frame-based methods can only estimate \textbf{discrete flow} with discrete integer frame intervals. 
Therefore, it has a limited ability to explain the motion with a high temporal resolution, such as the motion within frames. 
Here, we make a comparison for these different settings in Fig.~\ref{fig:continuous_vs_discrete}.
Note that the discrete flow in our paper is defined in the temporal domain, and it is different from the discrete flow defined in the previous two-frame-based approaches~\cite{flow:Lei_discrete_optimization_ctf_flow_ICCV_2009, flow:Guney_deepdiscreteflow_ACCV_2016}, which represents estimating optical flow by discrete optimization.

Therefore, it is difficult to estimate the \textbf{continuous flow} with variable time intervals under the two-frame setting. In contrast, the event cameras capture the brightness change of each pixel at high time resolution. It can represent the reliable internal motion within frames with high temporal resolution. Thus, we can use events to estimate the continuous flow with the theoretical frame rate as high as the event camera's \textit{eps} (events per second), which is very helpful for high-speed motion estimation or video analysis.

\subsubsection{Feature Correlation for Matching}
\label{sec:localcorr}
Most recent two-frame optical flow estimation networks use correlation to represent the pixel level matching similarity of two image features from a siamese encoder. 
The correlation is usually constructed by calculating the dot product similarity between the feature of the first image and the feature of the second image warped by the coarse flow.

Here, we review the general construction of local correlation in the two-frame setting~\cite{flow:Sun_PWCNet_TPAMI_2019, flow:Hofinger_ImprovingPyramid_ECCV_2020}. For the given two feature maps $P_{I_1}$ and $P_{I_2}$ (generated by two input images through a siamese encoder) and their corresponding optical flow $\bm{F}^{1\rightarrow2}$, the correlation volume are  constructed as:
\begin{equation}
\begin{aligned}
&C_{I}(\mathbf{x},\bm{\delta}_{uv})=P_{I_1}(\mathbf{x}) \cdot Warp \{ P_{I_2},\bm{F}^{1\rightarrow2} \} (\mathbf{x}+\bm{\delta}_{uv}) &\\
&=P_{I_1}(\mathbf{x}) \cdot Warp\{P_{I_2}(\mathbf{x}+\bm{\delta}_{uv}), \bm{F}^{1\rightarrow2}(\mathbf{x}+\bm{\delta}_{uv})\} &\\
&=P_{I_1}(\mathbf{x}) \cdot P_{I_2} (\mathbf{x}+\bm{F}^{1\rightarrow2} (\mathbf{x}+\bm{\delta}_{uv})+\bm{\delta}_{uv} ), &\\
\end{aligned}
\label{equ:corr}
\end{equation}
where $\cdot$ is the vector dot product, $\bm{F}^{1\rightarrow2}$ is the forward optical flow. $\bm{\delta}_{uv}=(\delta_u, \delta_v) \in ([-d_u,d_u], [-d_v,d_v])$ represents the horizontal and vertical search range, the values of $d_u,d_v$ are two manually set hyper-parameters determines the 2D range of built correlation.
The backward wrapping operation $Warp$ is implemented with bilinear interpolation and can compute the gradients for backpropagation~\cite{flow:Ilg_Flownet2_cvpr_2017}.

\subsubsection{Baselines}
\begin{figure}[tbp] 
\centering 

\subfigure[\textbf{Baseline-EV}]{
\begin{minipage}[t]{0.95\linewidth}
    \includegraphics[width=\textwidth]{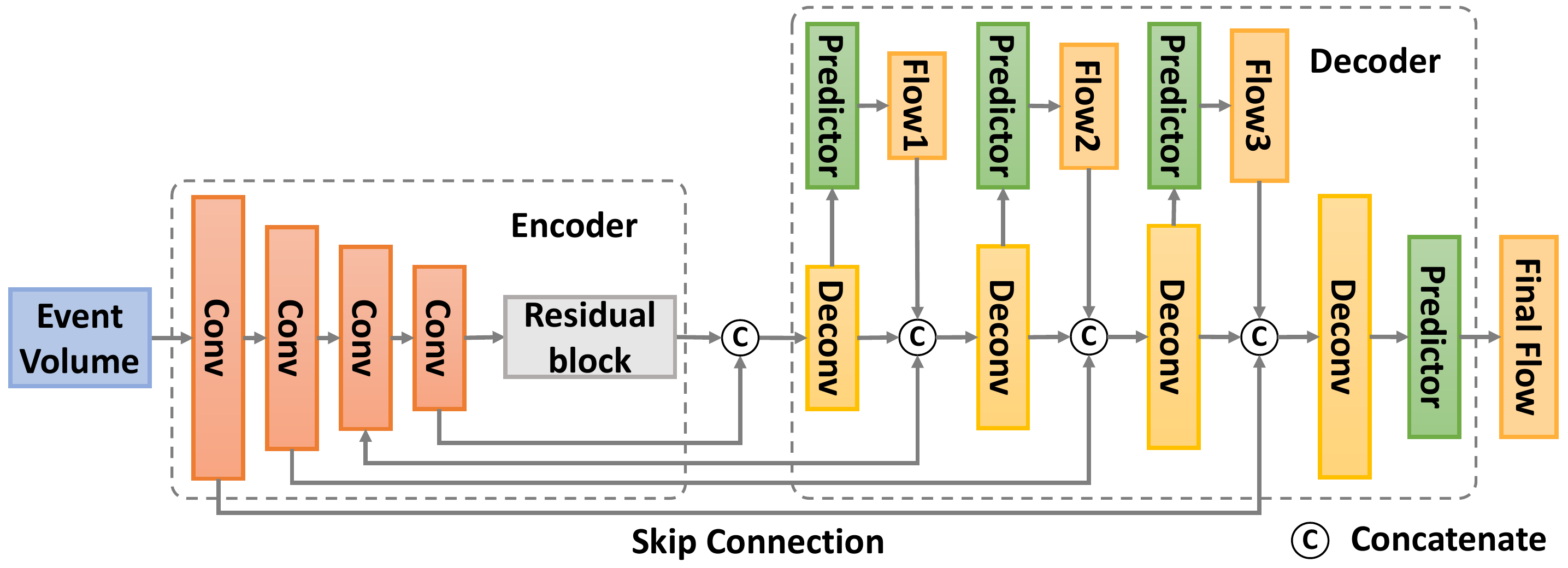}
    \label{fig:baeline_evflownet}
    \vspace{-10pt}
\end{minipage}%
}%

\vspace{-5pt}
\quad

\subfigure[\textbf{Baseline-EI}]{
\begin{minipage}[t]{0.95\linewidth}
    \includegraphics[width=\textwidth]{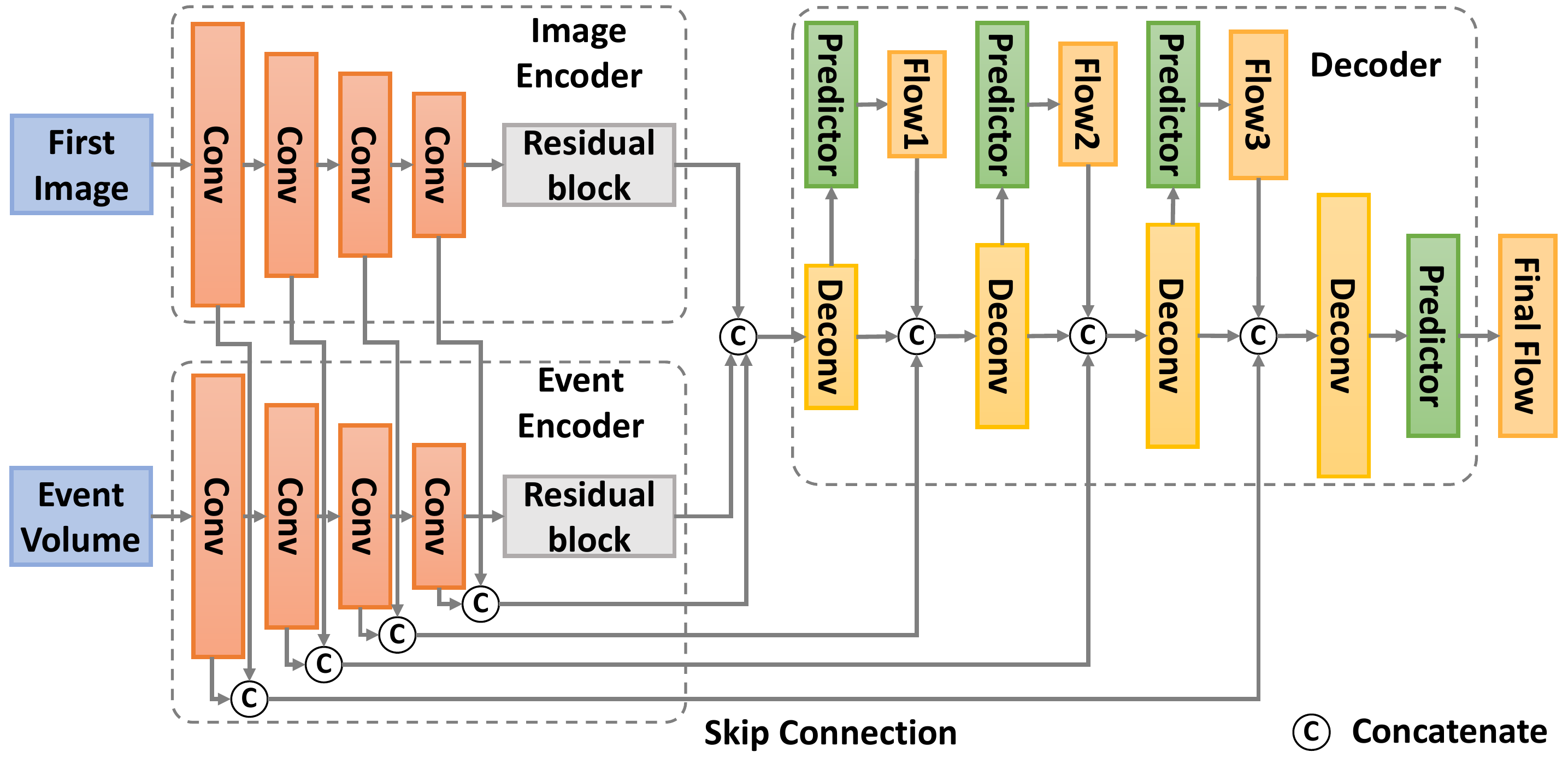} 
    \label{fig:baeline_evflownet_fusion}
    \vspace{-10pt}
\end{minipage}%
}%

\vspace{-5pt}
\caption{\textbf{The network structure of two baselines.} \textbf{\textit{Baseline-EV}} (a) is a U-Net-like network that only inputs the event volume. \textbf{\textit{Baseline-EI}} (b) adds an image encoder to get the image features and concatenates it with event features as the flow decoder input.}

\vspace{-10pt}
\label{fig:baselines}
\end{figure}

Most of the existing deep learning-based event-based flow methods employ a U-Net-like network, \eg, EV-FlowNet~\cite{eventflow:Zhu_EVFlowNet_CVPR_2019}. So we adopt it as a baseline network (called \textit{Baseline-EV} in Fig.~\ref{fig:baeline_evflownet}) using our event representation, training, and evaluation pipeline. \Fix{Inspired by FusionFlowNet~\cite{eventflow:Lee_Fusion_FlowNet_ICRA_2022}, we also introduce another simple network (called \textit{Baseline-EI} in Fig.~\ref{fig:baeline_evflownet_fusion}) modified upon EV-FlowNet to estimate flow from a single image and events.} We add another encoder to get the image feature and concatenate it with event features as the decoder input. The structures of these two baseline models are shown in Fig.~\ref{fig:baselines}.

\subsection{Event Representation}
Since the original event streams are composed of a series of discrete events, following the setting in~\cite{eventflow:Zhu_EVFlowNet_CVPR_2019, eventapp:Rebecq_Eventstovideo_TPAMI_2019, eventapp:Rebecq_HighSpeedHDREvent_TPAMI_2019, eventflow:Stoffregen_ReducingGAP_ECCV_2020}, we aggregate it into a three-dimensional event volume as the input of the convolution network. 
This process preserves most of the spatial-temporal information in the original event streams.

For an event stream $(e_i)^N, i \in [0, N]$ with $N$ events, we divide it into $B$ temporal bins as the channel dimension of an event volume for each polarity, then sum the normalized timestamps at different pixel positions in each bin as below:
\begin{equation}
E(b, \mathbf{x}_i, p_i) \!=\! \sum_{i=0}^{N}{ \!\max\!{ \left( 0,\! 1 \!-\! \left |b - \!\frac{t_i \!-\! t_{start}}{t_{end} \!-\!  t_{start}}\! (B \!-\! 1) \right| \right) }},
\end{equation}
where $b \in [0, B)$ indicates the index of temporal bins, $t_{start}$ and $t_{end}$ are the start and end timestamps of the event streams, respectively. Finally, we concatenate these temporal bins by two polarities to an event volume with $(2B \times H \times W)$:
\begin{equation}
\mathop{V(\mathbf{x})}\limits_{2B \, channels}=[\mathop{E(\mathbf{x},p=1)}\limits_{B \, channels}, \mathop{E(\mathbf{x},p=-1)}\limits_{B \, channels}],
\end{equation}
where $[~,~]$ is the concatenate operation and we perform it on the channel dimension. 

According to \cite{eventflow:Stoffregen_ReducingGAP_ECCV_2020, eventflow:Paredes_BackEventBasics_CVPR_2021, eventapp:Cadena_SPADE_E2VID_TIP_2021}, 
we divide the event stream into $B$=5 temporal bins for two polarities in our experiments, then the shape of represented event volume is $(10 \times H \times W)$.

\begin{figure*}[tbp] 
\centering 
\includegraphics[width=0.9\textwidth]{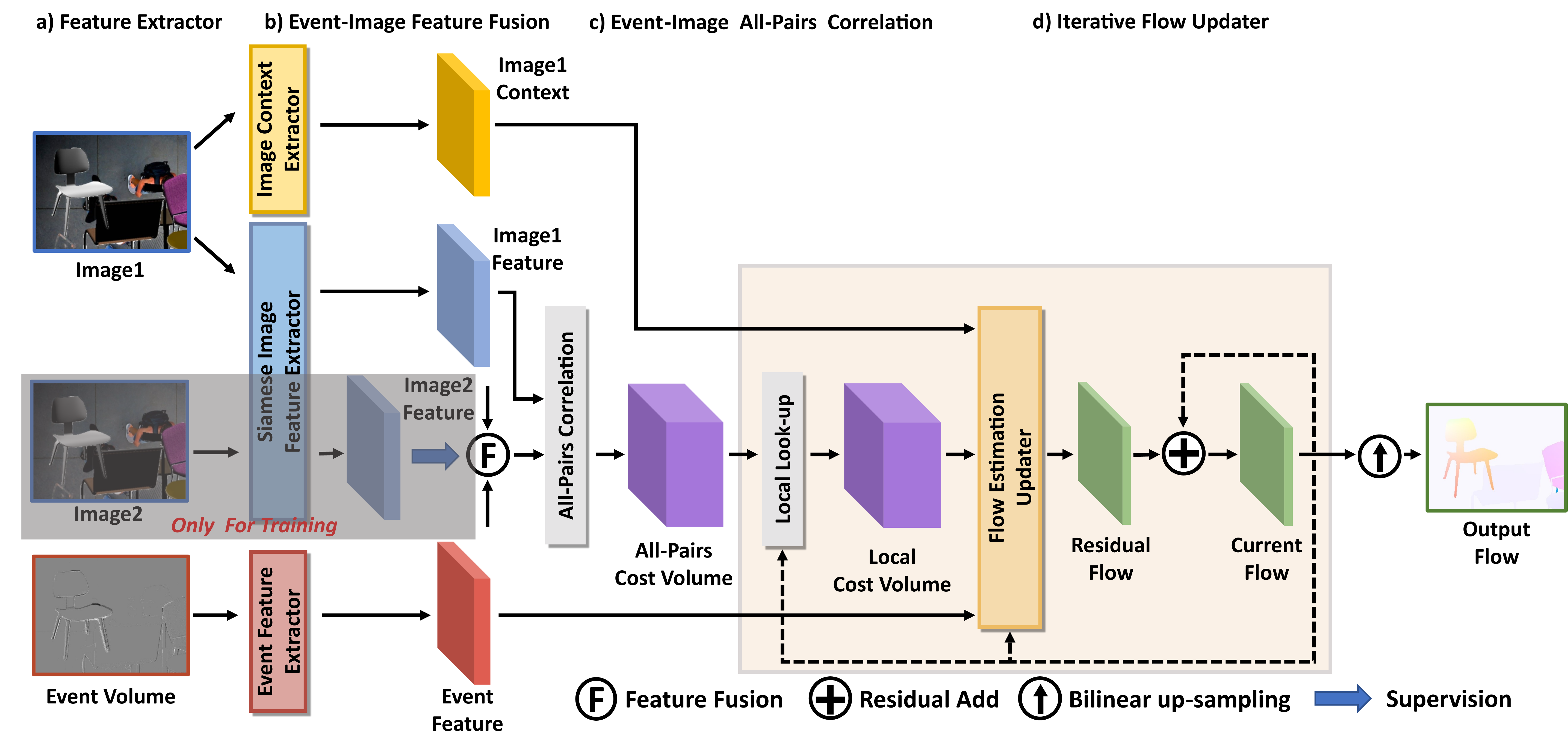}
\vspace{-5pt} 
\caption{\Fix{\textbf{Our DCEIFlow model structure.}
We use the feature extractor (left) to obtain event and image features and compute the matching correlation using our proposed event-image fusion and correlation construction module (middle). Then we feed them into the iterative flow updater (right) to update the flow iteratively. After the last iteration, we apply the up-sample operation to get the full-resolution output. The structures enclosed by the orange box need to be iteratively updated.}}
\label{fig:iterative_structure}
\vspace{-10pt}
\end{figure*}

\begin{figure}[tbp] 
\centering
\includegraphics[width=0.45\textwidth]{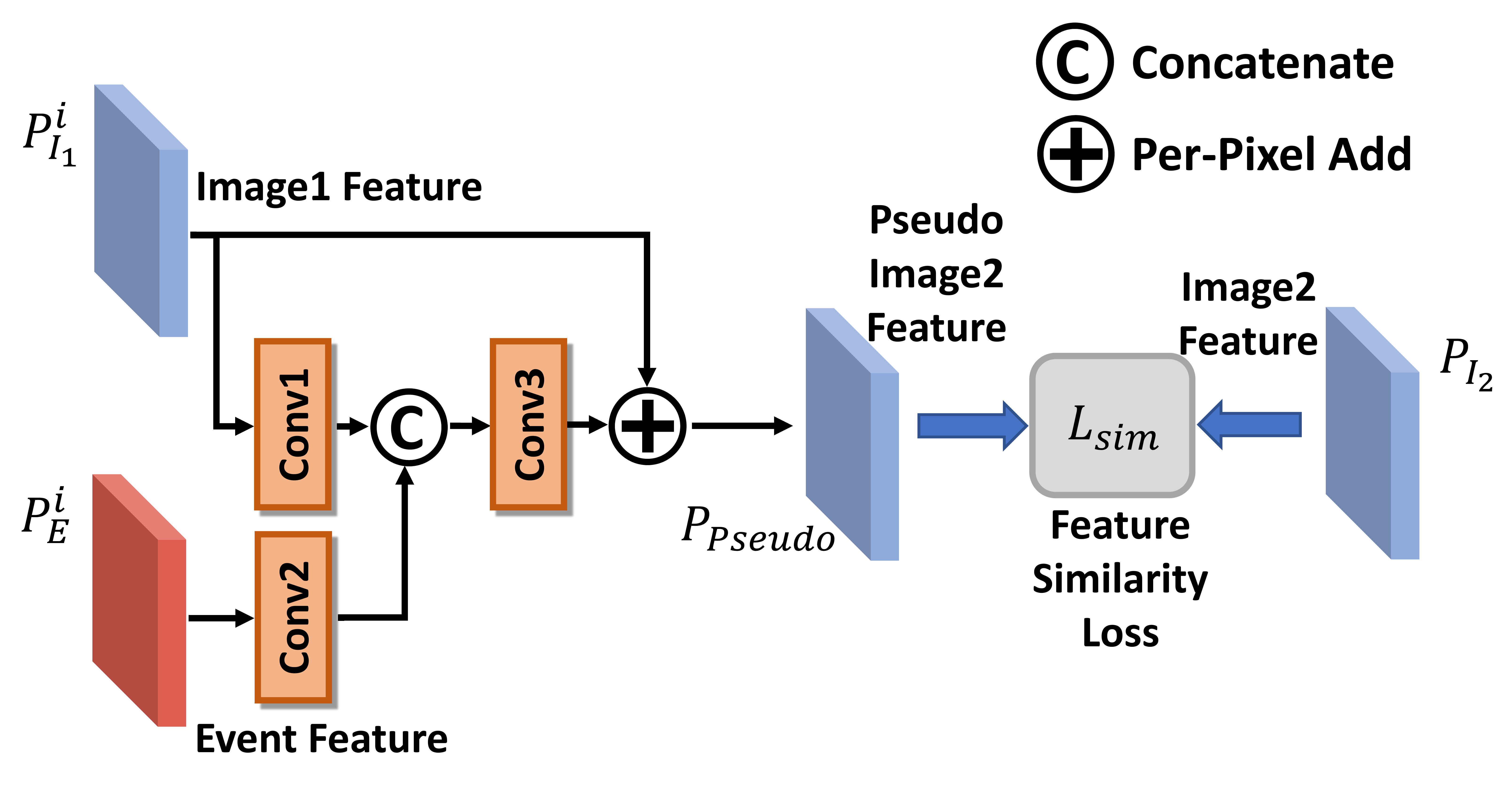}
\vspace{-5pt}
\caption{\textbf{The \textit{Fusion by Convolutions} structure of our Event-Image Fusion module}. The feature similarity loss $L_{sim}$ is only used in network training.}
\label{fig:feature_fusion_conv}
\vspace{-10pt}
\end{figure}

\subsection{Dense Iterative Event-Image Flow Network}

After representing the original discrete events as an event volume, we first extract the event and image feature using the feature extractor. Then we propose an event-image feature fusion module and construct the event-image correlation to effectively exploit the internal motion from two different modalities.
After that, we adopt the event-image correlation to the iterative flow update structure for accurate dense optical flow estimation.
Overall, our iterative network structure is shown in Fig.~\ref{fig:iterative_structure}, which consists of four modules: event and image feature extractor, event-image feature fusion, event-image all-pairs correlation module, and residual flow updater. 

\textbf{Event and Image Feature Extractor}.
We use two convolutional encoders with the same structure but without sharing weights to extract the image and event features.
Each encoder consists of 6 residual blocks with three times downsampling performed by convolution with a stride of 2. Then the encoders extract the input event volume size from ($H \times W \times 2B$) to feature size ($H/8 \times W/8 \times C$), and image size from ($H \times W \times 3$) to ($H/8 \times W/8 \times C$), where $C=256$ is the channel size of feature maps. In addition, the image encoder also extracts features from the input second frame image during training.

\textbf{Event-Image Feature Fusion}.
\label{section:featurefusion}
The correlation plays an important role in the two-frame optical flow estimation, thus we extend it to events. 
However, within our setting, the first image and event features are different data types, we cannot directly use the basic construction in Eq.~\ref{equ:corr}. 
Therefore, we propose to fuse the single image feature with the event feature to build the correlation, as shown in Fig.~\ref{fig:feature_fusion_conv}. We use the motion contained in the event feature to establish the conversion relationship with the first image, and generate pseudo second frame image features. Then we construct the correlation of these two image features as the input of the flow updater.

We first consider the simple addition operation, the \textit{Fusion by Add} structure. This method directly adds the event feature $P_E$ and the first image feature $P_{I_1}$ to obtain the pseudo second image feature $P_{pesudo}$.
\begin{equation}
\begin{aligned}
P_{pesudo} &= EIF_{add}( P_{I_1}, P_E) = P_{I_1} + P_E.
\end{aligned}
\end{equation}

Since the modality between the two features is different, direct addition is not an intuitive choice. Therefore, we propose \textit{Fusion by Convolutions} as shown in Fig.~\ref{fig:feature_fusion_conv}. There are three convolution layers in this module. The first two are used to encode the first image feature $P_{I_1}$ and event feature $P_E$, respectively. The last one is used for fusion by concatenating operation. The final pseudo second image feature $P_{pesudo}$ is obtained by residual addition.
\begin{equation}
\begin{aligned}
P_{pesudo} &= EIF_{conv}( P_{I_1}, P_E) \\
&=Conv3([Conv1(P_{I_1}), Conv2(P_E)]) + P_{I_1}.
\end{aligned}
\end{equation}

We propose a feature similarity loss $L_{sim}$ to supervise the similarity between the fused pseudo second image feature with the real second image feature.
We finally choose \textit{Fusion by Convolutions} for better performance. \textit{Fusion by Add} is used for ablation studies~\ref{section:ablation}.

\textbf{Event-Image All-Pairs Correlation Module}.
We construct the all-pairs correlation to enlarge the previous local path size described in~\ref{sec:localcorr} to full feature size. 
For the given first image feature $P_{I_1}$ and fused pseudo second image feature $P_{pseudo}$, the all-pair correlation $C^0_{EI}$ can be obtained by calculating the matrix multiplication:
\begin{equation}
\begin{aligned}
&C^0_{EI}(\mathbf{x}, \mathbf{y}) = P_{I_1}(\mathbf{x}, c) P_{pesudo}^T(\mathbf{y}, c),
\end{aligned}
\end{equation}
where $c$ is the channel of the feature map, $\mathbf{x}, \mathbf{y}$ are the coordinate vectors of the two features.

For the input feature map with size ($H \times W \times C$), the constructed correlation size is ($H \times W \times H \times W$).
We also introduce the pyramid correlation construction by three times average pooling to involve both large and small search ranges.
\begin{equation}
\begin{aligned}
&C^k_{EI}(\mathbf{x},\! \mathbf{y}') \!=\! \frac{1}{2^{2k}} \!\sum_{\mathbf{q}}^{(2^k,2^k)}{ C^0_{EI}\left(\mathbf{x}, 2^k \!\times\! \mathbf{y}' \!+\! \mathbf{q}\right) },
\end{aligned}
\end{equation}
where $\mathbf{y}'$ is the pooled coordinates of the last two dimensions, $k \in [1, 3]$ is the pyramid level, and $k$=0 means the original correlation. Thus the size of each correlation is ($H \times W \times H/2^k \times W/2^k$). 

Finally, we perform the lookup operation by the coarse flow in the defined search range $(\delta_u, \delta_v)$ on each correlation.
\begin{equation}
\begin{aligned}
&LC^k_{EI}(\mathbf{x}, \bm{\delta}_{uv}) = C^k_{EI}\left(\mathbf{x}, \frac{\mathbf{x}+\bm{F}^{1\rightarrow2}(\mathbf{x})}{2^{k}}+\bm{\delta}_{uv}\right).
\end{aligned}
\end{equation}

The size of $k$ level local correlation $LC^k_{EI}$ is ($H \times W \times (2 \times d_u + 1) \times (2 \times d_v + 1)$). We merge the last two dimensions and concatenate all of the pyramids $\{LC^0_{EI}, LC^1_{EI}, LC^2_{EI}, LC^3_{EI}\}$, then feed into the iterative flow updater.

\textbf{Iterative Flow Updater}.
The iterative flow updater consists of a ConvGRU (Convolutional Gated Recurrent Unit~\cite{Cho_GRU_EMNLP_2014}) and several convolution layers, which can estimate the residual flow $\Delta \bm{F}$ from the concatenation of image and event features, as well as the pyramid correlation volume. 
At each iteration, the residual flow $\Delta \bm{F}$ output by the flow updater is used to update the estimated flow $\bm{F}$.
The flow updater iterates $N$ times with the correlation lookup operation. \Fix{The $i$-th updated optical flow $\bm{F}^i$ is the sum of the previous and the current estimations: $\bm{F}^i = \bm{F}^{i-1} + \Delta \bm{F}$.}

Because the resolution of the extracted feature map is reduced to $1/8$ of the input image, the predicted optical flow needs to be upsampled to the input size. We use $8\times$ bilinear upsampling to the updated flow after the last iteration as the final predicted optical flow.

\begin{figure}[tbp] 
\centering 
\includegraphics[width=0.5\textwidth]{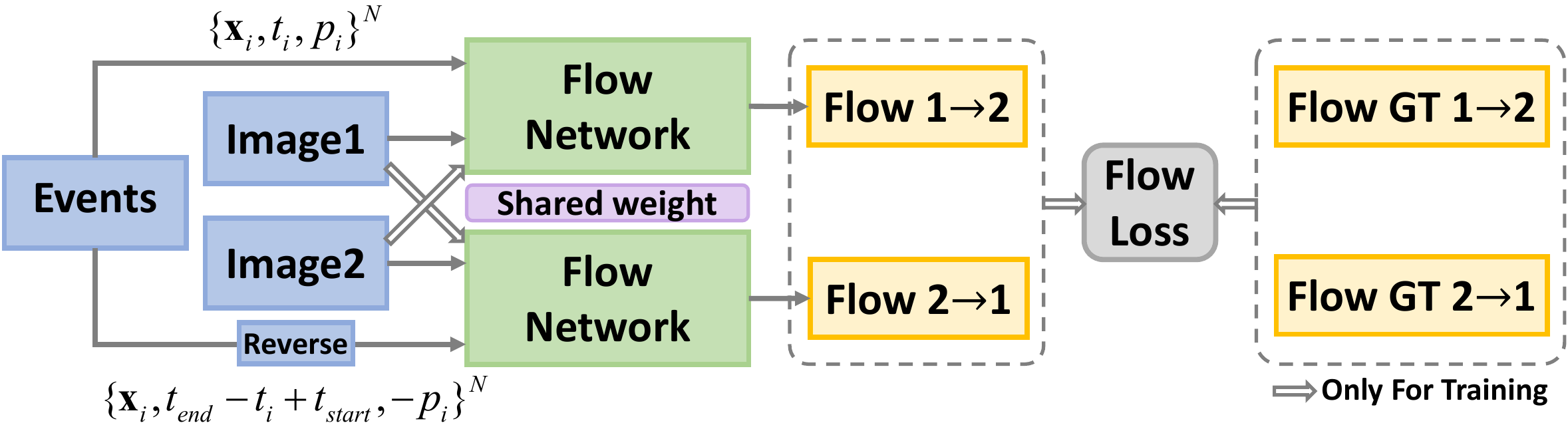} 
\caption{\textbf{Our bidirectional Event-Image flow training framework.} It takes the first image and original events as input and outputs the forward flow. Meanwhile, the second image and reversed events are taken as input to output the backward flow. The ground truth forward and backward flow are used as supervision for training.}
\label{fig:bidirection_training}
\vspace{-10pt}
\end{figure}

\subsection{Bidirectional Training}

Bidirectional flow training has been widely used in two-frame-based approaches~\cite{flow:Hur_IRR_CVPR_2019, flow:Liu_SelFlow_CVPR_2019}, which shows that the network with shared weight can estimate the forward and backward flow by exchanging the input order of two images.
For event-based flow estimation, we can know the order of each event from the event timestamps.
If we reverse the order of event timestamps and reverse the polarity of brightness changes, the motion becomes reversed. 
Thus the estimated flow can be considered equivalent to the inverse motion.

Therefore, we propose our bidirectional flow training mechanism in Fig.~\ref{fig:bidirection_training}.
If we train our model on the datasets with bidirectional flow annotation, we can input the original and reversed events and the corresponding image to the network, respectively, and use the ground-truths to supervise the output forward and backward flow. 
However, it should be noted that the backward flow requires the second image as input, so we cannot obtain the backward flow using the first image during the inference stage.
Our proposed bidirectional training mechanism is helpful to get more accurate flow results and improve the generalization ability without adding any network parameters, which has been verified in our ablation studies.

\subsection{Training Loss}

\subsubsection{Flow Loss}

Assume the ground truth flow is $\bm{F}^{gt}$, the predicted flow at the $i$-th iteration is $\bm{F}^i$, where $i=1,2,...,N$, $N$ is the total number of iterations. Then the predicted flows are supervised by using the following flow loss $L_{f}$:
\begin{equation}
L_{f} = \sum_{i = 1}^N{ {\phi}^{N-i+1} \rho(\parallel {\bm{F}^i-\bm{F}^{gt}} \parallel_2)},
\end{equation}
where the robust function $\rho(x) = (x^2+\epsilon)^q$, $q\in(0, 1)$ is less sensitive to outliers, $\epsilon$ is a small number which is close to 0. $\phi$ is a hyper-parameter used to balance the loss weights of each prediction. In our experiments, we set $N=6$ to balance the computation cost and performance and $\phi=0.8$ to make the later predictions with bigger weights.

For bidirectional training, we define the bidirectional flow loss $L_{fb}$ as:
\begin{equation}
\begin{aligned}
L_{fb} = \frac{1}{2} \cdot \sum_{i = 1}^N{{\phi}^{N-i+1}} \big[ & \rho(\parallel {\bm{F}^i_{1\rightarrow2} \!-\! \bm{F}^{gt}_{1\rightarrow2}} \parallel_2) \\ 
+ &\rho(\parallel {\bm{F}^i_{2\rightarrow1} \!-\! \bm{F}^{gt}_{2\rightarrow1}} \parallel_2) \big],
\end{aligned}
\end{equation}
where the ground truth forward and backward flow are $\bm{F}^{gt}_{1\rightarrow2}$ and $\bm{F}^{gt}_{2\rightarrow1}$, the predicted flow are $\bm{F}^i_{1\rightarrow2}$ and $\bm{F}^i_{2\rightarrow1}$.

\subsubsection{Feature similarity Loss}
\label{section:similarityloss}

In our event-image fusion module, the feature similarity loss $L_{sim}$ is used to supervise the pseudo second image feature $P_{pesudo}$ similar to the real second image feature $P_{I_2}$. Thus, we use $L_{sim}$ to compute $L_2$ distance between these two features:
\begin{equation}
L_{sim} = \parallel P_{I_2} - P_{pesudo} \parallel_2.
\end{equation}

\subsubsection{Total Training Loss}

The total training loss is a weighted sum of those two losses.
When training on the dataset with both forward and backward flow annotations, such as FlyingChairs2~\cite{flowdatasets:Ilg_FlyingChairs2_ECCV_2018}, the bidirectional loss $L_{bi}$ is used. 
When training on the dataset with the only forward flow, such as MVSEC~\cite{eventdatasets:Zhu_MVSEC_RAL_2018}, the unidirectional loss $L_{un}$ is used.

\begin{equation}
\begin{aligned}
L_{bi} &= L_{fb} + \lambda \cdot L_{sim}, \\
L_{un} &= L_{f} + \lambda \cdot L_{sim}.
\label{equ:loss}
\end{aligned}
\end{equation}

In our experiments, the feature similarity loss $L_{sim}$ can quickly converge to a small order of magnitude, so we set $\lambda = 100.0$ to balance the losses.

%%%%%%%%%%%%%%%%%%% Experiments %%%%%%%%%%%%%%%%%%%%%
\section{Experiments}

In this section, we first introduce our implementation details, including datasets, simulation, training details, and evaluation metrics. Then, we show the evaluation results of our model on both simulated and real datasets with comparisons to several baselines and existing methods. We further prove the effectiveness of each component in our network and the advantages of our network in dense and continuous optical flow estimation by model analysis. We conclude with discussions on failure cases and the limitations of our model.

\subsection{Implementation details}

\begin{figure*}[tbp]
\centering
\vspace{-0.2cm}

\begin{minipage}[t]{1\linewidth}
\footnotesize
\centering
    \begin{minipage}[t]{0.13\linewidth}
    \centering
    \includegraphics[width=\textwidth]{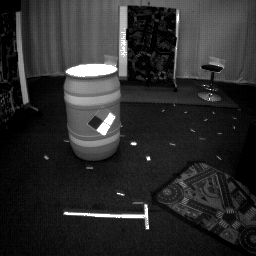} \\
    \vspace{0.02cm}
    \includegraphics[width=\textwidth]{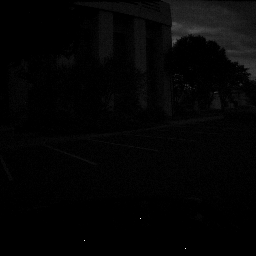} \\
    Image
    \end{minipage}%
    \quad
    \begin{minipage}[t]{0.13\linewidth}
    \centering
    \includegraphics[width=\textwidth]{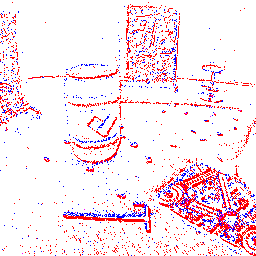} \\
    \vspace{0.02cm}
    \includegraphics[width=\textwidth]{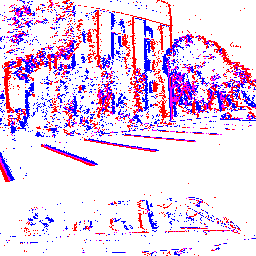} \\
    Events
    \end{minipage}%
    \quad
    \begin{minipage}[t]{0.13\linewidth}
    \centering
    \includegraphics[width=\textwidth]{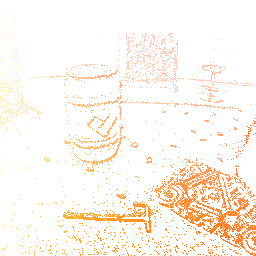} \\ 
    \vspace{0.02cm}
    \includegraphics[width=\textwidth]{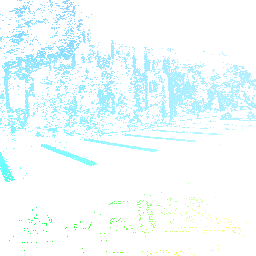} \\ 
    Flow GT~\textbf{M}
    \end{minipage}%
    \quad
    \begin{minipage}[t]{0.13\linewidth}
    \centering
    \includegraphics[width=\textwidth]{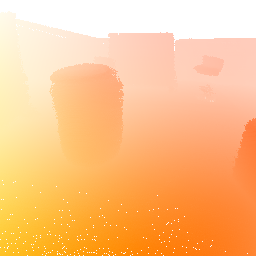} \\ 
    \vspace{0.02cm}
    \includegraphics[width=\textwidth]{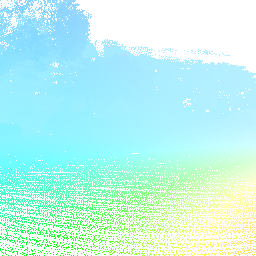} \\ 
    Flow GT
    \end{minipage}%
    \quad
    \begin{minipage}[t]{0.13\linewidth}
    \centering
    \includegraphics[width=\textwidth]{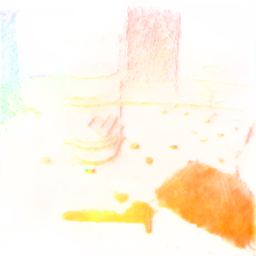} \\ 
    \vspace{0.02cm}
    \includegraphics[width=\textwidth]{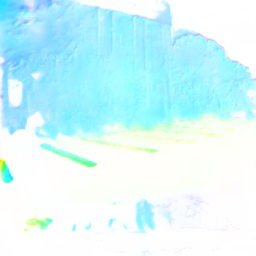} \\ 
    EV-FlowNet~\cite{eventflow:Zhu_EVFlowNet_RSS_2018}
    \end{minipage}%
    \quad
\\
    \begin{minipage}[t]{0.13\linewidth}
    \centering
    \includegraphics[width=\textwidth]{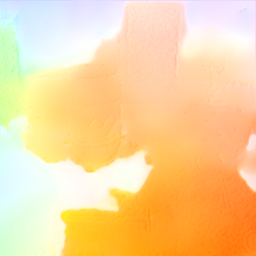} \\ 
    \vspace{0.02cm}
    \includegraphics[width=\textwidth]{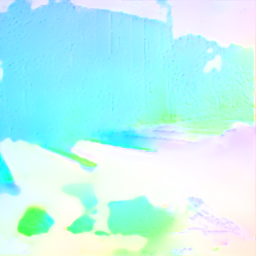} \\ 
    Stoffregen~\etal~\cite{eventflow:Stoffregen_ReducingGAP_ECCV_2020}
    \end{minipage}%
    \quad
    \begin{minipage}[t]{0.13\linewidth}
    \centering
    \includegraphics[width=\textwidth]{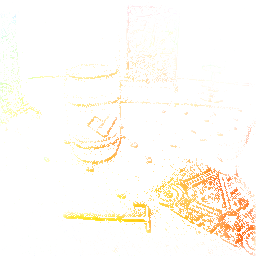} \\ 
    \vspace{0.02cm}
    \includegraphics[width=\textwidth]{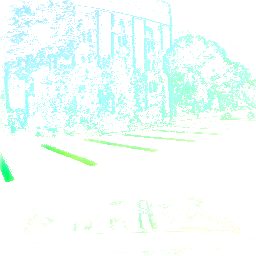} \\ 
    SpikeFlowNet~\textbf{M}~\cite{eventflow:Lee_SpikeFlowNet_ECCV_2020}
    \end{minipage}%
    \quad
    \begin{minipage}[t]{0.13\linewidth}
    \centering
    \includegraphics[width=\textwidth]{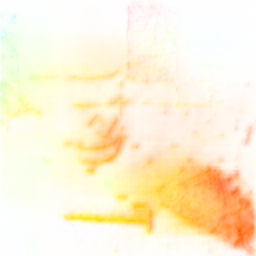} \\ 
    \vspace{0.02cm}
    \includegraphics[width=\textwidth]{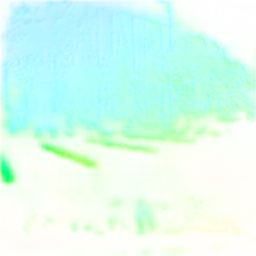} \\ 
    SpikeFlowNet~\cite{eventflow:Lee_SpikeFlowNet_ECCV_2020}
    \end{minipage}%
    \quad
    \begin{minipage}[t]{0.13\linewidth}
    \centering
    \includegraphics[width=\textwidth]{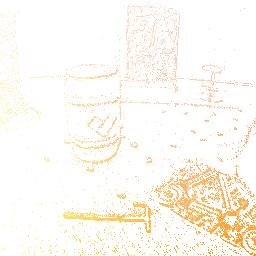} \\ 
    \vspace{0.02cm}
    \includegraphics[width=\textwidth]{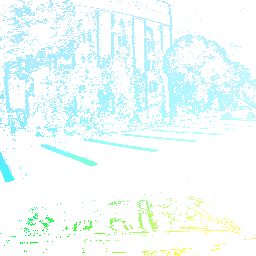} \\ 
    DCEIFlow~(Ours)~\textbf{M}
    \end{minipage}%
    \quad
    \begin{minipage}[t]{0.13\linewidth}
    \centering
    \includegraphics[width=\textwidth]{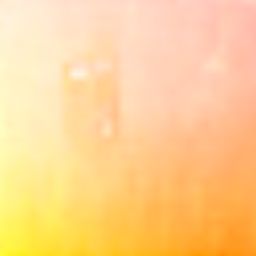} \\ 
    \vspace{0.02cm}
    \includegraphics[width=\textwidth]{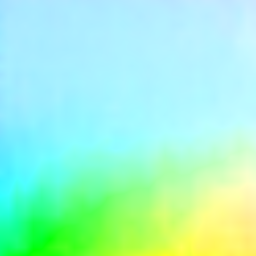} \\ 
    DCEIFlow~(Ours)
    \end{minipage}%
    \quad
\end{minipage}%
\centering
\vspace{-5pt}

\caption{\textbf{Visual comparisons on the MVSEC dataset~\cite{eventdatasets:Zhu_MVSEC_RAL_2018}.} \textbf{M} is the masked flow at the pixels with events. Our model gets better visual results in both indoor (top) and outdoor scenes (bottom).
Best viewed on screen.
\vspace{-10pt}
}
\label{viz:mvsec}
\end{figure*}

\begin{table*}[btp]
  \captionsetup{format=plain,labelformat=simple,justification=centering, labelsep=newline, singlelinecheck=false, textfont={sc}}
  \caption{\Fix{\textbf{Performance evaluation on the MVSEC dataset~\cite{eventdatasets:Zhu_MVSEC_RAL_2018} compared with existing event-based methods.}
  The results of the compared methods are directly extracted from the original papers.
  Note that existing event-based methods usually train two separate models for different time intervals ($dt\!=\!1$ and $dt\!=\!4$ frames), but our results are obtained on the same pre-trained model. }}
  \label{table:mvsec_frame_result}

  \resizebox{\textwidth}{!}{
  \Large
  \begin{threeparttable}
  \begin{tabular}{cc|ccc|c|cccccccccc}
    \toprule
    Input & Method & Train & Train & Train & Eval. & \multicolumn{2}{c}{\textit{indoor\_flying1}}& \multicolumn{2}{c}{\textit{indoor\_flying2}}& \multicolumn{2}{c}{\textit{indoor\_flying3}}& \multicolumn{2}{c}{\textit{outdoor\_day1}} & \multicolumn{2}{c}{\textit{outdoor\_day2}} \cr
    \textbf{$dt\!=\!1$} & Reference & Mann. & D.Type & D.Set & Metric & EPE & \%Out & EPE & \%Out & EPE & \%Out & EPE & \%Out & EPE & \%Out \cr
    \midrule
    \multirow{11}*{$\bf{E}$} &
    EV-FlowNet~\cite{eventflow:Zhu_EVFlowNet_RSS_2018} & USL & $\mathrm{I_1}$,$\mathrm{I_2}$,$\mathrm{E}$ & M & sparse & (1.03) & (2.2) & (1.72) & (15.1) & (1.53) & (11.9) & [0.49] & [0.2] & $\times$ & $\times$ \cr
    & Zhu~\etal~\cite{eventflow:Zhu_EVFlowNet_CVPR_2019} & USL & $\mathrm{E}$ & M & sparse & (0.58) & (0.0) & (1.02) & (4.0) & (0.87) & (3.0) & [0.32] & [0.0] & $\times$ & $\times$ \cr
    & EST~\cite{eventapp:Gehrig_event_representations_ICCV_2019} & SL & $\mathrm{E}$ & M & sparse & (0.97) & (0.91) & (1.38) & (8.20) & (1.43) & (6.47) & - & - & $\times$ & $\times$ \cr
    & Matrix-LSTM~\cite{eventflow:Cannici_Matrix-LSTM_ECCV_2020} & USL & $\mathrm{I_1}$,$\mathrm{I_2}$,$\mathrm{E}$ & M & sparse & (0.82) & (0.53) & (1.19) & (5.59) & (1.08) & (4.81) & - & - & $\times$ & $\times$ \cr
    & Spike-FlowNet\cite{eventflow:Lee_SpikeFlowNet_ECCV_2020} & USL & $\mathrm{I_1}$,$\mathrm{I_2}$,$\mathrm{E}$ & M & sparse & [0.84] & - & [1.28] & - & [1.11] & - & [0.49] & - & $\times$ & $\times$ \cr
    & Stoffregen~\etal~\cite{eventflow:Stoffregen_ReducingGAP_ECCV_2020} & SL & $\mathrm{E}$ & ESIM & dense & \textbf{0.56} & 1.00 & 0.66 & 1.00 & 0.59 & 1.00 & 0.68 & 0.99 & 0.82 & \textbf{0.96} \cr
    & Paredes~\etal~\cite{eventflow:Paredes_BackEventBasics_CVPR_2021} & USL & $\mathrm{E}$ & M & sparse & (0.79) & (1.2) & (1.40) & (10.9) & (1.18) & (7.4) & [0.92] & [5.4] & $\times$ & $\times$ \cr
    & LIF-EV-FlowNet~\cite{eventflow:Hagenaars_SelfSpikeFlow_Neurips_2021} & USL & $\mathrm{E}$ & FPV & sparse & 0.71 & 1.41 & 1.44 & 12.75 & 1.16 & 9.11 & \textbf{0.53} & 0.33 & - & - \cr
    & Deng~\etal~\cite{eventflow:Deng_DistillationLearning_TIP_2021} & USL & $\mathrm{I_1}$,$\mathrm{I_2}$,$\mathrm{E}$ & M & sparse & (0.89) & (0.66) & (1.31) & (6.44) & (1.13) & (3.53) & - & - & $\times$ & $\times$ \cr
    & Li~\etal~\cite{eventflow:Li_LightWeightEventFlow_ICPR_2021} & USL & $\mathrm{I_1}$,$\mathrm{I_2}$,$\mathrm{E}$ & M & sparse & (0.59) & (0.83) & (\textbf{0.64}) & (2.26) & - & - & \textbf{[0.31]} & [0.03] & $\times$ & $\times$ \cr
    & STE-FlowNet~\cite{eventflow:Ding_STEFlowNet_aaai_2022} & USL & $\mathrm{I_1}$,$\mathrm{I_2}$,$\mathrm{E}$ & M & sparse & [0.57] & [0.1] & [0.79] & [1.6] & [0.72] & [1.3] & [0.42] & \textbf{[0.0]} & $\times$ & $\times$ \cr
    \midrule
    \multirow{1}*{$\bf{I_1}$+$\bf{I_2}$}
    & Fusion-FlowNet~\cite{eventflow:Lee_Fusion_FlowNet_ICRA_2022} & USL & $\mathrm{I_1}$,$\mathrm{I_2}$,$\mathrm{E}$ & M & dense & (0.62) & - & (0.89) & - & (0.85) & - & [1.02] & - & $\times$ & $\times$ \cr
    \multirow{1}*{+$\bf{E}$}
    & Fusion-FlowNet~\cite{eventflow:Lee_Fusion_FlowNet_ICRA_2022} & USL & $\mathrm{I_1}$,$\mathrm{I_2}$,$\mathrm{E}$ & M & sparse & (0.56) & - & (0.95) & - & (0.76) & - & [0.59] & - & $\times$ & $\times$ \cr
    \midrule
    \multirow{3}*{$\bf{I_1}$+$\bf{E}$}
    & Pan~\etal~\cite{eventflow:Pan_SingleImageFlow_CVPR_2020}$^{*}$ & MB & - & - & M & 0.93 & 0.48 & 0.93 & 0.48 & 0.93 & 0.48 & 0.93 & 0.48 & - & - \cr
    & DCEIFlow~(Ours) & SL & $\mathrm{I_1}$,$\mathrm{I_2}$,$\mathrm{E}$ & C2 & dense & \textbf{0.56} & \textbf{0.28} & \textbf{0.64} & \textbf{0.16} & \textbf{0.57} & \textbf{0.12} & 0.91 & 0.71 & \textbf{0.79} & 2.59 \cr
    & DCEIFlow~(Ours) & SL & $\mathrm{I_1}$,$\mathrm{I_2}$,$\mathrm{E}$ & C2 & sparse & 0.57 & 0.30 & 0.70 & 0.30 & 0.58 & 0.15 & 0.74 & \textbf{0.29} & 0.82 & 2.34 \cr
    \bottomrule
  \end{tabular}
  \begin{tablenotes}
    \Large
    \item[] * Only the average EPE and outlier results of four sequences are given in Pan \etal\cite{eventflow:Pan_SingleImageFlow_CVPR_2020}.
  \end{tablenotes}
  \end{threeparttable}
  }

  \resizebox{\textwidth}{!}{
  \Large
  \begin{threeparttable}
  \begin{tabular}{cc|ccc|c|cccccccccc}
    \toprule
    Input & Method & Train & Train & Train & Eval. & \multicolumn{2}{c}{\textit{indoor\_flying1}}& \multicolumn{2}{c}{\textit{indoor\_flying2}}& \multicolumn{2}{c}{\textit{indoor\_flying3}}& \multicolumn{2}{c}{\textit{outdoor\_day1}} & \multicolumn{2}{c}{\textit{outdoor\_day2}} \cr
    \textbf{$dt\!=\!4$} & Reference & Mann. & D.Type & D.Set & Metric & EPE & \%Out & EPE & \%Out & EPE & \%Out & EPE & \%Out & EPE & \%Out \cr
    \midrule
    \multirow{6}*{$\bf{E}$} &
    EV-FlowNet~\cite{eventflow:Zhu_EVFlowNet_RSS_2018} & USL & $\mathrm{I_1}$,$\mathrm{I_2}$,$\mathrm{E}$ & M & sparse & (2.25) & (24.7) & (4.05) & (45.3) & (3.45) & (39.7) & [1.23] & [7.3]& $\times$ & $\times$ \cr
    & Zhu~\etal~\cite{eventflow:Zhu_EVFlowNet_CVPR_2019} & USL & $\mathrm{E}$ & M & sparse & (2.18) & (24.2) & (3.85) & (46.8) & (3.18) & (47.8) & [1.30] & [9.7] & $\times$ & $\times$ \cr
    & Spike-FlowNet\cite{eventflow:Lee_SpikeFlowNet_ECCV_2020} & USL & $\mathrm{I_1}$,$\mathrm{I_2}$,$\mathrm{E}$ & M & sparse & [2.24] & - & [3.83] & - & [3.18] & - & [1.09] & - & $\times$ & $\times$ \cr % [0.87]
    & LIF-EV-FlowNet~\cite{eventflow:Hagenaars_SelfSpikeFlow_Neurips_2021} & USL & $\mathrm{E}$ & FPV & sparse & 2.63 & 29.55 & 4.93 & 51.10 & 3.88 & 41.49 & 2.02 & 18.91 & - & - \cr
    & Li~\etal~\cite{eventflow:Li_LightWeightEventFlow_ICPR_2021} & USL & $\mathrm{I_1}$,$\mathrm{I_2}$,$\mathrm{E}$ & M & sparse & (2.08) & (26.4) & (3.76) & (43.2) & - & - & [1.24] & [8.16] & $\times$ & $\times$ \cr
    & STE-FlowNet~\cite{eventflow:Ding_STEFlowNet_aaai_2022} & USL & $\mathrm{I_1}$,$\mathrm{I_2}$,$\mathrm{E}$ & M & sparse & [1.77] & [14.7] & [2.52] & [26.1] & [2.23] & [22.1] & [0.99] & [3.9] & $\times$ & $\times$ \cr
    \midrule
    \multirow{1}*{$\bf{I_1}$+$\bf{I_2}$}
    & Fusion-FlowNet~\cite{eventflow:Lee_Fusion_FlowNet_ICRA_2022} & USL & $\mathrm{I_1}$,$\mathrm{I_2}$,$\mathrm{E}$ & M & dense & (1.81) & - & (2.90) & - & (2.46) & - & [3.06] & - & $\times$ & $\times$ \cr
    \multirow{1}*{+$\bf{E}$}
    & Fusion-FlowNet~\cite{eventflow:Lee_Fusion_FlowNet_ICRA_2022} & USL & $\mathrm{I_1}$,$\mathrm{I_2}$,$\mathrm{E}$ & M & sparse & (1.68) & - & (3.24) & - & (2.43) & - & [1.17] & - & $\times$ & $\times$ \cr
    \midrule
    \multirow{2}*{$\bf{I_1}$+$\bf{E}$}
    & DCEIFlow~(Ours) & SL & $\mathrm{I_1}$,$\mathrm{I_2}$,$\mathrm{E}$ & C2 & dense & \textbf{1.49} & \textbf{8.14} & \textbf{1.97} & \textbf{17.37} & \textbf{1.69} & \textbf{12.34} & 1.87 & 19.13 & 1.62 & 14.73 \cr
    & DCEIFlow~(Ours) & SL & $\mathrm{I_1}$,$\mathrm{I_2}$,$\mathrm{E}$ & C2 & sparse & 1.52 & 8.79 & 2.21 & 22.13 & 1.74 & 13.33 & \textbf{1.37} & \textbf{8.54} & \textbf{1.61} & \textbf{14.38} \cr
    \bottomrule
  \end{tabular}
  \begin{tablenotes}
    \Large
    \item[] - indicates these methods do not provide the corresponding results.
    \item[] $\times$ indicates these methods are trained on the \textit{ourdoor\_day2} sequence.
    \item[] The results with (~) are obtained by evaluating the model trained on both \textit{ourdoor\_day1} and \textit{ourdoor\_day2} sequences. 
    \item[] The results with [~] are obtained by evaluating the model trained on the \textit{ourdoor\_day2} sequence. 
    \item[] The results not enclosed by any brackets indicate that the model is not trained on any sequence of MVSEC.
  \end{tablenotes}
  \end{threeparttable}
  }

\end{table*}

\begin{table*}[t!]
  \captionsetup{format=plain,labelformat=simple,justification=centering, labelsep=newline, singlelinecheck=false, textfont={sc}}
  \caption{\Fix{\textbf{Extended evaluation results on the MVSEC dataset~\cite{eventdatasets:Zhu_MVSEC_RAL_2018} compared with existing state-of-the-art two frame-based methods and event-based method E-RAFT~\cite{eventflow:Gehrig_DenseRAFTFlow_3DV_2021}.}
  We evaluate the results of the compared methods on their open-source pre-trained models by using the same dataset splitting (following~\cite{eventflow:Stoffregen_ReducingGAP_ECCV_2020}). We use the same pre-trained model to get the results of each method under two input interval settings  ($dt\!=\!1$ and $dt\!=\!4$ frames). }}
  \label{table:mvsec_more_result}

  \resizebox{\textwidth}{!}{
  \Large
  \begin{threeparttable}
  \begin{tabular}{cc|ccc|c|cccccccccc}
    \toprule
    Input & Method & Train & Train & Train & Eval. & \multicolumn{2}{c}{\textit{indoor\_flying1}}& \multicolumn{2}{c}{\textit{indoor\_flying2}}& \multicolumn{2}{c}{\textit{indoor\_flying3}}& \multicolumn{2}{c}{\textit{outdoor\_day1}} & \multicolumn{2}{c}{\textit{outdoor\_day2}} \cr
    \textbf{$dt\!=\!1$} & Reference & Mann. & D.Type & D.Set & Metric & EPE & \%Out & EPE & \%Out & EPE & \%Out & EPE & \%Out & EPE & \%Out \cr
    \midrule
    \multirow{6}*{$\bf{I_1}$+$\bf{I_2}$} &
    PWC-Net~\cite{flow:Sun_PWCNet_TPAMI_2019} & SL & $\mathrm{I_1}$,$\mathrm{I_2}$ & C+T & dense & 1.57 & 3.11 & 1.62 & 3.29 & 1.55 & 2.70 & 1.83 & 11.50 & 1.67 & 7.88 \cr
    & PWC-Net~\cite{flow:Sun_PWCNet_TPAMI_2019} & SL & $\mathrm{I_1}$,$\mathrm{I_2}$ & C+T & sparse & 1.59 & 3.27 & 1.69 & 4.80 & 1.58 & 3.10 & 1.75 & 7.68 & 1.64 & 8.39 \cr
    \multirow{6}*{As ref-}& RAFT~\cite{flow:Teed_RAFT_ECCV_2020} & SL & $\mathrm{I_1}$,$\mathrm{I_2}$ & C & dense & 0.44 & 0.13 & 0.54 & 0.05  & 0.50 & 0.00 & 0.86 & 0.21 & 0.62 & 2.81 \cr
    \multirow{6}*{erence} & RAFT~\cite{flow:Teed_RAFT_ECCV_2020} & SL & $\mathrm{I_1}$,$\mathrm{I_2}$ & C & sparse & 0.48 & 0.12 & 0.62 & 0.07 & 0.54 & 0.00 & 0.77 & 0.21 & 0.56 & 2.37 \cr
    & ARFlow~\cite{flow:liu_ARFlow_cvpr_2020} & USL & $\mathrm{I_1}$,$\mathrm{I_2}$ & SR+S & dense & 0.39 & 0.13 & 0.46 & 0.07 & 0.43 & 0.02 & 1.44 & 12.29 & 0.86 & 7.19 \cr
    & ARFlow~\cite{flow:liu_ARFlow_cvpr_2020} & USL & $\mathrm{I_1}$,$\mathrm{I_2}$ & SR+S & sparse & 0.38 & 0.11 & 0.48 & 0.05 & 0.41 & 0.00 & 0.89 & 5.44 & 0.70 & 4.49 \cr
    & SMURF~\cite{flow:stone_smurf_cvpr_2021} & USL & $\mathrm{I_1}$,$\mathrm{I_2}$ & C & dense & 0.42 & 0.14 & 0.50 & 0.27 & 0.46 & 0.15 & 1.50 & 12.90 & 0.95 & 8.79 \cr
    & SMURF~\cite{flow:stone_smurf_cvpr_2021} & USL & $\mathrm{I_1}$,$\mathrm{I_2}$ & C & sparse & 0.39 & 0.11 & 0.50 & 0.09 & 0.43 & 0.04 & 0.99 & 6.37 & 0.72 & 5.09 \cr
    \midrule
    \multirow{2}*{$\bf{E}$}
    & E-RAFT~\cite{eventflow:Gehrig_DenseRAFTFlow_3DV_2021} & SL & $\mathrm{E}$ & DSEC & dense & 0.70 & \textbf{0.16} & 0.94 & 2.97 & 0.82 & 1.48 & 0.95 & 4.55 & 1.04 & 6.47 \cr
    & E-RAFT~\cite{eventflow:Gehrig_DenseRAFTFlow_3DV_2021} & SL & $\mathrm{E}$ & DSEC & sparse & 0.78 & 0.33 & 1.20 & 5.70 & 0.93 & 2.25 & \textbf{0.65} & 2.19 & 0.92 & 4.73 \cr
    \midrule
    \multirow{2}*{$\bf{I_1}$+$\bf{E}$}
    & DCEIFlow~(Ours) & SL & $\mathrm{I_1}$,$\mathrm{I_2}$,$\mathrm{E}$ & C2 & dense & \textbf{0.56} & 0.28 & \textbf{0.64} & \textbf{0.16} & \textbf{0.57} & \textbf{0.12} & 0.91 & 0.71 & \textbf{0.79} & 2.59 \cr
    & DCEIFlow~(Ours) & SL & $\mathrm{I_1}$,$\mathrm{I_2}$,$\mathrm{E}$ & C2 & sparse & 0.57 & 0.30 & 0.70 & 0.30 & 0.58 & 0.15 & 0.74 & \textbf{0.29} & 0.82 & \textbf{2.34} \cr
    \bottomrule
  \end{tabular}
  \end{threeparttable}
  }

  \resizebox{\textwidth}{!}{
  \Large
  \begin{threeparttable}
  \begin{tabular}{cc|ccc|c|cccccccccc}
    \toprule
    Input & Method & Train & Train & Train & Eval. & \multicolumn{2}{c}{\textit{indoor\_flying1}}& \multicolumn{2}{c}{\textit{indoor\_flying2}}& \multicolumn{2}{c}{\textit{indoor\_flying3}}& \multicolumn{2}{c}{\textit{outdoor\_day1}} & \multicolumn{2}{c}{\textit{outdoor\_day2}} \cr
    \textbf{$dt\!=\!4$} & Reference & Mann. & D.Type & D.Set & Metric & EPE & \%Out & EPE & \%Out & EPE & \%Out & EPE & \%Out & EPE & \%Out \cr
    \midrule
    \multirow{6}*{$\bf{I_1}$+$\bf{I_2}$} &
    PWC-Net~\cite{flow:Sun_PWCNet_TPAMI_2019} & SL & $\mathrm{I_1}$,$\mathrm{I_2}$ & C+T & dense & 1.94 & 14.35 & 2.19 & 21.01 & 2.03 & 17.06 & 3.03 & 37.99 & 2.33 & 19.52 \cr
    & PWC-Net~\cite{flow:Sun_PWCNet_TPAMI_2019} & SL & $\mathrm{I_1}$,$\mathrm{I_2}$ & C+T & sparse & 1.96 & 14.95 & 2.31 & 24.60 & 2.05 & 17.46 & 2.48 & 26.62 & 2.28 & 19.44 \cr
    \multirow{6}*{As ref-}& RAFT~\cite{flow:Teed_RAFT_ECCV_2020} & SL & $\mathrm{I_1}$,$\mathrm{I_2}$ & C & dense & 1.45 & 7.85 & 1.80 & 13.89 & 1.65 & 11.02 & 3.10 & 38.77 & 1.43 & 12.50 \cr
    \multirow{6}*{erence} & RAFT~\cite{flow:Teed_RAFT_ECCV_2020} & SL & $\mathrm{I_1}$,$\mathrm{I_2}$ & C & sparse & 1.48 & 7.82 & 1.91 & 15.94 & 1.67 & 11.29 & 2.47 & 27.15 & 1.39 & 11.65 \cr
    & ARFlow~\cite{flow:liu_ARFlow_cvpr_2020} & USL & $\mathrm{I_1}$,$\mathrm{I_2}$ & SR+S & dense & 1.31 & 6.21 & 1.58 & 9.51 & 1.44 & 8.05 & 3.43 & 39.55 & 1.53 & 13.06 \cr
    & ARFlow~\cite{flow:liu_ARFlow_cvpr_2020} & USL & $\mathrm{I_1}$,$\mathrm{I_2}$ & SR+S & sparse & 1.31 & 6.59 & 1.72 & 12.05 & 1.47 & 8.81 & 1.89 & 17.36 & 1.42 & 11.50 \cr
    & SMURF~\cite{flow:stone_smurf_cvpr_2021} & USL & $\mathrm{I_1}$,$\mathrm{I_2}$ & C & dense & 1.34 & 6.80 & 1.63 & 10.38 & 1.49 & 8.76 & 3.98 & 46.49 & 1.73 & 15.49 \cr
    & SMURF~\cite{flow:stone_smurf_cvpr_2021} & USL & $\mathrm{I_1}$,$\mathrm{I_2}$ & C & sparse & 1.32 & 6.76 & 1.73 & 12.27 & 1.50 & 9.17 & 2.53 & 25.65 & 1.41 & 11.57 \cr
    \midrule
    \multirow{2}*{$\bf{E}$}
    & E-RAFT~\cite{eventflow:Gehrig_DenseRAFTFlow_3DV_2021} & SL & $\mathrm{E}$ & DSEC & dense & 1.82 & 15.58 & 2.64 & 25.47 & 2.12 & 17.60 & 1.93 & 19.55 & 1.66 & 14.05 \cr
    & E-RAFT~\cite{eventflow:Gehrig_DenseRAFTFlow_3DV_2021} & SL & $\mathrm{E}$ & DSEC & sparse & 1.89 & 16.41 & 3.22 & 33.23 & 2.27 & 19.81 & 1.43 & 9.17 & \textbf{1.59} & \textbf{11.83} \cr
    \midrule
    \multirow{2}*{$\bf{I_1}$+$\bf{E}$}
    & DCEIFlow~(Ours) & SL & $\mathrm{I_1}$,$\mathrm{I_2}$,$\mathrm{E}$ & C2 & dense & \textbf{1.49} & \textbf{8.14} & \textbf{1.97} & \textbf{17.37} & \textbf{1.69} & \textbf{12.34} & 1.87 & 19.13 & 1.62 & 14.73 \cr
    & DCEIFlow~(Ours) & SL & $\mathrm{I_1}$,$\mathrm{I_2}$,$\mathrm{E}$ & C2 & sparse & 1.52 & 8.79 & 2.21 & 22.13 & 1.74 & 13.33 & \textbf{1.37} & \textbf{8.54} & 1.61 & 14.38 \cr
    \bottomrule
  \end{tabular}
  \end{threeparttable}
  }
\end{table*}

\subsubsection{Datasets}\ 

- \textit{Selection}: The commonly used event camera optical flow dataset is MVSEC~\cite{eventdatasets:Zhu_MVSEC_RAL_2018}. 
However, only using it to train our network is not a good practice because it only has sparse flow annotation with low spatial resolution.
Most existing learning-based two-frame flow approaches usually pre-train on synthetic datasets and then finetune on other datasets to get benchmark results. 
Thus we use ESIM~\cite{eventdatasets:Rebecq_ESIM_CoRL_2018} to simulate the event data between two frames on the FlyingChairs2 dataset~\cite{flowdatasets:Ilg_FlyingChairs2_ECCV_2018} to pre-train our model, because it provides full ground truth annotation of forward and backward flow. 

The flow annotations of MVSEC are computed from the depth by LIDAR with the ego-motion by IMU, and there are only rigid scenes.
Therefore, to verify the performance of our model in non-rigid dynamic scenes, we use an event-based highly dynamic moving object segmentation dataset, EV-IMO~\cite{eventdatasets:mitrokhin_EVIMO_IROS_2019}. 
In addition, we also use the Sintel dataset~\cite{flowdatasets:Butler_Sintel_ECCV_2012} because it is commonly used in the two-frame methods.

- \textit{Details}: 
Following the split of Stoffregen~\etal~\cite{eventflow:Stoffregen_ReducingGAP_ECCV_2020}, each indoor and outdoor sequence in MVSEC~\cite{eventdatasets:Zhu_MVSEC_RAL_2018} contains 1,880$\sim$ and 2,700$\sim$ images with corresponding events, respectively. FlyingChairs2~\cite{flowdatasets:Ilg_FlyingChairs2_ECCV_2018} contains 22,232 training and 640 validation samples. Each sample includes two image pairs, forward \& backward flow annotations, and our simulated event data. Sintel~\cite{flowdatasets:Butler_Sintel_ECCV_2012} provides naturalistic movie sequences with challenging long-range non-rigid motion, which includes Clean and Final passes with 1,041 pairs of training sets.
Because it has no event data, and the flow annotations of the test set are not publicly available, we only simulate the event data on the training set to evaluate the generalization ability. 
The test set of EV-IMO~\cite{eventdatasets:mitrokhin_EVIMO_IROS_2019} dataset includes 21 sequences, with a total of 8258 pairs of data captured by the DAVIS346C camera. Due to the lack of optical flow annotation, we only use its image and event data to estimate optical flow for qualitative visual analysis.

\subsubsection{Event data simulation}

Due to the lack of an event-based dataset with high-quality optical flow annotations for training, we use the open-source ESIM simulator~\cite{eventdatasets:Rebecq_ESIM_CoRL_2018} to simulate events on FlyingChairs2.
To simulate realistic events, ESIM requires a small displacement of the corresponding pixels between two frames. However, the pixel displacement of this dataset is not guaranteed to be always small, so we cannot directly input it into the simulator with the original two frames.
Following Gehrig~\etal~\cite{eventdatasets:Gehrig_VideoToEvent_CVPR_2020}, we first use  Super-SloMo~\cite{Jiang_SuperSlomo_CVPR2018} to interpolate the two frames to more, and then use ESIM to simulate events on them. 
The amount of interpolating frames depends on the motion range between two frames.

\subsubsection{Model training details}
\label{section:training_details}

\Fix{We train our model on the FlyingChairs2 training set by $L_{bi}$ in Eq.~\eqref{equ:loss} for 100 epochs with a random cropped size $[368, 496]$ and a batch size of 8. Our model needs 25 minutes for one epoch, and it takes about 42 hours to complete the whole training process for 100 epochs. 
We train our model on two NVIDIA 2080Ti GPUs using PyTorch~\cite{Paszke_Pytorch_nips_2019}. We use the AdamW optimizer~\cite{Kingma_Adam_ICLR_2015} for training with a weight decay of $10^{-4}$ and default parameters $\beta_1=0.9$, $\beta_2=0.999$, $\epsilon=10^{-8}$. We use several geometric and photometric augmentations, including random resize and crop, horizontal and vertical flips, contrast and brightness changes, etc. We use the OneCycle~\cite{Smith_OneCycleLR_2019} policy and set the maximal learning rate to $4\times10^{-4}$. After model training, we directly use the same pre-trained model to evaluate on MVSEC under different time intervals ($dt\!=\!1$ and $dt\!=\!4$). }

\Fix{In the baseline comparison, we also conduct the experiments of training only on the \textit{outdoor\_day2} sequence of the MVSEC dataset~\cite{eventdatasets:Zhu_MVSEC_RAL_2018} (\ien, M). Each baseline and our model follow the same pre-training settings on the MVSEC dataset (Train D.Set M). Both are trained for 200 epochs with a batch size of 16. Our model needs about 2 minutes for one epoch, and it takes about 10 hours to complete the training process for 300 epochs. 
Then we use the same pre-trained model to obtain the results of each method under two input interval settings ($dt\!=\!1$ and $dt\!=\!4$ frames).}

\subsubsection{Evaluation metrics}
A commonly used metric for optical flow evaluation is the average End Point Error (\textbf{EPE}), which calculates the Euclidean distance between the predicted flow and the ground truth.
\begin{equation}
\begin{aligned}
EPE \!=\! \frac{1}{m} \cdot \sum_{m}{\sqrt{(\bm{F}^{pred}_x \!-\! \bm{F}^{gt}_x)^2 + (\bm{F}^{pred}_y \!-\! \bm{F}^{gt}_y)^2} }.
\end{aligned}
\end{equation}

For \textit{dense} evaluation, $m$ is the pixels with valid flow annotation. 
For \textit{sparse} evaluation, $m$ is the pixels with valid flow annotation and triggered at least one event.
$\bm{F}^{pred}$ is the predicted flow vector and $\bm{F}^{gt}$ is ground-truth flow vector, the $x$ and $y$ subscripts indicate horizontal and vertical directions. Following KITTI~\cite{flowdatasets:Menze_jointkitti2015_2015} and EV-FlowNet~\cite{eventflow:Zhu_EVFlowNet_RSS_2018}, we also use the \textbf{outlier} metric to report the percentage of points with endpoint error greater than 3 pixels and 5\% of the magnitude.
Both of these two metrics are smaller the better. 

\Fix{In addition, we introduce a metric called ${\bf Dense~Ratio}$ to measure the output density of an event-based optical flow estimation model, which is defined as follows,
\begin{equation}
  {\bf Dense~Ratio} = \frac{{\bf EPE}_{Dense} + {\bf EPE}_{Event~Masked}}{{\bf EPE}_{Dense} + {\bf EPE}_{Event~Excluded}},
\end{equation}
where ${\bf EPE}_{Dense}$ calculates the dense error of pixels annotated by the valid optical flow, ${\bf EPE}_{Event~Masked}$ calculates the sparse error of pixels with the valid flow and triggered at least one event, and ${\bf EPE}_{Event~Excluded}$ calculates the sparse error of pixels which do not trigger any event. Because events are usually triggered at moving objects or texture edges, we measure the dense flow prediction ability by calculating the ratio of the masked and excluded EPE. When the ratio is less than 1, the model has a smaller error at the edges than at other locations. The underlined results in Table~\ref{table:densesparse} are obtained using unsupervised pre-trained models. Since the unsupervised objective function usually has higher energy at the edges, this result illustrates that unsupervised training tends to make the model fit to the edges. }

\begin{table*}[tbp]
  \captionsetup{format=plain,labelformat=simple,justification=centering, labelsep=newline, singlelinecheck=false, textfont={sc}}
  \caption{\Fix{\textbf{Ablation study results.} Both models are pre-trained on the FlyingChairs2~\cite{flowdatasets:Ilg_FlyingChairs2_ECCV_2018} training set with the same training setting, and directly evaluated on the FlyingChairs2 validation set, Sintel~\cite{flowdatasets:Butler_Sintel_ECCV_2012} training set and MVSEC~\cite{eventdatasets:Zhu_MVSEC_RAL_2018} \textit{indoor\_flying1-3} sequences with $dt=1$.}}
  \label{table:ablation}
  \centering
  \resizebox{\textwidth}{!}{
  \begin{threeparttable}
  \begin{tabular}{c|cccccc|c|cc|cc|cc}
    \toprule
    Model & Event & Corr. & E-I & Sim & Bi. & Network & Param. & \multicolumn{2}{c|}{FlyingChairs2} & \multicolumn{2}{c|}{Sintel} & \multicolumn{2}{c}{MVSEC} \cr
    ID & Pol. & Module & Fusion & Loss & Train. & Structure & Num.~(M) & EPE & \%Out & EPE & \%Out & EPE & \%Out \cr
    \midrule
    (a) & $\times$ & $\times$ & $\times$ & $\times$ & $\times$ & Pyramid & 10.88 & 2.01 & 12.00 & 8.96 & 45.33 & 1.16 & 2.58 \cr 
    (b) & \checkmark & $\times$ & $\times$ & $\times$ & $\times$ & Pyramid & 10.88 & 1.97 & 11.71 & 8.52 & 42.41 & 1.05 & 2.64 \cr 
    (c) & \checkmark & \checkmark & Add & \checkmark & $\times$ & Pyramid & 12.28 & 1.90 & 11.36 & 8.12 & 38.41 & 0.88 & 1.49 \cr 
    (d) & \checkmark & \checkmark & Add & \checkmark & $\times$ & Iterative & 6.08 & 1.85 & 10.51 & 7.69 & 36.42 & 0.76 & 0.72 \cr 
    (e) & \checkmark & \checkmark & Conv & \checkmark & $\times$ & Pyramid & 13.27 & 1.83 & 10.79 & 7.55 & 36.01 & 0.74 & 0.93 \cr 
    (f) & \checkmark & \checkmark & Conv & \checkmark & $\times$ & Iterative & 7.07 & 1.66 & 8.92 & 7.01 & 34.51 & 0.62 & 0.40 \cr 
    (g) & \checkmark & \checkmark & Conv & \checkmark & \checkmark & Pyramid & 13.27 & 1.80 & 10.23 & 7.21 & 35.29 & 0.68 & 0.81 \cr 
    \midrule
    (h) & \checkmark & \checkmark & Conv & $\times$ & \checkmark & Iterative & 7.07 & 1.74 & 9.20 & 7.50 & 35.46 & 0.73 & 0.89 \cr 
    (i) & \checkmark & \checkmark & Conv & \checkmark & \checkmark & Iterative & 7.07 & \textbf{1.58} & \textbf{7.88} & \textbf{6.47} & \textbf{32.23} & \textbf{0.59} & \textbf{0.18} \cr 
    \bottomrule
  \end{tabular}
  \end{threeparttable}
  }
\end{table*}

\begin{figure*}[htbp]
\centering

\begin{minipage}[t]{1.0\linewidth}
\scriptsize
\centering

\begin{minipage}[t]{0.11\linewidth}
    \centering
    \includegraphics[width=\textwidth]{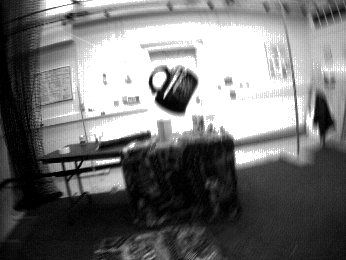} \\
    \vspace{0.03\linewidth}
    \includegraphics[width=\textwidth]{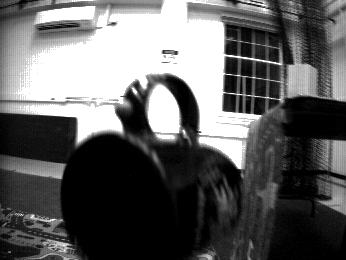} \\
    \vspace{0.03\linewidth}
    \includegraphics[width=\textwidth]{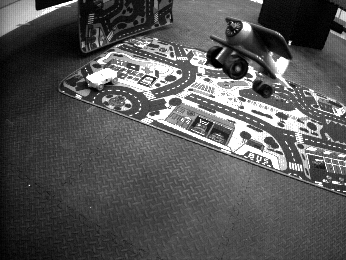} \\
    Image1
\end{minipage}%
\hspace{0.001\linewidth}
\begin{minipage}[t]{0.11\linewidth}
    \centering
    \includegraphics[width=\textwidth]{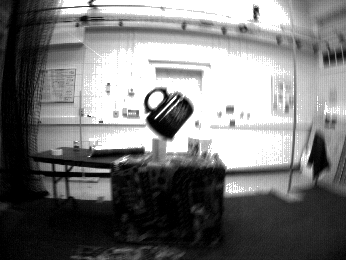} \\
    \vspace{0.03\linewidth}
    \includegraphics[width=\textwidth]{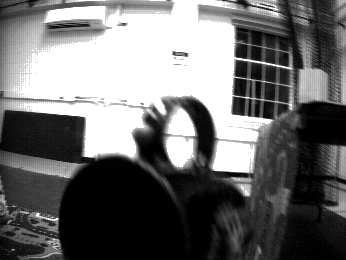} \\
    \vspace{0.03\linewidth}
    \includegraphics[width=\textwidth]{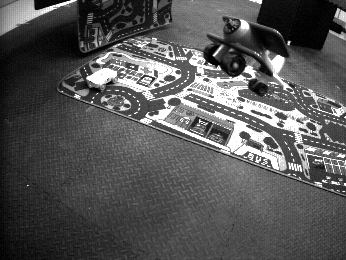} \\
    Image2
\end{minipage}%
\hspace{0.001\linewidth}
\begin{minipage}[t]{0.11\linewidth}
    \centering
    \includegraphics[width=\textwidth]{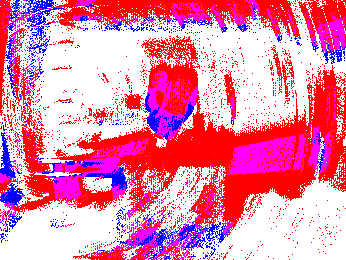} \\
    \vspace{0.03\linewidth}
    \includegraphics[width=\textwidth]{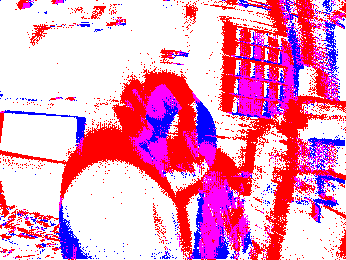} \\
    \vspace{0.03\linewidth}
    \includegraphics[width=\textwidth]{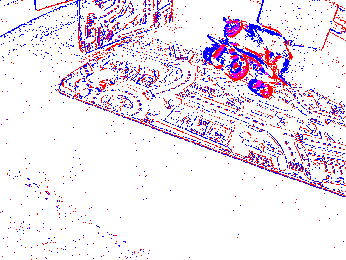} \\
    Events
\end{minipage}%
\hspace{0.001\linewidth}
\begin{minipage}[t]{0.11\linewidth}
    \centering
    \includegraphics[width=\textwidth]{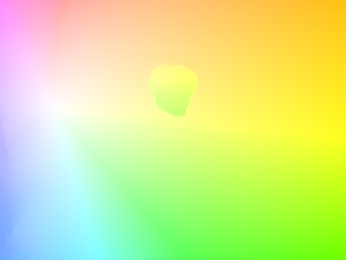} \\
    \vspace{0.03\linewidth}
    \includegraphics[width=\textwidth]{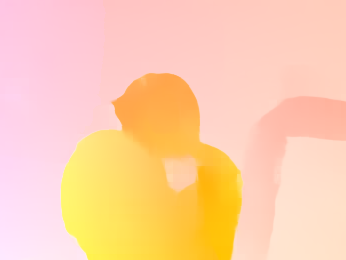} \\
    \vspace{0.03\linewidth}
    \includegraphics[width=\textwidth]{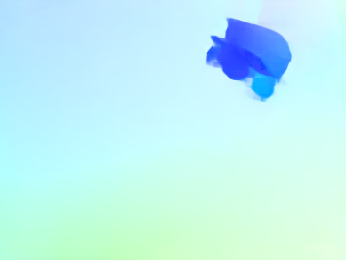} \\
    RAFT~\cite{flow:Teed_RAFT_ECCV_2020}
\end{minipage}%
\hspace{0.001\linewidth}
\begin{minipage}[t]{0.11\linewidth}
    \centering
    \includegraphics[width=\textwidth]{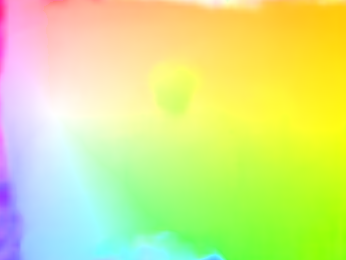} \\
    \vspace{0.03\linewidth}
    \includegraphics[width=\textwidth]{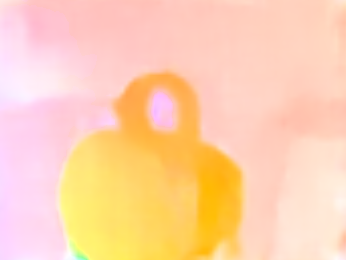} \\
    \vspace{0.03\linewidth}
    \includegraphics[width=\textwidth]{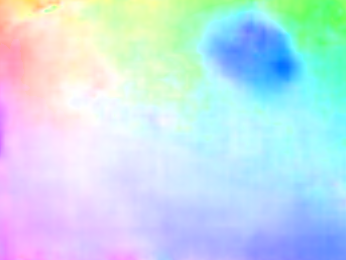} \\
    PWC-Net~\cite{flow:Sun_PWCNet_TPAMI_2019}
\end{minipage}%
\hspace{0.001\linewidth}
\begin{minipage}[t]{0.11\linewidth}
    \centering
    \includegraphics[width=\textwidth]{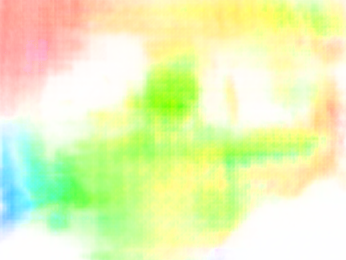} \\
    \vspace{0.03\linewidth}
    \includegraphics[width=\textwidth]{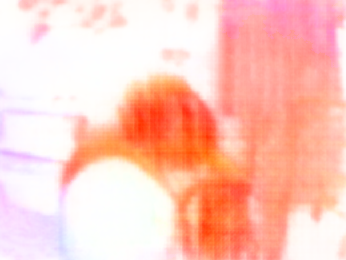} \\
    \vspace{0.03\linewidth}
    \includegraphics[width=\textwidth]{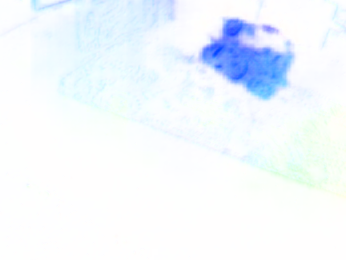} \\
    SpikeFlowNet~\cite{eventflow:Lee_SpikeFlowNet_ECCV_2020}
\end{minipage}%
\hspace{0.001\linewidth}
\begin{minipage}[t]{0.11\linewidth}
    \centering
    \includegraphics[width=\textwidth]{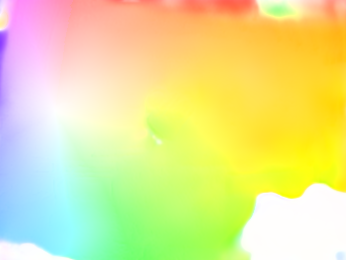} \\
    \vspace{0.03\linewidth}
    \includegraphics[width=\textwidth]{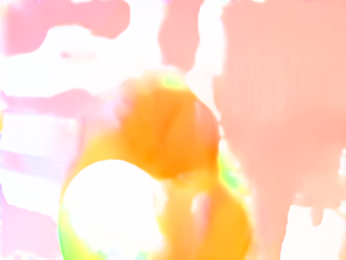} \\
    \vspace{0.03\linewidth}
    \includegraphics[width=\textwidth]{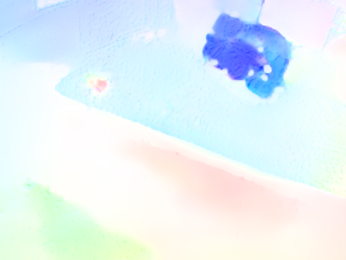} \\
    Stoffregen~\etal~\cite{eventflow:Stoffregen_ReducingGAP_ECCV_2020}
\end{minipage}%
\hspace{0.001\linewidth}
\begin{minipage}[t]{0.11\linewidth}
    \centering
    \includegraphics[width=\textwidth]{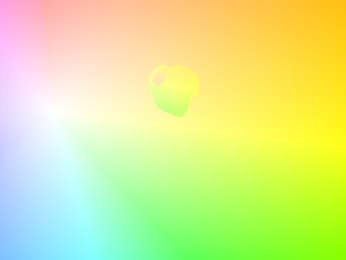} \\
    \vspace{0.03\linewidth}
    \includegraphics[width=\textwidth]{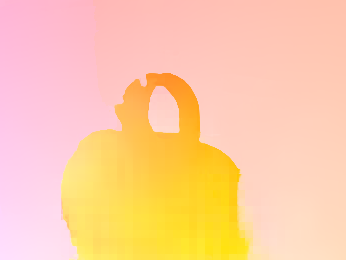} \\
    \vspace{0.03\linewidth}
    \includegraphics[width=\textwidth]{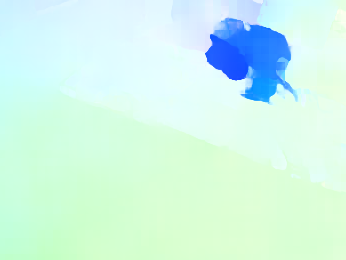} \\
    DCEIFlow~(Ours)
\end{minipage}%

\end{minipage}%

\vspace{-5pt}
\centering
\caption{\textbf{Visual comparisons on the EV-IMO~\cite{eventdatasets:mitrokhin_EVIMO_IROS_2019} dataset.} The models used for inference are all the pre-trained models that have not been trained on the EV-IMO dataset. Since the ground-truth flow annotations are not publicly available, we use the results of the two-frame methods as a reference. Best viewed on screen.}
\label{viz:evimo}
\vspace{-10pt}
\end{figure*}

\subsection{Results on the MVSEC dataset}

\Fix{Most event-based optical flow estimation methods report results on the MVSEC~\cite{eventdatasets:Zhu_MVSEC_RAL_2018} dataset. Therefore, we compare our model pre-trained on the FlyingChairs2 dataset~\cite{flowdatasets:Ilg_FlyingChairs2_ECCV_2018} (\ien, FC2) with them in Table~\ref{table:mvsec_frame_result}. }
Existing event-based methods usually train two different models to separately evaluate on different time intervals ($dt$=1 or $dt$=4 frames) respectively, but we only use one model for these two settings. 
Note that the results in Shedligeri~\etal~\cite{eventflow:Shedligeri_HighFPSEventFlow_CVIU_2021} and Mostafavi~\etal~\cite{event:Mostafavi_LearningEventHDR_IJCV_2021} are calculated with a fixed number of events ($dt=15000$ or $dt=30000$ events).
So we do not compare with them because we follow the commonly used fixed frame intervals setting.
\Fix{We also show the results from four two-frame-based methods only as a reference comparison in Table~\ref{table:mvsec_more_result}, including two supervised methods (PWC-Net~\cite{flow:Sun_PWCNet_TPAMI_2019} and RAFT~\cite{flow:Teed_RAFT_ECCV_2020}) and two unsupervised methods (ARFlow~\cite{flow:liu_ARFlow_cvpr_2020} and SMURF~\cite{flow:stone_smurf_cvpr_2021}).}

\Fix{We annotate the training manners for each method in Table~\ref{table:mvsec_frame_result}. In the column of training manners (\textbf{Train Mann.}), \textbf{SL} represents that the method is trained in a supervised manner, while \textbf{USL} represents an unsupervised manner. In addition, we annotate the data types (\textbf{Train D.Type}) used in the training process for each method. For the USL methods, the data type with $\mathrm{I_1}$,$\mathrm{I_2}$,$\mathrm{E}$ indicates that the loss function of this method is based on the warping of the APS images, while the data type with $\mathrm{E}$ indicates that it is based on the warping of events. In particular, for the model-based optimization method Pan~\etal~\cite{eventflow:Pan_SingleImageFlow_CVPR_2020}, we annotate it as \textbf{MB}. The model-based methods usually do not need pre-training but need to manually adjust the hyper-parameters for different input data.}

As shown in Table~\ref{table:mvsec_frame_result}, our model achieves state-of-the-art performance on indoor sequences for both EPE and outlier metrics. Although we do not get the best EPE on the $ourdoor\_day1$ sequence, the outlier metric is significantly lower than others, and the performance on $indoor$1-3 sequences is good enough to verify our advantage.
Especially for the longer time and larger motion evaluation setting with $dt$=4, our results have been greatly improved compared with others. 
Moreover, as a dense optical flow estimation method using a single image with events, our performance is comparable to the existing two-frame-based SOTAs.
This superior performance shows the effectiveness of our framework in fusing the first image and events for accurate dense flow prediction. 
\Fix{As shown in Table~\ref{table:mvsec_more_result}, we use the pre-trained model of E-RAFT~\cite{eventflow:Gehrig_DenseRAFTFlow_3DV_2021} on DSEC~\cite{eventdatasets:Gehrig_dsec_ral_2021} to compare with our model pre-trained on FlyingChairs2~\cite{flowdatasets:Ilg_FlyingChairs2_ECCV_2018}. 
The DSEC dataset used for pre-training E-RAFT is real captured on outdoor vehicles, while the FlyingChairs2 dataset we used is simulated with multiple chairs superimposed on a random image. 
Therefore, the performance of E-RAFT is comparable to our DCEIFlow model when evaluated on the outdoor sequences of the MVSEC dataset, while our model performs significantly better in the indoor sequences due to the introduction of image data.
Moreover, using a single 2080ti GPU, our model only takes 28ms to process data with MVSEC size and get flow prediction, while E-RAFT takes 62ms. }

\begin{table*}[tbp]
  \captionsetup{format=plain,labelformat=simple,justification=centering, labelsep=newline, singlelinecheck=false, textfont={sc}}
  \caption{
  \textbf{\Fix{Performance comparison between different baselines and our proposed DCEIFlow model on the MVSEC~\cite{eventdatasets:Zhu_MVSEC_RAL_2018} dataset.
  We use the same pre-trained model to get the results of each method under two input interval settings  ($dt\!=\!1$ and $dt\!=\!4$ frames).}}}
  \label{table:baselines_mvsec}

  \resizebox{\textwidth}{!}{
  \Large
  \begin{threeparttable}
  \begin{tabular}{cc|ccc|c|cccccccccc}
    \toprule
    Input & Method & Train & Train & Train & Eval. & \multicolumn{2}{c}{\textit{indoor\_flying1}}& \multicolumn{2}{c}{\textit{indoor\_flying2}}& \multicolumn{2}{c}{\textit{indoor\_flying3}}& \multicolumn{2}{c}{\textit{outdoor\_day1}} & \multicolumn{2}{c}{\textit{outdoor\_day2}} \cr
    \textbf{$dt\!=\!1$} &  & Mann. & D.Type & D.Set & Metric & EPE & \%Out & EPE & \%Out & EPE & \%Out & EPE & \%Out & EPE & \%Out \cr
    \midrule
    \multirow{4}*{$\bf{E}$}
    & Baseline-EV & SL & $\mathrm{E}$ & C2 & dense & 0.91 & 1.06 & 0.98 & 1.04 & 1.02 & 1.60 & 1.09 & 1.56 & 0.99 & 4.07 \cr
    & Baseline-EV & SL & $\mathrm{E}$ & C2 & sparse & 0.93 & 1.11 & 1.04 & 2.47 & 1.00 & 1.22 & 0.95 & 0.82 & 0.98 & 2.52 \cr
    & Baseline-EV & SL & $\mathrm{E}$ & M & dense & 0.80 & 0.65 & 0.95 & 2.04 & 0.89 & 1.56 & 0.44 & 0.03 & $\times$ & $\times$ \cr
    & Baseline-EV & SL & $\mathrm{E}$ & M & sparse & 0.89 & 1.17 & 1.14 & 4.31 & 0.97 & 2.50 & 0.47 & 0.10 & $\times$ & $\times$ \cr
    \midrule
    \multirow{8}*{$\bf{I_1}$+$\bf{E}$}
    & Baseline-EI & SL & $\mathrm{I_1}$,$\mathrm{E}$ & C2 & dense & 0.78 & 0.49 & 0.81 & 0.58 & 0.80 & 0.25 & 0.92 & 0.75 & 0.84 & 3.33 \cr
    & Baseline-EI & SL & $\mathrm{I_1}$,$\mathrm{E}$ & C2 & sparse & 0.82 & 0.60 & 0.89 & 1.37 & 0.82 & 0.40 & 0.80 & 0.54 & 0.83 & 2.69 \cr
    & Baseline-EI & SL & $\mathrm{I_1}$,$\mathrm{E}$ & M & dense & 0.75 & 0.54 & 0.80 & 0.77 & 0.80 & 0.93 & 0.36 & 0.00 & $\times$ & $\times$ \cr
    & Baseline-EI & SL & $\mathrm{I_1}$,$\mathrm{E}$ & M & sparse & 0.78 & 0.68 & 0.91 & 1.45 & 0.82 & 1.03 & 0.36 & 0.00 & $\times$ & $\times$ \cr
    & DCEIFlow~(Ours) & SL & $\mathrm{I_1}$,$\mathrm{I_2}$,$\mathrm{E}$ & C2 & dense & \textbf{0.56} & \textbf{0.28} & \textbf{0.64} & \textbf{0.16} & \textbf{0.57} & \textbf{0.12} & 0.91 & 0.71 & \textbf{0.79} & 2.59 \cr
    & DCEIFlow~(Ours) & SL & $\mathrm{I_1}$,$\mathrm{I_2}$,$\mathrm{E}$ & C2 & sparse & 0.57 & 0.30 & 0.70 & 0.30 & 0.58 & 0.15 & \textbf{0.74} & \textbf{0.29} & 0.82 & \textbf{2.34} \cr
    & DCEIFlow~(Ours) & SL & $\mathrm{I_1}$,$\mathrm{I_2}$,$\mathrm{E}$ & M & dense & 0.64 & 0.87 & 0.74 & 1.16 & 0.70 & 1.08 & 0.20 & 0.00 & $\times$ & $\times$ \cr
    & DCEIFlow~(Ours) & SL & $\mathrm{I_1}$,$\mathrm{I_2}$,$\mathrm{E}$ & M & sparse & 0.75 & 1.55 & 0.90 & 2.10 & 0.80 & 1.77 & 0.22 & 0.00 & $\times$ & $\times$ \cr
    \bottomrule
  \end{tabular}
  \end{threeparttable}
  }

  \resizebox{\textwidth}{!}{
  \Large
  \begin{threeparttable}
  \begin{tabular}{cc|ccc|c|cccccccccc}
    \toprule
    Input & Method & Train & Train & Train & Eval. & \multicolumn{2}{c}{\textit{indoor\_flying1}}& \multicolumn{2}{c}{\textit{indoor\_flying2}}& \multicolumn{2}{c}{\textit{indoor\_flying3}}& \multicolumn{2}{c}{\textit{outdoor\_day1}} & \multicolumn{2}{c}{\textit{outdoor\_day2}} \cr
    \textbf{$dt\!=\!4$} & & Mann. & D.Type & D.Set & Metric & EPE & \%Out & EPE & \%Out & EPE & \%Out & EPE & \%Out & EPE & \%Out \cr
    \midrule
    \multirow{4}*{$\bf{E}$}
    & Baseline-EV & SL & $\mathrm{E}$ & C2 & dense & 1.76 & 13.21 & 2.05 & 19.09 & 1.99 & 18.73 & 2.63 & 30.84 & 2.10 & 20.83 \cr
    & Baseline-EV & SL & $\mathrm{E}$ & C2 & sparse & 1.72 & 12.37 & 2.24 & 22.98 & 1.91 & 16.57 & 1.96 & 17.91 & 2.01 & 18.86 \cr
    & Baseline-EV & SL & $\mathrm{E}$ & M & dense & 2.59 & 29.67 & 3.53 & 39.82 & 2.87 & 31.21 & 1.55 & 12.24 & $\times$ & $\times$ \cr
    & Baseline-EV & SL & $\mathrm{E}$ & M & sparse & 3.04 & 36.87 & 4.55 & 52.58 & 3.38 & 36.72 & 1.64 & 13.64 & $\times$ & $\times$ \cr
    \midrule
    \multirow{8}*{$\bf{I_1}$+$\bf{E}$}
    & Baseline-EI & SL & $\mathrm{I_1}$,$\mathrm{E}$ & C2 & dense & 1.66 & 11.17 & 2.13 & 20.17 & 1.78 & 14.56 & 2.25 & 25.19 & 1.92 & 19.39 \cr
    & Baseline-EI & SL & $\mathrm{I_1}$,$\mathrm{E}$ & C2 & sparse & 1.65 & 11.21 & 2.37 & 25.70 & 1.79 & 14.57 & 1.66 & 13.20 & 1.83 & 17.55 \cr
    & Baseline-EI & SL & $\mathrm{I_1}$,$\mathrm{E}$ & M & dense & 2.08 & 21.13 & 2.77 & 29.20 & 2.32 & 24.59 & 1.16 & 5.98 & $\times$ & $\times$ \cr
    & Baseline-EI & SL & $\mathrm{I_1}$,$\mathrm{E}$ & M & sparse & 2.21 & 23.61 & 3.37 & 36.99 & 2.47 & 26.50 & 1.11 & 5.01 & $\times$ & $\times$ \cr
    & DCEIFlow~(Ours) & SL & $\mathrm{I_1}$,$\mathrm{I_2}$,$\mathrm{E}$ & C2 & dense & \textbf{1.49} & \textbf{8.14} & \textbf{1.97} & \textbf{17.37} & \textbf{1.69} & \textbf{12.34} & 1.87 & 19.13 & 1.62 & 14.73 \cr
    & DCEIFlow~(Ours) & SL & $\mathrm{I_1}$,$\mathrm{I_2}$,$\mathrm{E}$ & C2 & sparse & 1.52 & 8.79 & 2.21 & 22.13 & 1.74 & 13.33 & \textbf{1.37} & \textbf{8.54} & \textbf{1.61} & \textbf{14.38} \cr
    & DCEIFlow~(Ours) & SL & $\mathrm{I_1}$,$\mathrm{I_2}$,$\mathrm{E}$ & M & dense & 1.90 & 17.43 & 2.97 & 34.38 & 2.32 & 26.07 & 0.87 & 3.12 & $\times$ & $\times$ \cr
    & DCEIFlow~(Ours) & SL & $\mathrm{I_1}$,$\mathrm{I_2}$,$\mathrm{E}$ & M & sparse & 2.08 & 21.47 & 3.48 & 42.05 & 2.51 & 29.73 & 0.89 & 3.19 & $\times$ & $\times$ \cr
    \bottomrule
  \end{tabular}
  \end{threeparttable}
  }
\end{table*}

\begin{table*}[htbp]
  \captionsetup{format=plain,labelformat=simple,justification=centering, labelsep=newline, singlelinecheck=false, textfont={sc}}
  \caption{\Fix{\textbf{Dense flow prediction analysis.} The results are averaged on the MVSEC \textit{indoor\_flying1-3} sequences.}}
  \label{table:densesparse}
  \centering
  \resizebox{\textwidth}{!}{
  \small
  \begin{threeparttable}
  \begin{tabular}{ccccc|cccccc|c}
    \toprule
    Input & Method & Claim & Train. & Eval. & \multicolumn{2}{c}{Full valid pixels} & \multicolumn{2}{c}{Event Masked} & \multicolumn{2}{c|}{Event Excluded} & Dense \cr
    & & Setting & set~\textbf{($dt$)} & \textbf{$dt$} & EPE & \%Out & EPE & \%Out & EPE & \%Out & Ratio \cr
    \midrule
    \multirow{8}*{$\bf{I_1}$+$\bf{I_2}$}
    & PWC-Net~\cite{flow:Sun_PWCNet_TPAMI_2019} & dense & C+T & 1 & 1.58 & 3.03 & 1.62 & 3.73 & 1.58 & 2.96 & 1.013 \cr
    & RAFT~\cite{flow:Teed_RAFT_ECCV_2020} & dense & C & 1 & 0.49 & 0.06 & 0.55 & 0.07 & 0.49 & 0.06 & 1.060 \cr
    & ARFlow~\cite{flow:liu_ARFlow_cvpr_2020} & dense & SR+S & 1 & 0.43 & 0.07 & 0.42 & 0.05 & 0.43 & 0.08 & \underline{0.996} \cr
    & SMURF~\cite{flow:stone_smurf_cvpr_2021} & dense & C & 1 & 0.46 & 0.19 & 0.44 & 0.08 & 0.46 & 0.20 & \underline{0.981} \cr
    \cmidrule(r){2-12}
    & PWC-Net~\cite{flow:Sun_PWCNet_TPAMI_2019} & dense & C+T & 4 & 2.05 & 17.47 & 2.11 & 18.99 & 2.03 & 16.83 & 1.019 \cr
    & RAFT~\cite{flow:Teed_RAFT_ECCV_2020} & dense & C & 4 & 1.63 & 10.92 & 1.69 & 11.68 & 1.61 & 10.49 & 1.024 \cr
    & ARFlow~\cite{flow:liu_ARFlow_cvpr_2020} & dense & SR+S & 4 & 1.44 & 7.93 & 1.50 & 9.15 & 1.41 & 7.20 & 1.034 \cr
    & SMURF~\cite{flow:stone_smurf_cvpr_2021} & dense & C & 4 & 1.53 & 9.29 & 1.54 & 9.72 & 1.52 & 8.86 & 1.007 \cr
    \midrule
    \multirow{12}*{$\bf{E}$}
    & SpikeFlowNet~\cite{eventflow:Lee_SpikeFlowNet_ECCV_2020} & sparse & M~(1) & 1 & 1.14 & 4.67 & 1.08 & 4.03 & 1.15 & 5.16 & \underline{0.969} \cr
    & Stoffregen~\etal~\cite{eventflow:Stoffregen_ReducingGAP_ECCV_2020} & dense & ESIM & 1 & 0.75 & 0.57 & 0.65 & 0.41 & 0.76 & 0.60 & \underline{0.927} \cr
    & E-RAFT~\cite{eventflow:Gehrig_DenseRAFTFlow_3DV_2021} & dense & DSEC & 1 & 0.82 & 1.54 & 0.97 & 2.76 & 0.80 & 1.33 & 1.105 \cr
    & Baseline-EV & dense & M~(1) & 1 & 0.83 & 2.38 & 0.99 & 4.60 & 0.80 & 1.96 & 1.114 \cr
    & Baseline-EV & dense & M~(1\&4) & 1 & 0.88 & 1.42 & 1.00 & 2.66 & 0.86 & 1.15 & 1.080 \cr
    & Baseline-EV & dense & C2 & 1 & 0.97 & 1.24 & 0.99 & 1.60 & 0.97 & 1.18 & 1.010 \cr
    \cmidrule(r){2-12}
    & SpikeFlowNet~\cite{eventflow:Lee_SpikeFlowNet_ECCV_2020} & sparse & M~(4) & 4 & 3.65 & 45.42 & 3.08 & 33.45 & 3.78 & 49.01 & \underline{0.906} \cr
    & Stoffregen~\etal~\cite{eventflow:Stoffregen_ReducingGAP_ECCV_2020} & dense & ESIM & 4 & 3.08 & 35.91 & 2.29 & 21.03 & 3.35 & 40.79 & \underline{0.835} \cr
    & E-RAFT~\cite{eventflow:Gehrig_DenseRAFTFlow_3DV_2021} & dense & DSEC & 4 & 2.19 & 19.55 & 2.46 & 23.15 & 2.05 & 17.97 & 1.098 \cr
    & Baseline-EV & dense & M~(4) & 4 & 3.12 & 35.51 & 3.60 & 39.99 & 2.84 & 33.36 & 1.128 \cr
    & Baseline-EV & dense & M~(1\&4) & 4 & 3.00 & 33.57 & 3.66 & 42.06 & 2.66 & 30.01 & 1.177 \cr
    & Baseline-EV & dense & C2 & 4 & 1.93 & 17.01 & 1.96 & 17.31 & 1.90 & 16.46 & 1.015 \cr
    \midrule
    \multirow{12}*{$\bf{I_1}$+$\bf{E}$}
    & Baseline-EI & dense & M~(1) & 1 & 0.77 & 1.91 & 0.91 & 3.61 & 0.75 & 1.59 & 1.106 \cr
    & Baseline-EI & dense & M~(1\&4) & 1 & 0.78 & 0.75 & 0.84 & 1.05 & 0.77 & 0.68 & 1.040 \cr
    & Baseline-EI & dense & C2 & 1 & 0.80 & 0.44 & 0.84 & 0.79 & 0.79 & 0.40 & 1.033 \cr
    & DCEIFlow~(Ours) & dense & M~(1) & 1 & 0.78 & 1.60 & 0.92 & 2.55 & 0.76 & 1.45 & 1.102 \cr
    & DCEIFlow~(Ours) & dense & M~(1\&4) & 1 & 0.67 & 0.48 & 0.77 & 0.89 & 0.66 & 0.42 & 1.080 \cr
    & DCEIFlow~(Ours) & dense & C2 & 1 & \textbf{0.59} & \textbf{0.18} & \textbf{0.62} & \textbf{0.25} & \textbf{0.58} & \textbf{0.18} & 1.027 \cr
    \cmidrule(r){2-12}
    & Baseline-EI & dense & M~(4) & 4 & 2.73 & 32.52 & 3.00 & 36.79 & 2.57 & 30.34 & 1.081 \cr
    & Baseline-EI & dense & M~(1\&4) & 4 & 2.39 & 24.97 & 2.69 & 29.03 & 2.22 & 23.02 & 1.100 \cr
    & Baseline-EI & dense & C2 & 4 & 1.85 & 15.30 & 1.94 & 17.16 & 1.80 & 14.18 & 1.036 \cr
    & DCEIFlow~(Ours) & dense & M~(4) & 4 & 2.58 & 22.00 & 2.93 & 25.80 & 2.41 & 20.41 & 1.106 \cr
    & DCEIFlow~(Ours) & dense & M~(1\&4) & 4 & 2.24 & 22.42 & 2.52 & 27.33 & 2.10 & 20.27 & 1.098 \cr
    & DCEIFlow~(Ours) & dense & C2 & 4 & \textbf{1.72} & \textbf{12.62} & \textbf{1.82} & \textbf{14.75} & \textbf{1.67} & \textbf{11.69} & 1.045 \cr
    \bottomrule
  \end{tabular}
  \end{threeparttable}
  }
\end{table*}

In addition, we make visual comparisons with several event-based methods, which have open-sourced models in Fig.~\ref{viz:mvsec}.
EV-FlowNet~\cite{eventflow:Zhu_EVFlowNet_RSS_2018} and SpikeFlowNet~\cite{eventflow:Lee_SpikeFlowNet_ECCV_2020} only use events, and their claim is to predict sparse flow. 
Their visualizations are also sparse and include many incorrect predictions (such as the upper left corner of the first sample in Fig.~\ref{viz:mvsec}). 
For another event-only method Stoffregen~\etal~\cite{eventflow:Stoffregen_ReducingGAP_ECCV_2020}, although its claim is dense prediction, it is difficult to predict a complete dense flow in the area without events.
\Fix{Most of the motion in the MVSEC dataset is caused by the camera, and the ground-truth optical flow labels are calculated from sparse depth and camera motion. Thus there are spatial mismatches between images and optical flow in some scenes, which also increases the difficulty of visual comparison.
Despite the slight mismatches, we can still conclude that our proposed DCEIFlow model produces not only fewer errors but also more dense estimations. This is consistent with the conclusion of the above quantitative comparison.}

\subsection{Results on the EV-IMO dataset}

To verify the performance of our model on a more challenging dataset EV-IMO~\cite{eventdatasets:mitrokhin_EVIMO_IROS_2019} with fast-moving objects, we also run SpikeFlowNet~\cite{eventflow:Lee_SpikeFlowNet_ECCV_2020} and Stoffregen~\etal~\cite{eventflow:Stoffregen_ReducingGAP_ECCV_2020} with their open sourced models, and the visual comparisons are shown in Fig.~\ref{viz:evimo}. 
Because there is no ground-truth optical flow, we take the results of the two-frame methods (PWC-Net~\cite{flow:Sun_PWCNet_TPAMI_2019}, and RAFT~\cite{flow:Teed_RAFT_ECCV_2020}) as a reference for comparative analysis.
We found that RAFT achieved perfect flow visualization using two frames, but we also found its shortcomings in some detailed areas compared with ours, especially the edge or hole position.
We believe this is the advantage of introducing the events which contain detailed motion. 
For event-based methods, it is obvious that their predictions are not dense and accurate enough, especially SpikeFlowNet makes some unique estimates for foreground objects. 
Our results are denser than them, and most consistent with RAFT. 
This further verifies the superior generalization performance of our model compared with the existing event-based methods.

\subsection{Model analysis}

\subsubsection{Baselines for comparison}

\Fix{To fully illustrate the superiority of our model, we experiment with the same training setting for both baseline models, i.e., 100 epochs on FlyingChairs2 or 300 epochs on the MVSEC dataset using the same hyper-parameters. See Sec.~\ref{section:training_details} for more training details. }
Table~\ref{table:baselines_mvsec} show the results of two baselines on the MVSEC dataset. 
In addition, we also evaluate their dense flow prediction ability in Table~\ref{table:densesparse}. 
By comparing the two baselines, our proposed DCEIFlow model improves significantly when the model size is less than theirs. 
This shows that our model is more suitable and powerful than the baselines for event-based dense flow estimation. 
\Fix{When trained with only one frame interval setting, our model also achieves better performance than the compared baselines and the two event-based methods SpikeFlownet~\cite{eventflow:Lee_SpikeFlowNet_ECCV_2020} and Stoffregen~\etal~\cite{eventflow:Stoffregen_ReducingGAP_ECCV_2020}.
In addition, we found that the model trained with two interval settings can achieve better performance than with only one. 
We think this is because larger data sizes help supervised learning to achieve a better model. }

\Fix{Furthermore, Table~\ref{table:baselines_mvsec} shows that our proposed DCEIFlow model outperforms the competing event-based baseline models even when directly trained on the MVSEC dataset without pre-training. Compared with the existing event-based unsupervised methods in Table~\ref{table:mvsec_frame_result}, our model achieves comparable, if not better, performance.
These results show that the methods with unsupervised training usually have better generalization ability. The methods with supervised training may not achieve superior results when trained on limited data. 
Thus our model achieves a significant performance improvement when pre-trained with a large-scale dataset FlyingChairs2 (\ien, FC2).}

\subsubsection{\Fix{Supervised and unsupervised training}}
\label{section:supervised_vs_un}
\Fix{As shown in Table~\ref{table:mvsec_frame_result}, some existing methods use unsupervised loss functions for training, while our model is obtained by supervised training using ground-truth supervision. 
For the four methods with input data is $\bf{I_1}+\bf{I_2}$, we only evaluate their pre-trained model for reference comparisons.
Because SMURF and ARFlow have similar structures to RAFT and PWC-Net, respectively, the experiments on the MVSEC dataset (\cf~Table~\ref{table:mvsec_more_result}) indicate that the unsupervised methods usually have better generalization performance.
However, in the dense prediction analysis (\cf~Table~\ref{table:densesparse}), the unsupervised methods usually have worse results in the weakly textured regions (i.e., without events) than in the richly textured regions (i.e., with events), while the supervised methods are usually better in the weakly textured regions. 
This conclusion has also been verified by the results of the event-based unsupervised methods. 
We conclude that the supervised methods can generally achieve better dense flow estimation, but their generalization ability is worse than unsupervised methods under the same training protocols. }

\subsubsection{Ablation study}
\label{section:ablation}
We conducted ablations to confirm the effectiveness of each module in our framework, including 1) event polarity representation, 2) event-image correlation module, 3) event-image feature fusion, and 4) bidirectional flow training. In addition, we also experimented with the commonly used pyramid structure~\cite{flow:Sun_PWCNet_TPAMI_2019} to verify the effectiveness of our iterative structure. More details of this pyramid structure are described in the supplementary material.

The results are shown in Table~\ref{table:ablation}. We use the same training setting to pre-train on the FlyingChairs2 dataset for the nine models composed of the above five parts. In models (a)\&(b), separating events by their polarity during represent events into event volume can reduce the information loss caused by positive and negative coexistence, and improve the accuracy. In models (b)\&(c), introducing the correlation construction can greatly improve the accuracy of flow prediction, which is also proved in the existing two-frame methods~\cite{flow:Sun_PWCNet_TPAMI_2019}. At the same time, because the input dimension of the decoder is increased, the number of network parameters is also increased. In models (c)\&(e) and (d)\&(f), our proposed fusion by convolution structure is more suitable compared to simple addition. In models (e)\&(g) and (f)\&(i), the proposed bidirectional training mechanism can further improve the performance without increasing the number of network parameters, and the results of the iterative structure are better than pyramid structure with smaller parameters. 
\Fix{In models (h)\&(i), since there is no constraint on the output of the fusion module, the correlation module is challenging to realize the function to search for matching points in the neighborhood. The comparison results also illustrate that the fusion module and the similar loss need to be used together to perform better.} In summary, the model with both parts obtains the best results, which demonstrates the effectiveness of each component in our framework. 
\Fix{In addition, we also compare the dense flow estimation performance of each model on the \textit{indoor\_flying1-3} sequences of the MVSEC dataset~\cite{eventdatasets:Zhu_MVSEC_RAL_2018} with $dt=1$. The comparison shows that models with better results on other datasets are usually superior on MVSEC, where the iteration-based models usually perform fewer outliers (\%Out) than the pyramid-based models.}

\begin{table*}[tbp]
  \setlength\tabcolsep{1pt}
  % \small
  \centering
    \begin{tabular}{p{1.6cm} p{1.7cm} p{1.7cm} p{1.7cm} p{1.7cm} p{1.7cm} p{1.7cm} p{1.7cm} p{1.7cm} p{1.7cm}}
    \begin{minipage}[b]{1.5cm}
    \centering
    Image1\\
    \raisebox{-.5\height}{($\bf{I_1}$)}
  \end{minipage}& 
    \begin{minipage}[b]{1.6cm}
    \centering
    \raisebox{-.5\height}{\includegraphics[width=\linewidth]{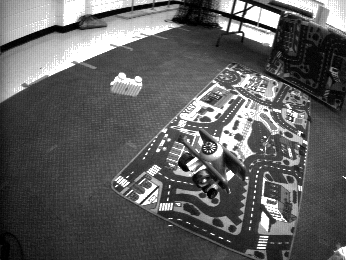}}
  \end{minipage}& 
    \begin{minipage}[b]{1.6cm}
    \centering
    \raisebox{-.5\height}{\includegraphics[width=\linewidth]{viz/continuous/table_seq_01_40/table_seq_01_40_image1.png}}
  \end{minipage}& 
    \begin{minipage}[b]{1.6cm}
    \centering
    \raisebox{-.5\height}{\includegraphics[width=\linewidth]{viz/continuous/table_seq_01_40/table_seq_01_40_image1.png}}
  \end{minipage}& 
    \begin{minipage}[b]{1.6cm}
    \centering
    \raisebox{-.5\height}{\includegraphics[width=\linewidth]{viz/continuous/table_seq_01_40/table_seq_01_40_image1.png}}
  \end{minipage}& 
    \begin{minipage}[b]{1.6cm}
    \centering
    \raisebox{-.5\height}{\includegraphics[width=\linewidth]{viz/continuous/table_seq_01_40/table_seq_01_40_image1.png}}
  \end{minipage}& 
    \begin{minipage}[b]{1.6cm}
    \centering
    \raisebox{-.5\height}{\includegraphics[width=\linewidth]{viz/continuous/table_seq_01_40/table_seq_01_40_image1.png}}
  \end{minipage}& 
    \begin{minipage}[b]{1.6cm}
    \centering
    \raisebox{-.5\height}{\includegraphics[width=\linewidth]{viz/continuous/table_seq_01_40/table_seq_01_40_image1.png}}
  \end{minipage}& 
    \begin{minipage}[b]{1.6cm}
    \centering
    \raisebox{-.5\height}{\includegraphics[width=\linewidth]{viz/continuous/table_seq_01_40/table_seq_01_40_image1.png}}
  \end{minipage}& 
    \begin{minipage}[b]{1.6cm}
    \centering
    \raisebox{-.5\height}{\includegraphics[width=\linewidth]{viz/continuous/table_seq_01_40/table_seq_01_40_image1.png}}
  \end{minipage} \cr
    \specialrule{0em}{0.5pt}{0.5pt}

    \begin{minipage}[b]{1.5cm}
    \centering
    Image2\\
    \raisebox{-.5\height}{($\bf{I_2}$)}
  \end{minipage}& 
    \begin{minipage}[b]{1.6cm}
    \centering
    -
  \end{minipage}& 
    \begin{minipage}[b]{1.6cm}
    \centering
    -
  \end{minipage}& 
    \begin{minipage}[b]{1.6cm}
    \centering
    \raisebox{-.5\height}{\includegraphics[width=\linewidth]{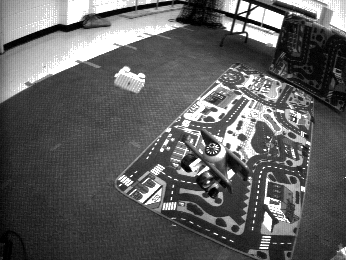}}
  \end{minipage}& 
    \begin{minipage}[b]{1.6cm}
    \centering
    -
  \end{minipage}& 
    \begin{minipage}[b]{1.6cm}
    \centering
    \raisebox{-.5\height}{\includegraphics[width=\linewidth]{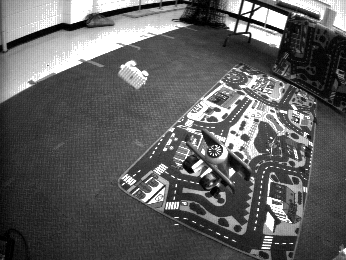}}
  \end{minipage}& 
    \begin{minipage}[b]{1.6cm}
    \centering
    -
  \end{minipage}& 
    \begin{minipage}[b]{1.6cm}
    \centering
    \raisebox{-.5\height}{\includegraphics[width=\linewidth]{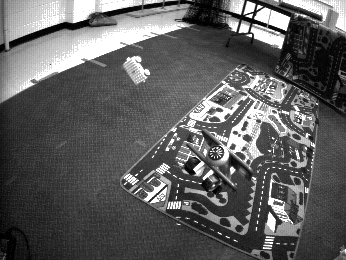}}
  \end{minipage}& 
    \begin{minipage}[b]{1.6cm}
    \centering
    -
  \end{minipage}& 
    \begin{minipage}[b]{1.6cm}
    \centering
    \raisebox{-.5\height}{\includegraphics[width=\linewidth]{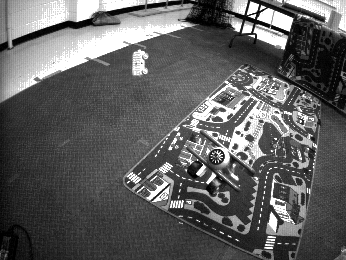}}
  \end{minipage} \cr
    \specialrule{0em}{0.5pt}{0.5pt}

    \begin{minipage}[b]{1.5cm}
    \centering
        \raisebox{-.5\height}{PWC-Net} \raisebox{-.5\height}{\cite{flow:Sun_PWCNet_TPAMI_2019} ($\bf{I_1}$+$\bf{I_2}$)}
  \end{minipage}& 
    \begin{minipage}[b]{1.6cm}
    \centering
    -
  \end{minipage}& 
    \begin{minipage}[b]{1.6cm}
    \centering
    -
  \end{minipage}& 
    \begin{minipage}[b]{1.6cm}
    \centering
    \raisebox{-.5\height}{\includegraphics[width=\linewidth]{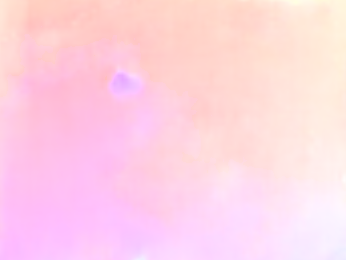}}
  \end{minipage}& 
    \begin{minipage}[b]{1.6cm}
    \centering
    -
  \end{minipage}& 
    \begin{minipage}[b]{1.6cm}
    \centering
    \raisebox{-.5\height}{\includegraphics[width=\linewidth]{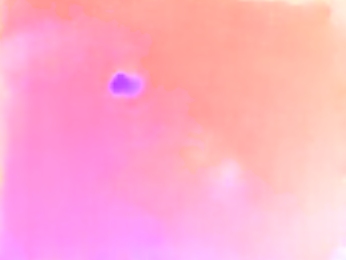}}
  \end{minipage}& 
    \begin{minipage}[b]{1.6cm}
    \centering
    -
  \end{minipage}& 
    \begin{minipage}[b]{1.6cm}
    \centering
    \raisebox{-.5\height}{\includegraphics[width=\linewidth]{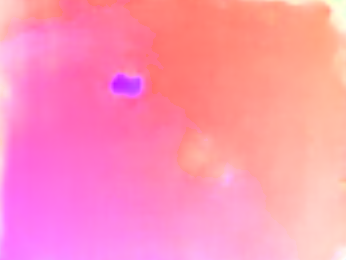}}
  \end{minipage}& 
    \begin{minipage}[b]{1.6cm}
    \centering
    -
  \end{minipage}& 
    \begin{minipage}[b]{1.6cm}
    \centering
    \raisebox{-.5\height}{\includegraphics[width=\linewidth]{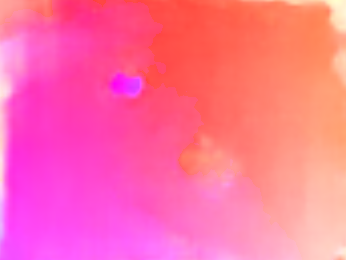}}
  \end{minipage} \cr
    \specialrule{0em}{0.5pt}{0.5pt}

    \begin{minipage}[b]{1.5cm}
    \centering
        \raisebox{-.5\height}{RAFT}
        \raisebox{-.5\height}{\cite{flow:Teed_RAFT_ECCV_2020} (\textbf{$\bf{I_1}$+$\bf{I_2}$})}
  \end{minipage}& 
    \begin{minipage}[b]{1.6cm}
    \centering
    -
  \end{minipage}& 
    \begin{minipage}[b]{1.6cm}
    \centering
    -
  \end{minipage}& 
    \begin{minipage}[b]{1.6cm}
    \centering
    \raisebox{-.5\height}{\includegraphics[width=\linewidth]{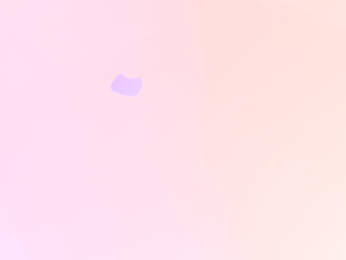}}
  \end{minipage}& 
    \begin{minipage}[b]{1.6cm}
    \centering
    -
  \end{minipage}& 
    \begin{minipage}[b]{1.6cm}
    \centering
    \raisebox{-.5\height}{\includegraphics[width=\linewidth]{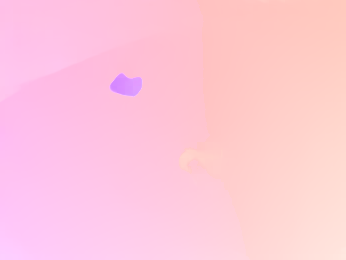}}
  \end{minipage}& 
    \begin{minipage}[b]{1.6cm}
    \centering
    -
  \end{minipage}& 
    \begin{minipage}[b]{1.6cm}
    \centering
    \raisebox{-.5\height}{\includegraphics[width=\linewidth]{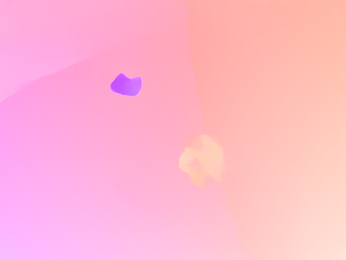}}
  \end{minipage}& 
    \begin{minipage}[b]{1.6cm}
    \centering
    -
  \end{minipage}& 
    \begin{minipage}[b]{1.6cm}
    \centering
    \raisebox{-.5\height}{\includegraphics[width=\linewidth]{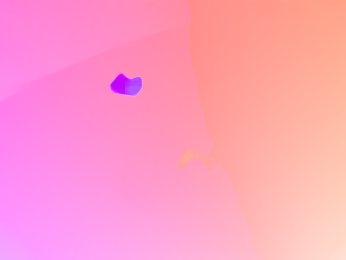}}
  \end{minipage} \cr
    \specialrule{0em}{0.5pt}{0.5pt}

    \begin{minipage}[b]{1.5cm}
    \centering
        Events\\
    \raisebox{-.5\height}{($\bf{E}$)}
  \end{minipage}& 
    \begin{minipage}[b]{1.6cm}
    \centering
    \raisebox{-.5\height}{\includegraphics[width=\linewidth]{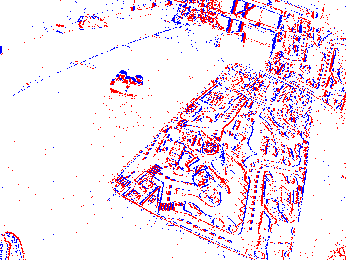}}
  \end{minipage}& 
    \begin{minipage}[b]{1.6cm}
    \centering
    \raisebox{-.5\height}{\includegraphics[width=\linewidth]{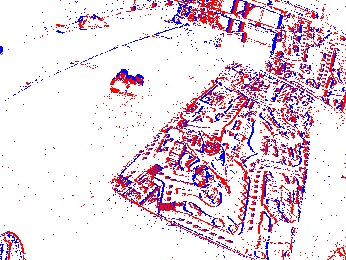}}
  \end{minipage}& 
    \begin{minipage}[b]{1.6cm}
    \centering
    \raisebox{-.5\height}{\includegraphics[width=\linewidth]{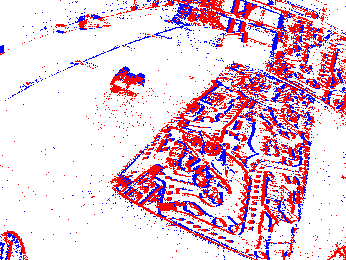}}
  \end{minipage}& 
    \begin{minipage}[b]{1.6cm}
    \centering
    \raisebox{-.5\height}{\includegraphics[width=\linewidth]{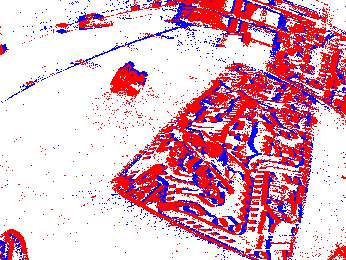}}
  \end{minipage}& 
    \begin{minipage}[b]{1.6cm}
    \centering
    \raisebox{-.5\height}{\includegraphics[width=\linewidth]{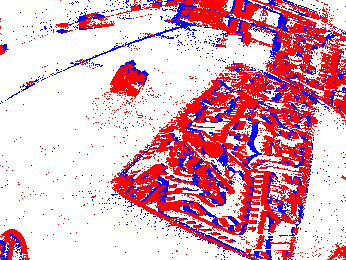}}
  \end{minipage}& 
    \begin{minipage}[b]{1.6cm}
    \centering
    \raisebox{-.5\height}{\includegraphics[width=\linewidth]{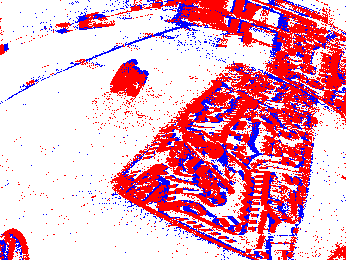}}
  \end{minipage}& 
    \begin{minipage}[b]{1.6cm}
    \centering
    \raisebox{-.5\height}{\includegraphics[width=\linewidth]{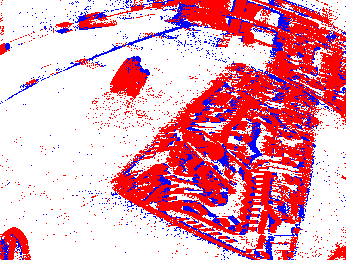}}
  \end{minipage}& 
    \begin{minipage}[b]{1.6cm}
    \centering
    \raisebox{-.5\height}{\includegraphics[width=\linewidth]{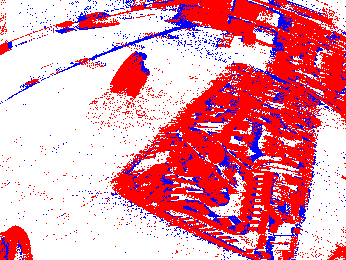}}
  \end{minipage}& 
    \begin{minipage}[b]{1.6cm}
    \centering
    \raisebox{-.5\height}{\includegraphics[width=\linewidth]{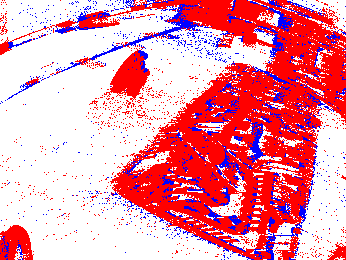}}
  \end{minipage} \cr
    \specialrule{0em}{0.5pt}{0.5pt}

    \begin{minipage}[b]{1.5cm}
    \centering
        \raisebox{-.5\height}{\scriptsize SpikeFlowNet} \raisebox{-.5\height}{\cite{eventflow:Lee_SpikeFlowNet_ECCV_2020} ($\bf{E}$)}
  \end{minipage}& 
    \begin{minipage}[b]{1.6cm}
    \centering
    \raisebox{-.5\height}{\includegraphics[width=\linewidth]{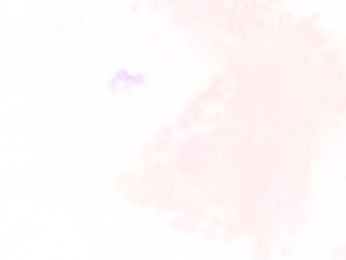}}
  \end{minipage}& 
    \begin{minipage}[b]{1.6cm}
    \centering
    \raisebox{-.5\height}{\includegraphics[width=\linewidth]{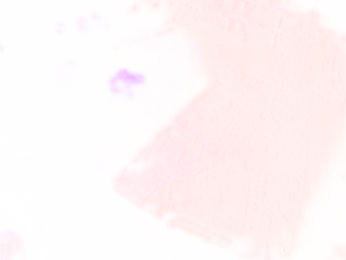}}
  \end{minipage}& 
    \begin{minipage}[b]{1.6cm}
    \centering
    \raisebox{-.5\height}{\includegraphics[width=\linewidth]{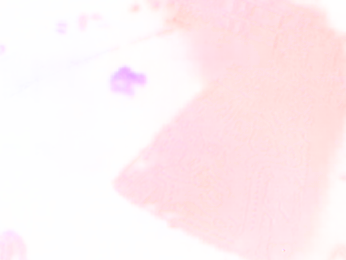}}
  \end{minipage}& 
    \begin{minipage}[b]{1.6cm}
    \centering
    \raisebox{-.5\height}{\includegraphics[width=\linewidth]{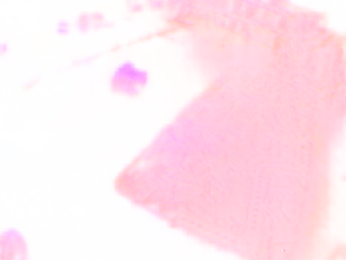}}
  \end{minipage}& 
    \begin{minipage}[b]{1.6cm}
    \centering
    \raisebox{-.5\height}{\includegraphics[width=\linewidth]{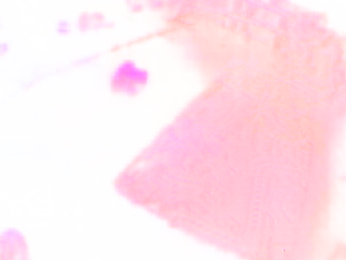}}
  \end{minipage}& 
    \begin{minipage}[b]{1.6cm}
    \centering
    \raisebox{-.5\height}{\includegraphics[width=\linewidth]{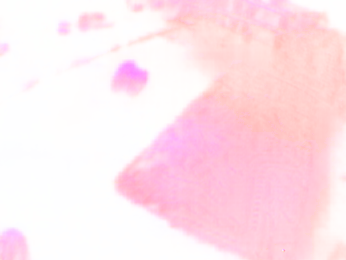}}
  \end{minipage}& 
    \begin{minipage}[b]{1.6cm}
    \centering
    \raisebox{-.5\height}{\includegraphics[width=\linewidth]{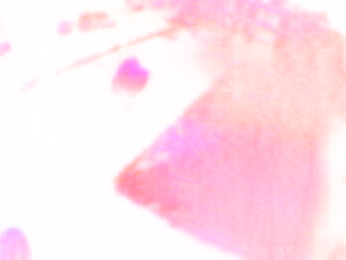}}
  \end{minipage}& 
    \begin{minipage}[b]{1.6cm}
    \centering
    \raisebox{-.5\height}{\includegraphics[width=\linewidth]{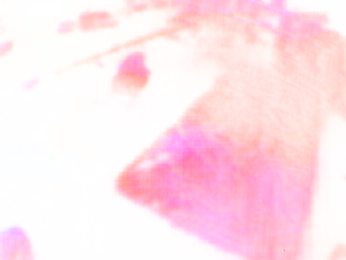}}
  \end{minipage}& 
    \begin{minipage}[b]{1.6cm}
    \centering
    \raisebox{-.5\height}{\includegraphics[width=\linewidth]{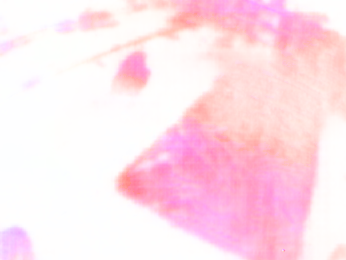}}
  \end{minipage} \cr
    \specialrule{0em}{0.5pt}{0.5pt}

    \begin{minipage}[b]{1.6cm}
    \centering
        \raisebox{-.5\height}{\scriptsize Stoffregen~\etal} \raisebox{-.5\height}{\cite{eventflow:Stoffregen_ReducingGAP_ECCV_2020} ($\bf{E}$)}
  \end{minipage}& 
    \begin{minipage}[b]{1.6cm}
    \centering
    \raisebox{-.5\height}{\includegraphics[width=\linewidth]{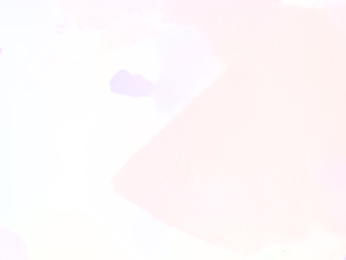}}
  \end{minipage}& 
    \begin{minipage}[b]{1.6cm}
    \centering
    \raisebox{-.5\height}{\includegraphics[width=\linewidth]{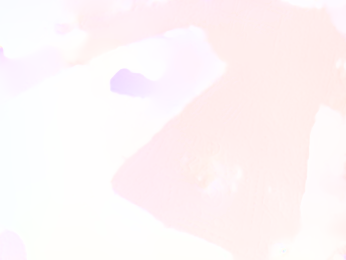}}
  \end{minipage}& 
    \begin{minipage}[b]{1.6cm}
    \centering
    \raisebox{-.5\height}{\includegraphics[width=\linewidth]{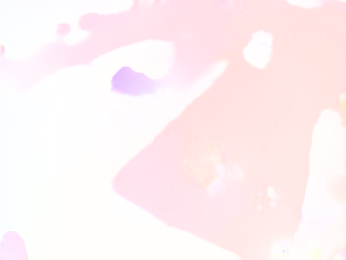}}
  \end{minipage}& 
    \begin{minipage}[b]{1.6cm}
    \centering
    \raisebox{-.5\height}{\includegraphics[width=\linewidth]{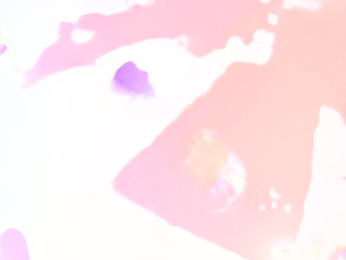}}
  \end{minipage}& 
    \begin{minipage}[b]{1.6cm}
    \centering
    \raisebox{-.5\height}{\includegraphics[width=\linewidth]{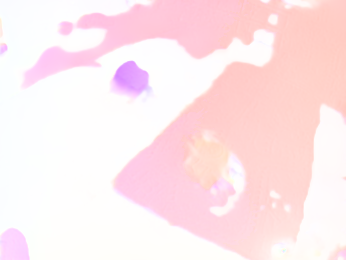}}
  \end{minipage}& 
    \begin{minipage}[b]{1.6cm}
    \centering
    \raisebox{-.5\height}{\includegraphics[width=\linewidth]{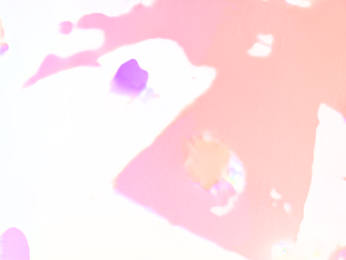}}
  \end{minipage}& 
    \begin{minipage}[b]{1.6cm}
    \centering
    \raisebox{-.5\height}{\includegraphics[width=\linewidth]{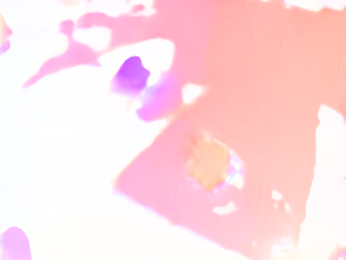}}
  \end{minipage}& 
    \begin{minipage}[b]{1.6cm}
    \centering
    \raisebox{-.5\height}{\includegraphics[width=\linewidth]{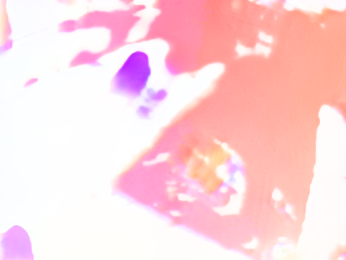}}
  \end{minipage}& 
    \begin{minipage}[b]{1.6cm}
    \centering
    \raisebox{-.5\height}{\includegraphics[width=\linewidth]{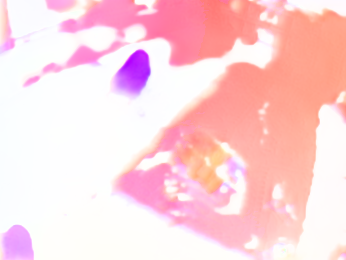}}
  \end{minipage} \cr
    \specialrule{0em}{0.5pt}{0.5pt}

    \begin{minipage}[b]{1.5cm}
    \centering
        \raisebox{-.5\height}{\scriptsize DCEIFlow} \raisebox{-.5\height}{Ours~($\bf{I_1}$+$\bf{E}$)}
  \end{minipage}& 
    \begin{minipage}[b]{1.6cm}
    \centering
    \raisebox{-.5\height}{\includegraphics[width=\linewidth]{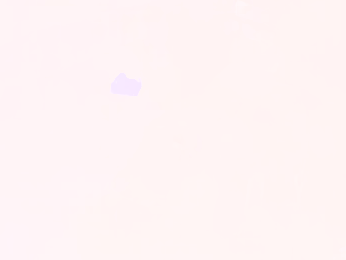}}
  \end{minipage}& 
    \begin{minipage}[b]{1.6cm}
    \centering
    \raisebox{-.5\height}{\includegraphics[width=\linewidth]{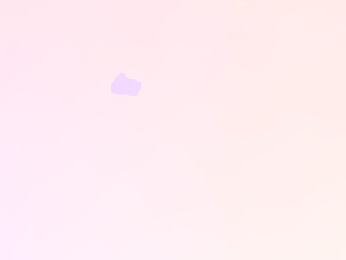}}
  \end{minipage}& 
    \begin{minipage}[b]{1.6cm}
    \centering
    \raisebox{-.5\height}{\includegraphics[width=\linewidth]{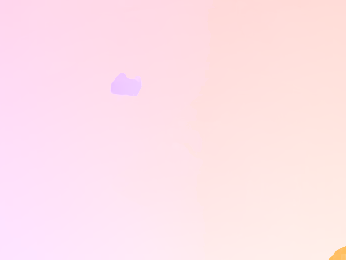}}
  \end{minipage}& 
    \begin{minipage}[b]{1.6cm}
    \centering
    \raisebox{-.5\height}{\includegraphics[width=\linewidth]{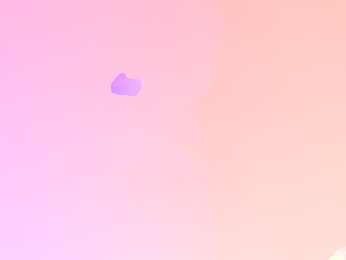}}
  \end{minipage}& 
    \begin{minipage}[b]{1.6cm}
    \centering
    \raisebox{-.5\height}{\includegraphics[width=\linewidth]{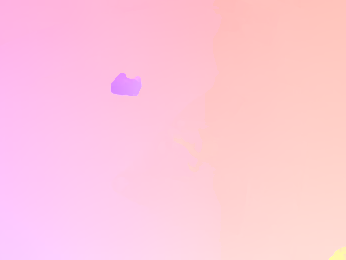}}
  \end{minipage}& 
    \begin{minipage}[b]{1.6cm}
    \centering
    \raisebox{-.5\height}{\includegraphics[width=\linewidth]{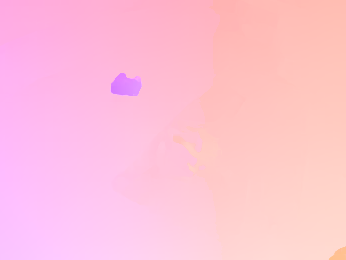}}
  \end{minipage}& 
    \begin{minipage}[b]{1.6cm}
    \centering
    \raisebox{-.5\height}{\includegraphics[width=\linewidth]{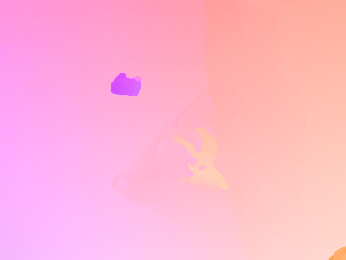}}
  \end{minipage}& 
    \begin{minipage}[b]{1.6cm}
    \centering
    \raisebox{-.5\height}{\includegraphics[width=\linewidth]{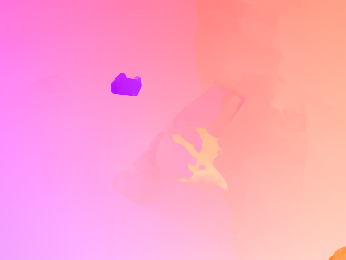}}
  \end{minipage}& 
    \begin{minipage}[b]{1.6cm}
    \centering
    \raisebox{-.5\height}{\includegraphics[width=\linewidth]{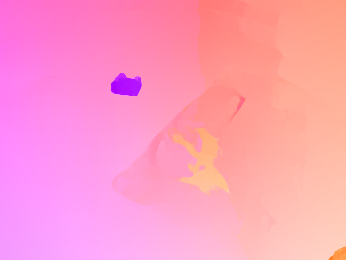}}
  \end{minipage} \cr
    \specialrule{0em}{0.5pt}{0.5pt}

    & 
    \begin{minipage}[b]{1.6cm}
    \centering
    $dt$=0.4
  \end{minipage}& 
    \begin{minipage}[b]{1.6cm}
    \centering
    $dt$=0.6
  \end{minipage}& 
    \begin{minipage}[b]{1.6cm}
    \centering
    $dt$=1.0
  \end{minipage}& 
    \begin{minipage}[b]{1.6cm}
    \centering
    $dt$=1.8
  \end{minipage}& 
    \begin{minipage}[b]{1.6cm}
    \centering
    $dt$=2.0
  \end{minipage}& 
    \begin{minipage}[b]{1.6cm}
    \centering
    $dt$=2.3
  \end{minipage}& 
    \begin{minipage}[b]{1.6cm}
    \centering
    $dt$=3.0
  \end{minipage}& 
    \begin{minipage}[b]{1.6cm}
    \centering
    $dt$=4.4
  \end{minipage}&
    \begin{minipage}[b]{1.6cm}
    \centering
    $dt$=5.0
  \end{minipage} \cr
    \specialrule{0em}{0.5pt}{0.5pt}
  \end{tabular}
  \vspace{-5pt}
  \captionof{figure}[foo]{\textbf{Another visual comparisons of continuous flow prediction with different time intervals on the EV-IMO dataset}. Best viewed on screen.
  }
  \label{viz:evimo_continuous_viz}
  \vspace{-10pt}
\end{table*}

\subsubsection{Dense and Continuous}

In Table~\ref{table:densesparse}, we evaluate the performance of dense flow prediction on MVSEC~\cite{eventdatasets:Zhu_MVSEC_RAL_2018}. 
Besides the dense and sparse (Event Masked) metrics, we also report the mean accuracy of pixels that do not trigger any events and have flow ground-truth (Event Excluded).
The event-only methods have consistently better masked results than the excluded results, even Stoffregen~\etal~\cite{eventflow:Stoffregen_ReducingGAP_ECCV_2020} and Baseline-EV use dense supervision.
On the contrary, the excluded results are better than those that use events with an image.
Our two models are much better than the others for each metric and frame interval setting. 
This shows the superiority of our framework in predicting dense flow using events.

For continuous flow estimation, we not only evaluate the results with $dt$=1 and $dt$=4 of each model on MVSEC in Table~\ref{table:mvsec_frame_result}, but also compare the results with the non-integer frame number time window on the EV-IMO dataset in Fig.~\ref{viz:continuous_viz} and Fig.~\ref{viz:evimo_continuous_viz}. 
Since there is no corresponding second frame image, the two-frame-based method cannot deal with non-integer intervals.
Our model obviously achieves higher accuracy and denser visualization results than other event-based methods.
This illustrates the advantages of our framework in predicting continuous and dense optical flow.

\begin{figure}[tbp]
\centering

\begin{minipage}[t]{\linewidth}
\scriptsize
\centering

\begin{minipage}[t]{0.22\linewidth}
    \centering
    \includegraphics[width=\textwidth]{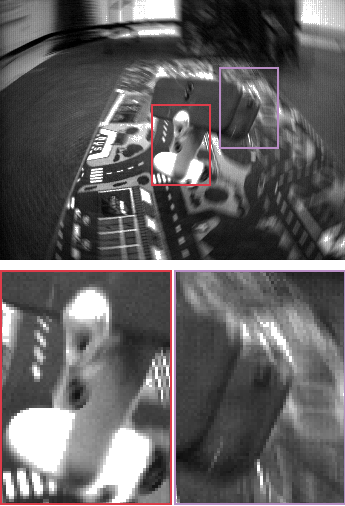} \\
    Image1
\end{minipage}%
\hspace{0.001\linewidth}
\begin{minipage}[t]{0.22\linewidth}
    \centering
    \includegraphics[width=\textwidth]{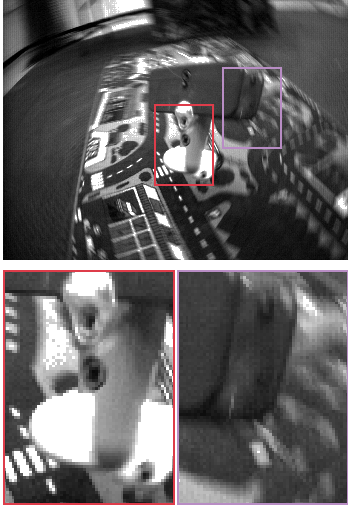} \\
    Image2
\end{minipage}%
\hspace{0.001\linewidth}
\begin{minipage}[t]{0.22\linewidth}
    \centering
    \includegraphics[width=\textwidth]{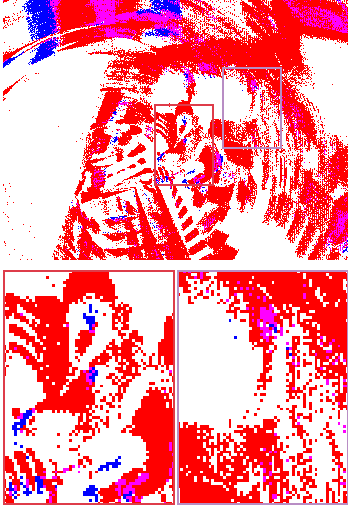} \\
    Events
\end{minipage}%
\hspace{0.001\linewidth}
\begin{minipage}[t]{0.22\linewidth}
    \centering
    \includegraphics[width=\textwidth]{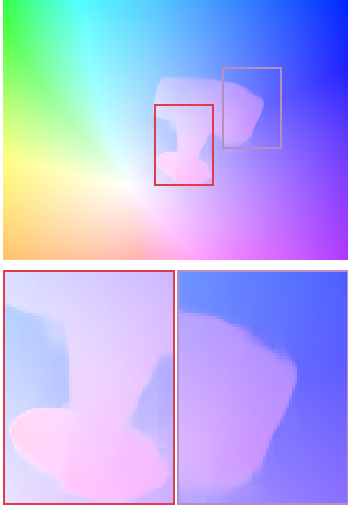} \\
    RAFT~\cite{flow:Teed_RAFT_ECCV_2020}
\end{minipage}%

\begin{minipage}[t]{0.22\linewidth}
    \centering
    \includegraphics[width=\textwidth]{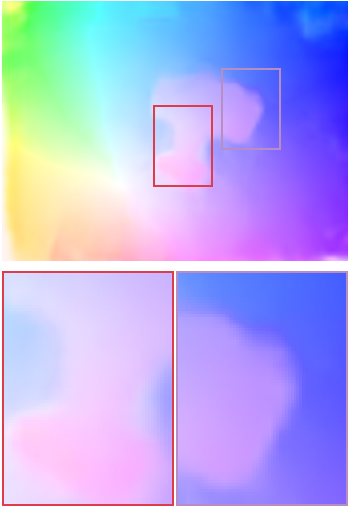} \\
    PWC-Net~\cite{flow:Sun_PWCNet_TPAMI_2019}
\end{minipage}%
\hspace{0.001\linewidth}
\begin{minipage}[t]{0.22\linewidth}
    \centering
    \includegraphics[width=\textwidth]{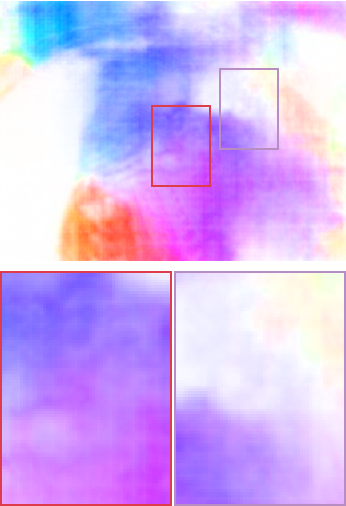} \\
    SpikeFlowNet~\cite{eventflow:Lee_SpikeFlowNet_ECCV_2020}
\end{minipage}%
\hspace{0.001\linewidth}
\begin{minipage}[t]{0.22\linewidth}
    \centering
    \includegraphics[width=\textwidth]{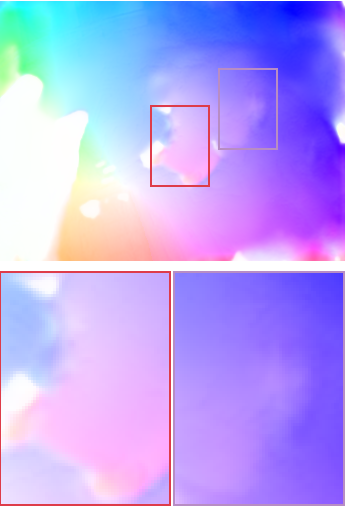} \\
    Stoffregen~\etal~\cite{eventflow:Stoffregen_ReducingGAP_ECCV_2020}
\end{minipage}%
\hspace{0.001\linewidth}
\begin{minipage}[t]{0.22\linewidth}
    \centering
    \includegraphics[width=\textwidth]{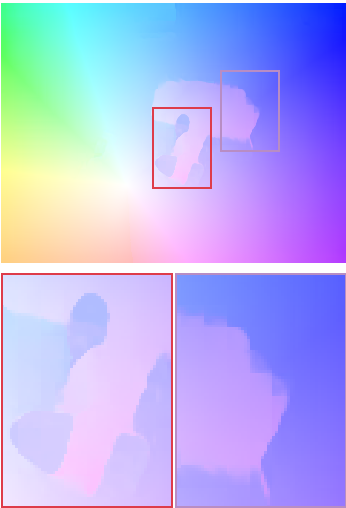} \\
    DCEIFlow~(Ours)
\end{minipage}%

\end{minipage}%

\begin{minipage}[t]{1\linewidth}
\scriptsize
\centering

\begin{minipage}[t]{0.22\linewidth}
    \centering
    \includegraphics[width=\textwidth]{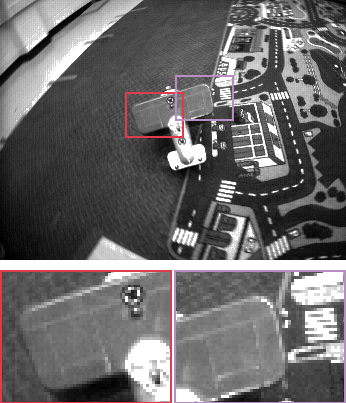} \\
    Image1
\end{minipage}%
\hspace{0.001\linewidth}
\begin{minipage}[t]{0.22\linewidth}
    \centering
    \includegraphics[width=\textwidth]{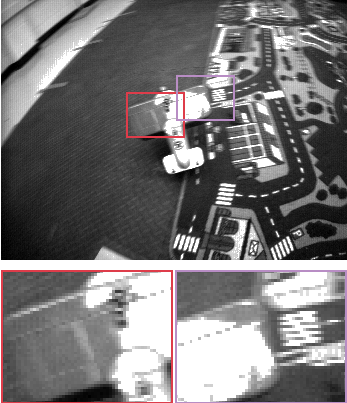} \\
    Image2
\end{minipage}%
\hspace{0.001\linewidth}
\begin{minipage}[t]{0.22\linewidth}
    \centering
    \includegraphics[width=\textwidth]{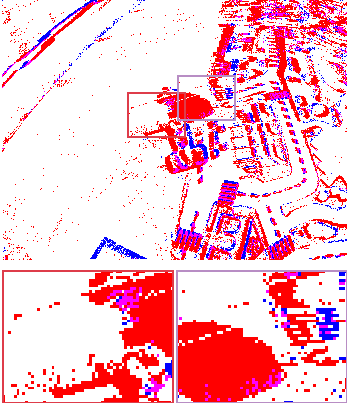} \\
    Events
\end{minipage}%
\hspace{0.001\linewidth}
\begin{minipage}[t]{0.22\linewidth}
    \centering
    \includegraphics[width=\textwidth]{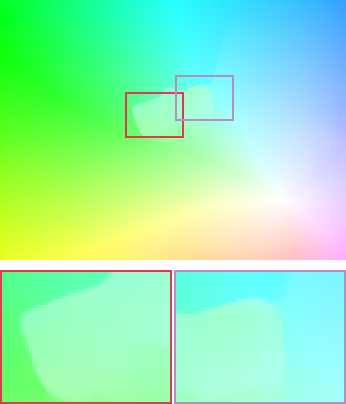} \\
    RAFT~\cite{flow:Teed_RAFT_ECCV_2020}
\end{minipage}%

\begin{minipage}[t]{0.22\linewidth}
    \centering
    \includegraphics[width=\textwidth]{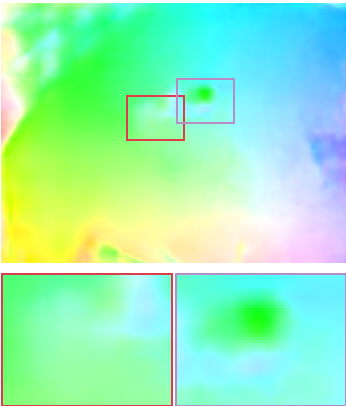} \\
    PWC-Net~\cite{flow:Sun_PWCNet_TPAMI_2019}
\end{minipage}%
\hspace{0.001\linewidth}
\begin{minipage}[t]{0.22\linewidth}
    \centering
    \includegraphics[width=\textwidth]{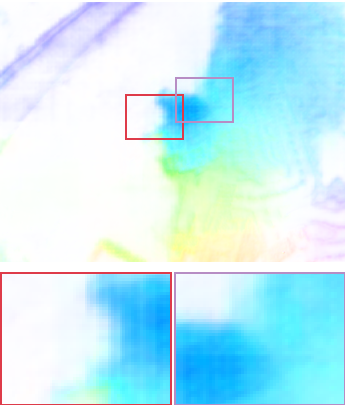} \\
    SpikeFlowNet~\cite{eventflow:Lee_SpikeFlowNet_ECCV_2020}
\end{minipage}%
\hspace{0.001\linewidth}
\begin{minipage}[t]{0.22\linewidth}
    \centering
    \includegraphics[width=\textwidth]{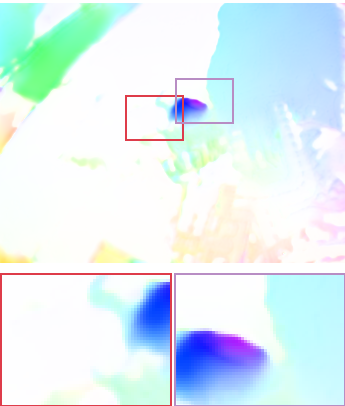} \\
    Stoffregen~\etal~\cite{eventflow:Stoffregen_ReducingGAP_ECCV_2020}
\end{minipage}%
\hspace{0.001\linewidth}
\begin{minipage}[t]{0.22\linewidth}
    \centering
    \includegraphics[width=\textwidth]{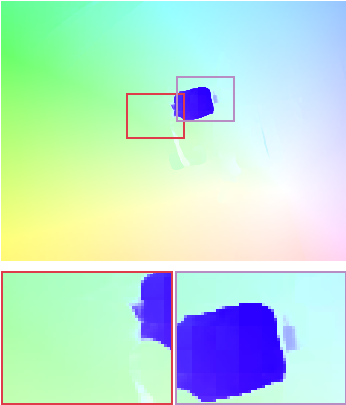} \\
    DCEIFlow~(Ours)
\end{minipage}%

\end{minipage}%

\centering
\vspace{-5pt}
\caption{\textbf{Failure cases on the EV-IMO dataset.} Best viewed on screen.}
\label{viz:failure_cases}
\vspace{-10pt}
\end{figure}

\subsection{Failure cases and limitation}

From the previous experiments, our predictions are better than the two-frame-based start-of-the-art approach RAFT~\cite{flow:Teed_RAFT_ECCV_2020} in some cases, especially in the detailed areas, but sometimes worse.
This aroused our interest in further analysis. 
We found two examples on the EV-IMO~\cite{eventdatasets:mitrokhin_EVIMO_IROS_2019} dataset, as shown in Fig.~\ref{viz:failure_cases}. 
For the first example, the tail part lacks texture, and events are rarely triggered in such weakly textured regions. Thus for our setting with a single image and events, our model outputs incorrect predictions in the tail part of the toy airplane. 
We think this shows that the two-frame-based methods such as RAFT can still match the structural association in these challenging regions. 
For the second example, a part of the object suddenly reflects light. 
However, the brightness change is not caused by motion, and the object did not move so much. 
The reflection produces a lot of events, which seriously affects the predictions of event-based methods, including ours. 

For event-based applications, the efficiency of the algorithm is worth considering. 
Using a single 2080ti GPU, our model takes 28ms to process data with MVSEC size. 
This result meets the \textit{real-time} standard (30 fps) and is much better than the model-based optimization method Pan~\etal~\cite{eventapp:Pan_BringingBlurryEvent_CVPR_2019}, which requires an uncertain running time of more than 1 second.
However, compared to EV-FlowNet~\cite{eventflow:Zhu_EVFlowNet_RSS_2018} and SpikeFlowNet~\cite{eventflow:Lee_SpikeFlowNet_ECCV_2020} that use only events and require only 5ms and 15ms, we have achieved better results with increased computation cost. 
Higher time consumption will limit the application, which is what we need to improve next.

%%%%%%%%%%%%%%%%%%% Conclusion %%%%%%%%%%%%%%%%%%%%%
\section{Conclusion}
In this paper, we have proposed a deep learning-based dense and continuous optical flow estimation approach from a single image with event streams.
Our network can effectively exploit the internal relation of two different modalities of data through an event-image fusion and correlation module, and predict the dense optical flow by the iterative flow update network structure, combined with our bidirectional training strategy. 
Thus our framework can reliably estimate dense flow as two-frame-based methods, as well as estimate continuous flow as event-based methods.
Extensive experimental evaluation on multiple datasets demonstrates the superiority of our proposed framework in estimating dense and continuous optical flow compared with existing state-of-the-art event-only or fused single-image methods.

\ifCLASSOPTIONcaptionsoff
  \newpage
\fi

\bibliographystyle{IEEEtran}
\bibliography{CV_Abbreviation, Event_Flow_Reference}

\begin{IEEEbiography}[{\includegraphics[width=1in,height=1.25in,clip,keepaspectratio]{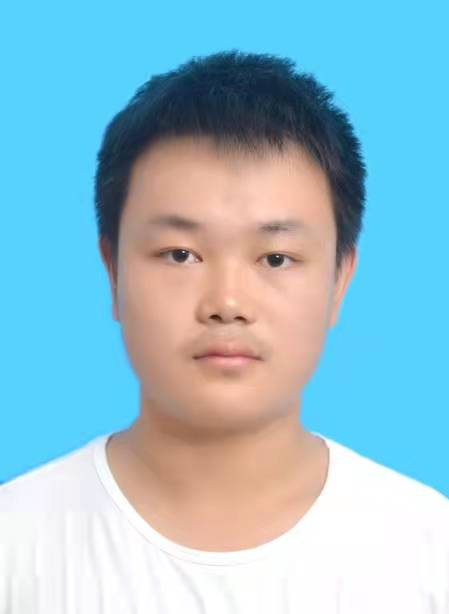}}]{Zhexiong Wan}
is currently a PhD student with School of Electronics and Information, Northwestern Polytechnical University, Xi'an, China. He received his Bachelor of Engineering degree from Northwestern Polytechnical University in 2019.
\end{IEEEbiography}

\begin{IEEEbiography}[{\includegraphics[width=1in,height=1.25in,clip,keepaspectratio]{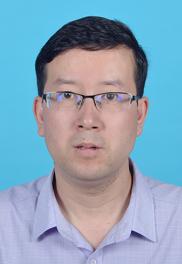}}]{Yuchao Dai}
is currently a Professor with School of Electronics and Information at the Northwestern Polytechnical University. He received the B.E. degree, M.E degree and Ph.D. degree all in signal and information processing from Northwestern Polytechnical University, Xi'an, China, in 2005, 2008 and 2012, respectively. He was an ARC DECRA Fellow with the Research School of Engineering at the Australian National University, Canberra, Australia from 2014 to 2017 and a Research Fellow with the Research School of Computer Science at the Australian National University, Canberra, Australia from 2012 to 2014. His research interests include 3D vision, multi-view geometry, low-level computer vision, deep learning, and optimization. He won the Best Paper Award in IEEE CVPR 2012, Best Paper Nominee in IEEE CVPR 2020, the DSTO Best Fundamental Contribution to Image Processing Paper Prize at DICTA 2014, the Best Algorithm Prize in NRSFM Challenge at CVPR 2017, the Best Student Paper Prize at DICTA 2017 and the Best Deep/Machine Learning Paper Prize at APSIPA ASC 2017. He served/serves as Area Chair at CVPR, ICCV, ACM MM, ACCV, etc. He serves as the Publicity Chair at ACCV 2022 and the Distinguished Lecturer of APSIPA.
\end{IEEEbiography}

\begin{IEEEbiography}[{\includegraphics[width=1in,height=1.25in,clip,keepaspectratio]{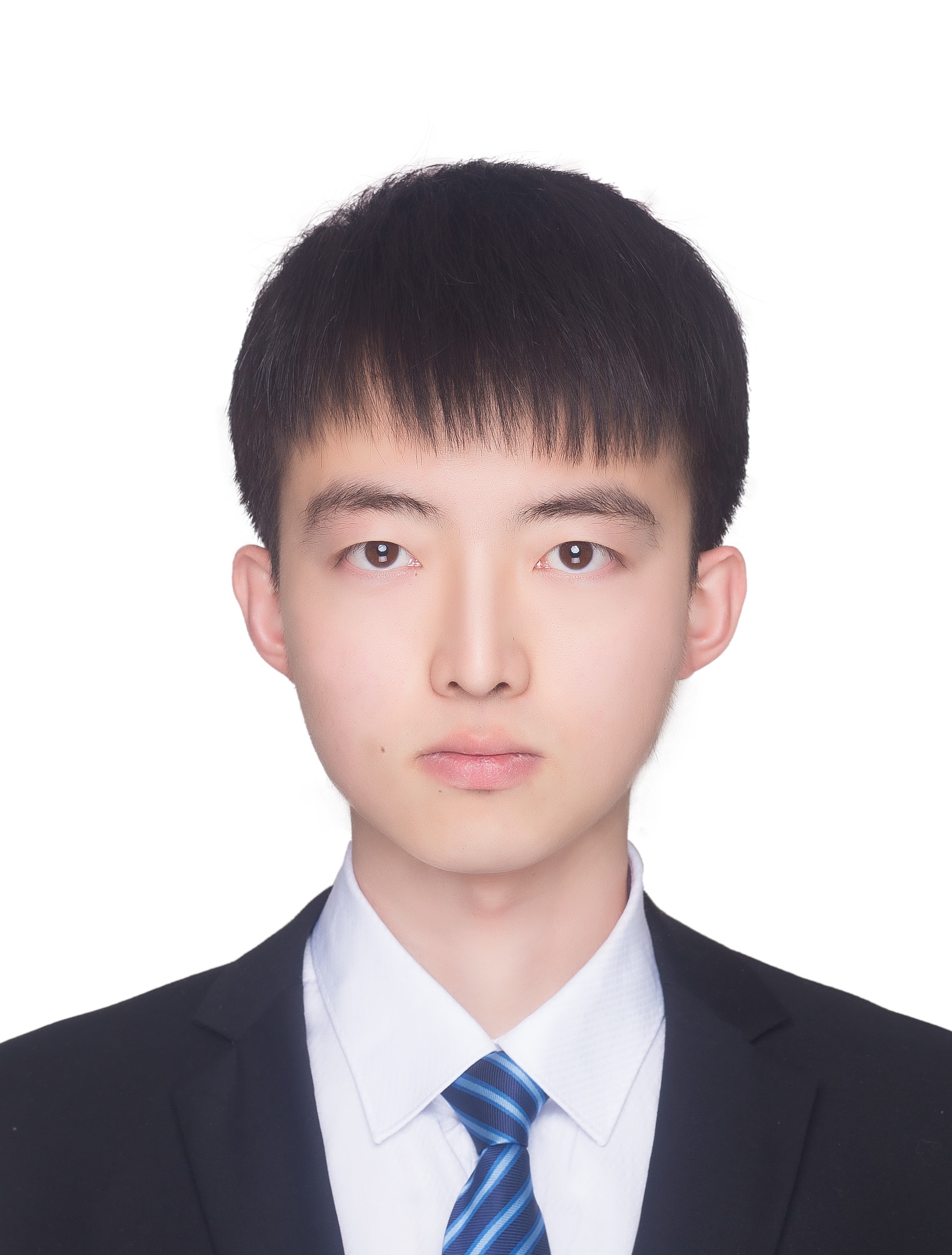}}]{Yuxin Mao}
is currently a PhD student with School of Electronics and Information, Northwestern Polytechnical University, Xi'an, China. He received his Bachelor of Engineering degree from Southwest Jiaotong University in 2020. He won the best Paper Award Nominee at ICIUS 2019.
\end{IEEEbiography}

\end{document}